%% file: tesi.tex
\documentstyle[a4wide,tesi,11pt]{book}

\include{epsf}

\author{Llu\'{\i}s Padr\'o i Cirera}
\title{A Hybrid Environment for Syntax-Semantic Tagging}
\director{Horacio Rodr\'{\i}guez Hontoria}
\dept{Departament de Llenguatges i Sistemes Inform\`atics}
\what{Inform\`{a}tica}
\where{Facultat d'Inform\`{a}tica de Barcelona}
\submitdate{15 de Desembre de 1997} 

\begin{document}
\stdtitlepage           
\pagestyle{empty}

\newpage
\mbox{}{}
\newpage

\vglue 3cm
\begin{flushright}
A la meva fam\'{\i}lia.\\
A la In\'es.
\end{flushright}

\newpage
\mbox{}{}
\newpage

{\bf Agra\"{\i}ments}
\bigskip

   Aquesta tesi no hauria estat possible sense l'ajuda i la col$\cdot$laboraci\'o 
de moltes persones a qui vull donar les gr\`acies.
\medskip

   La tasca m\'es feixuga i continuada ha caigut sobre les espatlles de
l'Horacio Rodr\'{\i}guez. Ha estat sempre disponible, amb un consell
encertat, una sugger\`encia interessant, una frase encoratjadora. 
La seva dedicaci\'o i paci\`encia han anat molt m\'es enll\`a del que
s'espera d'un director de tesi per arribar al que s'espera d'un amic.
\medskip

   Tamb\'e vull agra\"{\i}r a la Carme Torras i el Pedro Meseguer les 
orientacions que em van donar al principi d'aquesta recerca. A la primera 
li dec la idea que em va posar en aquest cam\'{\i}. Al seg\'on li agraeixo 
el seu inter\`es i sugger\`encies.
\medskip

   Una menci\'o molt especial mereixen els meus companys de feina. Especialment
per al Llu\'{\i}s M\`arquez, de qui m'he aprofitat del seu rigor i la seva 
capacitat de treball i pel German Rigau de qui he apr\`es el que \'es tenir 
esperit cient\'{\i}fic.

   Tamb\'e cal mencionar que aquesta tesi l'ha facilitat enormement l'agradable 
ambient que fan possible tots els altres components del Grup de Recerca en 
Processament del Llenguatge Natural. Voldria mencionar especialment a: Alicia Ageno, 
Jordi \'Alvarez, Jordi Atserias, N\'uria Castell, Irene Castell\'on, Salvador Climent, 
Xavier Farreres, Marta Gatius, Toni Mart\'{\i}, Mariona Taul\'e i Jordi Turmo.

   Un altre factor ambiental del que n'he tret profit es de donar la doc\`encia
a Vilanova. Cal atribu\"{\i}r a Rafel Camps, Neus Catal\`a, Jordi Daud\'e, Jordi Esteve,
\`Angels Hern\'andez, Mario Mart\'{\i}n i Anna Rosell\'o el bon ambient del que
he disfrutat els anys que ha durat aquesta recerca.
\medskip

   Tamb\'e vull agra\"{\i}r al Departament de Llenguatges i Sistemes Inform\`atics 
haver-me facilitat enormement la feina amb desc\`arregues docents. Sense elles, 
la feina d'aquesta tesi estaria un any endarrerida.
\medskip

    Pel que fa al cap\'{\i}tol personal, vull agra\"{\i}r als meus pares 
i germans el seu suport. I a la In\'es el seu entusiasme en els moments 
bons i la seva comprensi\'o i recolzament en els dif\'{\i}cils.
\medskip

   Aquest treball ha estat parcialment finan\c{c}at pels seg\"uents projectes
i iniciatives:
\smallskip

\indent ESPRIT-BRA 7315. Projecte ACQUILEX II \\
\indent CICYT. TIC96-1243-C03-02. Projecte ITEM \\
\indent EU. LE4003. Projecte EuroWordNet \\
\indent CIRIT. Grup de Recerca de Qualitat 1995SGR-00566
\newpage

{\bf Acknowledgments}
\bigskip
 
   This thesis wouldn't have been possible without the aid and collaboration
of many people whom I wish to thank.
\medskip

    The hardest and longest task has been on Horacio Rodr\'{\i}guez's shoulders.
He has been always available, with an accurate advice, an interesting suggestion
or an encouraging word. His devotion and patience has gone far beyond what one
expects from an advisor to get to what one expects from a friend.
\medskip

   I also want to thank Carme Torras and Pedro Meseguer for their orientations
in the early steps of this research. I owe the former the idea that put me on
this path. I thank the later for his interest and suggestions.
\medskip

   A very especial mention is deserved by my work mates. Specially by 
Llu\'{\i}s M\`arquez, whose rigour and working capacity I have taken advantage
of, and by German Rigau, from who I have learnt what scientific spirit is.

   It must also be stated that the friendly environment made possible by all
the members of the Natural Language Research Group has been a facility for
writing this thesis. I would like to specially mention: Alicia Ageno,
Jordi \'Alvarez, Jordi Atserias, Núria Castell, Irene Castell\'on, Salvador Climent, 
Xavier Farreres, Marta Gatius, Toni Mart\'{\i}, Mariona Taul\'e and Jordi Turmo.

   Another environmental factor I have taken advantage of is teaching in
Vilanova. I must thank Rafel Camps, Neus Catal\`a, Jordi Daud\'e, Jordi Esteve,
\`Angels Hern\'andez, Mario Mart\'{\i}n and Anna Rosell\'o the friendly environment
I have enjoyed during these years.
\medskip

   I also want to thank the Software Department for facilitating my research 
by means of teaching reductions. Without them, this work would be one year behind.
\medskip

   As for the personal gratitude, I want to thank my parents and siblings for their
support. And In\'es for her enthusiasm in the good moments and her sympathy
and support in the hard ones.
\medskip
  
   This work has been partially funded by the following projects and initiatives:
\smallskip

\indent ESPRIT-BRA 7315. Projecte ACQUILEX II \\
\indent CICYT. TIC96-1243-C03-02. Projecte ITEM \\
\indent EU. LE4003. Projecte EuroWordNet \\
\indent CIRIT. Grup de Recerca de Qualitat 1995SGR-00566
\newpage

\pagestyle{headings}
\pagenumbering{roman}
\bibliographystyle{alpha}
\tableofcontents

\include{introduction}
\include{state-of-art}
\include{application}
\include{experiments}
\include{results}
\include{conclusions}

\addcontentsline{toc}{chapter}{Bibliography}
\bibliography{}

\addtocontents{toc}{\contentsline {chapter}{Appendices}{}}
\appendix
\include{appendix1}

\include{appendix2}

\end{document}

%% file: introduction.tex
\chapter{Introduction}
\pagenumbering{arabic}
\label{cap:introduction}

  In recent years a growing amount of researchers on natural language 
processing (NLP) feel that the problems traditionally addressed
separately should --as available resources enable it-- be addressed 
as a whole. For instance, \cite{Wilks96b} showed that knowing the
part-of-speech tag of a word can help to disambiguate its sense in
a high percentage of the cases, thus, a system performing word
sense disambiguation using not only the context information related 
to words and senses but also part-of-speech information, would 
have higher performance. This idea is also present in works like 
those of \cite{Jung96,Ng96}, which presented models able to combine
different kinds of statistical information.

  This statement is quite obvious, since it seems logical that the more
information we have, the better results we will produce at a given task. 
But if we take this idea twofold, we can use each kind of information to help
to disambiguate the other {\em at the same time}, e.g. we can perform
POS tagging and WSD simultaneously, using all information available, 
and taking advantage of the interactions between the different kinds 
of information. This is more or less what we humans do when
understanding a NL utterance: we use all kinds of information
--lexical, syntactical, semantic, etc.-- at the same time to cut
out improper analysis and pick the right one. 
\medskip

In this thesis we are interested in the use of flexible algorithms
that can handle different kinds of information (semantic, syntactic, \ldots)
using different kinds of knowledge (linguistic, statistical, \ldots), in the style of 
Constraint Grammars \cite{Karlsson95}, where the properties that may be owned
by a word or referred to by a constraint are only limited by
which ones are available. Getting over the historical controversy 
between linguistic and knowledge-based and statistical methods, 
numerical information about natural language behaviour must not 
be let out, since work in recent years 
\cite{Klavans94,Jung96,Hajic97,Pedersen97a,DellaPietra97,Ristad97b}
confirms that it may deal very accurately with language ambiguities.
\medskip

The task of finding a set of relationships or interactions between all
information kinds such that they describe natural language behaviour,
has the category of language modelling and involves
linguistic, cognitive and psychological considerations which are 
beyond the scope of this thesis. Anyway, since our system can not work
without a reasonable language model, we will also use several 
existing alternatives for acquiring one, ranging from manual
development to n-gram collection, through the use of machine learning
algorithms.

\section{Goals of this Research}
\label{sec:goals}

  This thesis describes research into the use of energy--function 
optimization algorithms to solve natural language processing tasks. 
The main objective is to show that such algorithms can deal with 
hybrid information: combining statistical and linguistic information, and
with different classification dimensions (e.g. POS tags, senses, etc.).

  The problems addressed are mainly those of disambiguation nature, 
that is, those where the task to be done consists of disambiguating 
a given sequence of words somehow ambiguous (part-of-speech, syntactic
function, word sense, etc.). Most of the NLP tasks where a value has
to be assigned to a feature can be seen as disambiguation problems, since 
the task can be summarized as picking the most appropriate value from
a known set of possibilities.

  The optimization algorithm focused on is relaxation labelling, since  
there is a clear structural matching between disambiguation-like problems 
and the tasks the algorithm naturally applies to. The algorithm chooses
the most suitable label for each of the variables in the model. Our work will
consist of modelling the NLP task we want to perform in an appropriate
way for the algorithm.

\subsection{Finding a flexible NL modelling}
\label{sec:goals-1}

  The objective of enabling a hybrid model requires a way to express NL properties
that is able to include all kinds of information. This means that if we want to 
perform POS-tagging, we do not have to limit ourselves to use POS information about the words
in the sentence, but we can also include any information available: semantic, syntactic, 
morphological, etc.

  In addition, we want our model to be able to cope with imprecise or
incomplete information, and with flexible relationships between NL elements,
i.e. we want a robust model that can produce a reasonable result when faced to
a non-expected case. So we need to introduce a numerical, statistical, or 
probabilistic component in our model.
\medskip

  The way in which we will try to achieve this kind of model is the following:
We will use context constraints to express the relationships between linguistic
elements. These constraints will admit any kind of available linguistic information.
The choice of constraint modelling enables us to describe a wide range of patterns,
from a simple bigram --expressed as a constraint between two consecutive word
positions-- to a complex structure relating different features of several
words --e.g. checking the existence of an auxiliary verb to the left of 
a given word with no occurrences of a noun in between--.

  The possibility of using statistical information will be
introduced by assigning to each constraint a numerical value, which will 
be interpreted as its {\em weight} or {\em strength}, that is, as how strictly
must be that constraint applied. This enables pure classical linguistic models
--where all constraints are strictly applied--, statistical models, where all
constraints have a weight computed through some statistical method, or any 
hybrid model where some constraints are strictly applied and some others are not.

\subsubsection{Constraint Satisfaction}
\label{sec:g1-CSP}

  As described in the previous section, we chose our model to be a weighted 
constraint one. So, the disambiguation problems will consist of applying
the constraints and finding the combination that satisfies them all (or, 
at least, as many of them as possible). The natural approach to these problems are
constraint satisfaction algorithms.

  Since many useful and interesting problems can be stated as a constraint satisfaction
problem --travelling salesman, n-queens, corner and edge recognition, 
image smoothing, etc. \cite{Lloyd83,Richards81,Aarts87}--  this is a field where
we find many algorithms that have been long used to solve them. 
\medskip

  The best-known are those of basic operational research, such as gradient step or 
relaxation --for continuous spaces-- or mathematical programming --for discrete spaces--.
 In the later case, we can consider the optimization as a search in a state space, and
use classical artificial intelligence algorithms, from depth-first or breadth-first global
search to more sophisticated heuristic search algorithms such as hill-climbing, best-first 
or $A^*$.

\subsubsection{Relaxation Labelling}
\label{sec:g1-RL}

  Although any of the algorithms mentioned in the previous section could be
used to process a constraint model, we want to deal with {\em weighted} constraints,
which requires the algorithm to be able to move in a continuous space. This 
leads us to choose relaxation labelling since its objective function is 
expressed in terms of constraints, which makes it more suitable to our needs
than gradient step or other optimization algorithms for continuous space 
such as neural nets, genetic algorithms or simulated annealing which do not use 
constraints in such a natural way as relaxation labelling does. Different optimization
algorithms will be compared in section \ref{sec:optimization}.
\medskip

  Relaxation labelling is a well-known technique used to solve consistent labelling
problems (CLP).  The algorithm finds a combination of values for a set of variables such 
that satisfies -to the maximum possible degree- a set of given constraints. 
Since CLPs are closely related to 
constraint satisfaction problems \cite{Larrosa95a}, relaxation labelling will be a suitable
algorithm to apply our constraint-based language model. In addition,
since all of them perform function optimization based on
local information, relaxation is closely related to neural nets \cite{Torras89}
and gradient step \cite{Larrosa95b}. 

  Relaxation operations had been long used in 
engineering fields to solve systems of equations \cite{Southwell40}, but 
they got their biggest success when the extension to symbolic domain
--relaxation labelling-- was applied to constraint propagation field, 
specially in low-level vision problems \cite{Waltz75,Rosenfeld76}.
The possibility of applying it to NLP tasks was recently pointed out 
by \cite{Pelillo94a,Pelillo94b} who use a toy POS tagging problem to 
evaluate their method to estimate compatibility values.

\subsection{Application to Different NL tasks}
\label{sec:goals-2}

   A secondary goal of this research is proving that our approach works in practice,
applying it to several NLP tasks. As stated above, the most natural tasks for
this approach are those of disambiguation nature, so we will test our system in
this kind of tasks. Namely, at part-of-speech tagging, at combined POS-tagging plus 
shallow parsing, and at combined POS-tagging plus word sense disambiguation.
\medskip

  Part-of-speech tagging is the most widely known disambiguation problem in NLP, and
the results obtained by current systems are probably the best results ever obtained in
a NLP task. This is due in part to the irruption of statistical methods in this field
in the late 80's, but the good results are also reflecting that this task is structurally
simpler than others, and that a simple method can solve a great part of it. Nevertheless, 
the ambiguities that remain unresolved frequently belong to the class of those which could 
only be solved through the use of higher level information.

   We will apply relaxation labelling to POS-tagging, and check whether the addition
of higher level information results in a performance increase. 
 We will also use POS-tagging as a base problem to test the influence of cross-information
when solving different NLP tasks simultaneously.
\medskip

  Word sense disambiguation is a task right opposite to POS-tagging with respect
to complexity and achieved results.
From the impossible consensus on what should be considered a 
sense to the almost inexistent test set to perform experiments through the intrinsic 
task difficulty, the obstacles
that the researcher in this task must overcome are much greater than in the previous
case, and thus, the results reported by current works are much further away from what 
could be desired.

  While current methods tend to use only one kind of knowledge, we will 
try to solve WSD simultaneously with POS-tagging, to check whether the coalition of
both yields better results than each one of them separately.
\medskip

  Shallow parsing is a recent idea, which is half way between POS-tagging and 
parsing. It consists of assigning to each word --or at least to each important one--
its syntactic function, but only superficially, not building the whole parse tree.

  Shallow parsing and POS tagging are closely related: knowing that the part-of-speech
for a word is, say, {\em verb}, discards the possibility for that word to be the 
subject of the sentence, and the other way round: when the parser decides that
a word is acting as subject of a given verb, it is implying that it must have a
nominal part of speech. This relationship subscribes the idea that both tasks
can be solved in parallel combining the knowledge needed to solve each of them.
\medskip

  In all cases, we will combine language models obtained from different sources:
statistics collection, linguist-written rules and machine-learned rules.

\section{Setting}
\label{sec:setting}

\subsection{Utility}
\label{sec:utility}
  From a general perspective, constraint satisfaction and optimization
algorithms may be useful to NLP purposes, since they enable a basis where 
language models which take into account many linguistic phenomena and 
features as well as different relationships among them may be easily applied 
to real linguistic data.

  In addition, the model is not restricted, that is, it can be built
incrementally and it also allows the merging of information obtained from many
different sources. 
\medskip

  From a more specific point of view, a system as the one we are proposing enables 
linguists to combine different kinds of information to perform a single task, 
or even perform several disambiguation tasks in parallel, taking advantage of 
cross information between the different knowledge sources. This not only 
should help to improve the results that current systems obtain at tasks such
as part-of-speech tagging or word sense disambiguation, but it also opens a path
towards the development of wide--coverage knowledge--integrated linguistic models 
and its application to real data. 
\medskip

  The utility of such disambiguation tasks is well known: POS-tagging is very useful
in reducing the ambiguity amount that a parser must deal with \cite{Wauschkuhn95},
it is also used in speech recognition to anticipate the probabilities of the next word
to come and thus reduce the ambiguity \cite{Heeman97}, and it can also 
be used to extract syntactic knowledge from annotated corpora, for instance 
via grammatical inference \cite{Pereira92,Charniak93,Smith95,Lawrence95}. 

  Word sense disambiguation is a much more difficult task, and its obvious utilities
are the ambiguity reduction for further applications such as information retrieval, machine 
translation, document classification, etc. From a more linguistic or lexicographic 
point of view it can be used to study or to extract knowledge on selectional 
restrictions, sense co-occurrences, different uses of the same word, etc.

\subsection{Approaches}
\label{sec:approach}

  The current approaches to disambiguation problems such as POS-tagging or WSD,
can be classified in two broad families. The classical and most straightforward
is the linguistic approach, which 
uses linguist-written language models. Recently the statistical approach has achieved
great success due to the good results it yields using easily obtainable models based 
either on collecting statistics from a training corpus or using
machine-learning algorithms to extract the language model from
that training corpus.
\medskip

  The linguistic models are developed by introspection. This makes it a
high labour cost work to obtain a good language model. Transporting the model to
other languages means starting over again. They usually do not consider frequency
information and thus have a limited robustness and coverage. Their advantages are 
that the model 
is written from a linguistic point of view and explicitly describes linguistic
phenomena, and that the model may contain many and complex kinds of knowledge.
\medskip

  The statistical approaches are based on collecting statistics from existing corpora,
either tagged (supervised training) or untagged (unsupervised training). This makes 
the model development much shorter --specially
in the unsupervised version-- and the transportation to other languages much easier,
provided there are corpora in the desired language. They take into account frequency
information, which gives them great robustness and coverage. 

  The statistical approaches can be divided in two classes, according to the complexity
of the statistical model they acquire:

  First, we have the simple--model class, where
the language model consists of a set of co-occurrence frequencies for some predetermined
features. Typical representatives of this class are n-gram based models for
part-of-speech tagging or word form co-occurrence models for word sense disambiguation.
The main disadvantages of
these models are that they collect only simple information (usually co-occurrences) and 
that the language model is neither explicit (it is only a set of frequencies)
nor has any linguistically significant structure. 

  Second, there is the complex--model class which
consists of using a machine--learning algorithm to automatically acquire a high-level
language model from a training corpus, The knowledge acquired may take the form
of rules, decision trees, frames, etc. but it will be more complex than a simple
set of frequency countings.
In this case, the model is explicit, since usual machine--learning algorithms produce
symbolic knowledge, but it does not necessarily have any linguistic meaning. 
\medskip

 The previously described methods are approaches to acquire a language model. Once
the model is achieved it is applied through some algorithm to perform some NLP task. 
The model-applying algorithm is usually very dependent on the kind of model and task,
so a different model and algorithm is needed for each different task.

  We will present in this thesis the use of relaxation labelling algorithm to perform
NLP tasks, and we will show that it can be used either with models belonging to any of
the different families described above or with hybrid models. We will show also that
if the model contains information to perform different NLP tasks, the algorithm is
able to solve these tasks simultaneously.

\section{Summary}
\label{sec:summary}

\subsection{Contributions}
\label{sec:contrib}

  The research described in this thesis includes new contributions in the following 
aspects:

\subsubsection{Use of optimization techniques in NLP}
\label{sec:contrib-1}

   The main contribution of this work is taking a step further into the use
of optimization techniques to process natural language. The successful use of the
relaxation labelling algorithm confirms that previous works which used simulated 
annealing or neural nets were in a promising path. This approach enables the 
modelling of language through sets of constraints and through objective functions,
which can be optimized locally or globally --depending on the compromise efficiency vs. 
accuracy one wants to take-- using a suitable algorithm. It also makes the
application algorithm independent of the language model.
\medskip

   The main difficulty presented by this approach is the modelling of language in 
a way that enables the use of optimization algorithms. The proposed weighted constraint
model is only one possibility --other algorithms may require different modellings-- 
which seems adequate to the use of relaxation labelling
algorithm, while keeping the attractive of being easily readable and the ability
of accepting either manually written or automatically derived constraints.
 Other algorithms may require a different modelling

\subsubsection{Application of multi-feature models}
\label{sec:contrib-2}

  The used language model is based on context constraints which restrict
the values that a word feature may take, depending on the features of
neighbour words. 
It is able to represent different features for each word (i.e. part-of-speech,
lemma, semantic properties, etc.). Neither the number of features nor 
their meaning are restricted in any way. 

  The language model consists of a set of constraints, which relate the 
features of a word with those of the words in the context, and state whether
that situation is very likely or very unlikely to happen. 

  Since this schema is similar to that of Constraint Grammars \cite{Karlsson95}, 
we will use their formalism because its expressive power suits our needs and
its widespread diffusion will simplify the task of obtaining hand-written
language models.
  Original Constraint Grammars only state if such situations are 
possible and must be selected ({\bf SELECT} constraints) or
impossible and must be discarded ({\bf REMOVE} constraints).
Our extension introduces a new class of constraints: those to
which a numerical {\em compatibility value} is assigned. This value may 
range from a large positive value (very compatible) to a 
large negative value (very incompatible) with all 
intermediate degrees. The SELECT/REMOVE constraints are 
interpreted as stating a very strong compatibility/incompatibility value.

\subsubsection{Application of statistical-linguistic hybrid models}
\label{sec:contrib-3}

  The choice to model language through a set of constraints, each of 
them associated to a compatibility value, makes it possible to merge 
knowledge acquired from multiple sources. The way to achieve this is 
converting the different source knowledges to the common formalism of
our language model. 
\medskip

  The hand--written constraints can be written directly in the desired
formalism, and the automatically obtained models can be easily translated
to the common representation based on weighted constraints. 

 For instance,
a n-gram model can be converted to a set of constraints --one for each 
n-gram-- which will have a compatibility value computed according to
the n-gram probability. In the same way, a machine--learned model 
consisting of statistical decision trees can be converted into a set of 
constraints --one for each tree branch-- with a compatibility value computed 
from the conditional probabilities of the leaf nodes. Also, a representation consisting
of a vector of context words for each sense (as in \cite{Yarowsky92})
could be converted to a set of constraints --one for each pair sense plus 
context word in its vector-- with a compatibility depending on the relevance 
probability of that pair. Equally, a model based on conceptual distance 
between senses (e.g. \cite{Sussna93}) can be converted to constraints with 
compatibility values computed from that distance measure. 

All those compatibility values can be computed from
probabilities in many ways, as detailed in section \ref{sec:constraints}.
\medskip

  So, we can produce hybrid models with constraints obtained from any source
combining them in any desired proportion, by means of translating them all
to a common formalism. In our case we chose that formalism to be an extension 
of Constraint Grammars, due to their flexibility, successful performance and 
widespread diffusion.

\subsubsection{Simultaneous resolution of NLP tasks}
\label{sec:contrib-4}

  Due to the multi-feature nature of constraints, and to the
parallel way in which relaxation applies them, the algorithm 
can select simultaneously the combination of values for
several linguistic features that best suit a word in a certain context,
that its, it can solve different NLP disambiguation tasks at the same time,
taking advantage of the interactions between them.
\medskip

  For instance, if we have POS and sense ambiguities, we will have 
for each word several possible readings in the form of pairs $(POS,sense)$.
The model can contain constraints selecting --or refusing-- one POS, one sense, 
or one pair for that word in the current context. Obviously,
the constraints on only one feature, will affect all the pairs that
contain it. At the end, the pair with has collected more positive
contributions will be selected and thus a POS and a sense will be assigned to  
the word, i.e. it will have the two features disambiguated,
and the disambiguation has not been performed in a classical cascade-style
but in parallel. 

\subsection{Overview}
\label{sec:overview}

  The organization of rest of the thesis is presented in this section.
After a state of the art summary in chapter \ref{cap:state-of-art}, chapter
\ref{cap:application} describes the relaxation algorithm and its
application to NLP. Chapters \ref{cap:experiments} and \ref{cap:results} present
experiments performed and results obtained. Finally, chapter \ref{cap:conclusions}
contains conclusions yield by this work and outlines further research lines.

\subsubsection{Chapter \ref{cap:state-of-art}: Disambiguation and Optimization in NLP}

  In this chapter we overview the current trends in natural language
processing, specially on corpus linguistics. A special attention is 
paid to disambiguation tasks, since it is the main issue in this thesis.
\medskip

  Items are addressed from the artificial intelligence perspective
rather than from a linguistic point of view. Nevertheless, the 
indispensable contribution and complement that linguistics 
must provide to the presented work,
as well as the utility our contribution may represent
to those linguists working on large corpora, is also taken into account.
\medskip

  We also summarize in this chapter previous applications of different 
optimization algorithms to perform NLP tasks, and describe some of
the most representatives.

\subsubsection{Chapter \ref{cap:application}: Application of Relaxation Labelling to NLP}

   In this chapter we detail the relaxation labelling algorithm and its
possible parameterizations. We discuss which ones may be appropriate to our
purposes and some related problems.
\medskip

   Since the presented algorithm is based on modelling language by means of
context constraints, and the developing of a linguist--written model is 
highly costly, we also describe different ways to acquire the knowledge --in the
form of context constraints-- to be used by the algorithm.

\subsubsection{Chapter \ref{cap:experiments}: Experiments and Results}
   
  In this chapter we describe three groups of performed experiments which
were performed on several corpora and disambiguation tasks, 
using different parameterizations and knowledge obtained from various sources.
\medskip

    The first set of experiments aimed to establish the most appropriate
parameterization for the relaxation algorithm when applied to NLP 
disambiguation tasks. POS tagging was used as a testbench task for this
purpose.

    The second group of experiments aimed to perform POS tagging as
accurately as possible using relaxation labelling. Different language
models were used in this case to test the ability of the algorithm to
integrate constraints obtained from various knowledge sources.

    The last set of experiments consisted of broadening the range of
application to NLP tasks other than POS tagging. It also included
experiments on combining different word features and on simultaneous 
resolution of several NLP tasks. The selected tasks were shallow 
parsing and word sense disambiguation.

\subsubsection{Chapter \ref{cap:results}: Comparative Analysis of Results}

   This chapter contains a comparative analysis of the 
results obtained by relaxation labelling on the tested tasks.

  We focused on analyzing the influence of the use of multi-source and/or 
multi-feature information on the obtained results, as well as
studying whether parallel task solving yields some improvement.

  Also, the results produced by our system are compared to those of other 
current systems. Some considerations on the evaluation and 
comparison of systems performing NL corpus processing are
exposed.

\subsubsection{Chapter \ref{cap:conclusions}: Conclusions}

   In this chapter we summarize the research described in this thesis and
we outline some future lines of research to improve the performance of
our system and to broaden the range of optimization algorithms applied to NLP.


%% file: state-of-art.tex
\chapter{Disambiguation and Optimization in NLP}
\label{cap:state-of-art}

 In the previous chapter, we outlined the situation of disambiguation problems 
inside the NLP field, and more particularly in the corpus linguistics field, where
the objective is processing large amounts of linguistic data with reasonable
results, rather than obtaining very precise results over a small set of
linguistic features.

 In this chapter we will expose in a more detailed way which are the current trends
to approach these problems, and to what extent are optimization methods spread inside
the NLP research field.
\medskip

 First, a general look on usual methods to perform NLP disambiguation tasks is presented.
Then, we will describe how those methods are used in particular when facing the two most 
common disambiguation problems in this field: 
part-of-speech tagging and word sense disambiguation. 

 Second, an overview of how different energy--function optimization techniques have
been applied to NLP task is presented, and the most representative are described.

\section{The Disambiguation Problem}
\label{sec:disamb}

  Natural Language is an ambiguous mean to transmit information. This may be a 
desirable feature for joke-tellers, cartoonists or humor screenplay writers,
but it becomes a great problem when one wants a computer to process information
stored in this form.

  This makes ambiguity to be one of the main problems of NLP, and very probably the 
only one, since all NLP problems can be related to some kind of ambiguity.

  Ambiguity in natural language is manifold. We find part-of-speech ambiguity
(e.g. past vs. participle in regular verbs), semantic ambiguity in polysemic words,
syntactic ambiguity in parsing (e.g. PP-attachment), reference ambiguity
in anaphora resolution, etc. 
\medskip

   Methods to deal with ambiguity range from the brute-force: {\em compute all the 
possibilities and choose the best one}, which becomes impractical when dealing
with real data such as linguistic corpora; to more clever language representations
which avoid the combinatory explosion by taking into account the frequency, probability,
or any other criterion to select the best solution.
\medskip

  To enable a computer system to process natural language,
it is required that language is {\em modelled} in some way, that is, that 
the phenomena occurring in language are characterized and captured, in such
a way that it can be used to predict or recognize future uses of language:
\cite{Rosenfeld94} defines language modelling as {\em the attempt to characterize,
capture and exploit regularities in natural language}, and states that 
the need of language modelling arises from the great deal of variability
and uncertainty present in natural language.

  Different methods to process NL derive from different approaches to 
language modelling. These methods can be classified into three broad families, 
although, obviously, there exist also methods that would fit in more 
than one --an thus they may be considered as hybrid-- or that do not fit 
well in any category.
\medskip

  First, the linguistic or knowledge--based family, where language is modelled 
by a linguist who tries to explain the behaviour of 
ambiguity using some unambiguous formalism.
Some representative examples of this class are the works by 
\cite{Voutilainen94,Oflazer97}, where a large hand-written constraint 
grammar is used to perform part-of-speech tagging.
\medskip

  Second, we find the statistical family, where the language model is 
left to a data-collection process which stores thousands of occurrences 
of some kind of linguistic phenomenon and tries to derive a statistical 
law from them. This model acquisition is known as {\em training}.
Two main points of view are used in this family, the Bayesian point of
view and the Information Theory point of view. Both of them rely on 
the estimation of occurrence probabilities for each relevant event, but
while the former tries to obtain them computing the number of event
occurrences (Maximum Likelihood Estimation) --which may cause problems 
when an event is infrequent or data are scarce-- the later is based on 
assuming maximum ignorance and trying to minimize the model entropy, thus
unobserved events will only keep maximum uncertainty.

  Statistical methods constitute a very large family, and the one that has 
reported most successful results to NLP field in recent years. Some examples 
are the works by \cite{Rabiner90} who presents a tutorial on Hidden Markov 
Models and their application to speech recognition,
or \cite{Kupiec91,Briscoe94b} who apply statistical methods to
grammar development and parsing.
Other works on NLP using statistical models are that of \cite{Matsukawa93a,McKeown93},
who learn to cluster similar words, and \cite{Brants97} who identify the 
grammatical function of each word in a sentence.
Statistical methods have been specially successful --since the 1970s
to nowadays-- when applied to speech recognition tasks \cite{Rabiner90,Huang93,Heeman97}.
This success caused that they were also used in other NLP areas, such as
optical character recognition, spelling correction, POS tagging, parsing,
translation, lexicography, text compression and information retrieval.
\medskip

  Finally, the machine-learning family, where the model acquisition 
is also automatic, but the knowledge acquired belongs to a higher level 
than simple occurrence frequencies. For instance, \cite{Yarowsky94} learns decision
lists to properly restore accents in Spanish and French texts, 
the system described \cite{Daelemans96b} learns morphological rules,
and as a secondary effect, a relevant classification of phonemes appears, 
and in \cite {Mooney96}, several classical machine--learning algorithms are 
applied to learn to disambiguate word senses, and the results of
the different methods compared.
\medskip

   The following section summarizes some basic issues on linguistic 
corpora compilation and overviews some well-known corpora.
   Sections \ref{sec:tagging-POS} and \ref{sec:tagging-semantic}
summarize the application of methods belonging to the families above
to the particular disambiguation problems of POS tagging and WSD.

\subsection{Corpuses, corpi and corpora}

  The aforementioned success of statistical methods in natural
language processing would not have been possible without the 
existence of large amounts of machine-readable text from which
statistical data could be collected. A compilation of 
naturally occurring linguistic phenomena in newspapers, 
literature, parliament acts, etc. is known as a {\em linguistic corpus}.

  The compilation of raw text corpora is no longer a problem, since
nowadays most documents, books and publications are written on a
computer. But corpus have a higher linguistic value when they are
{\em annotated}, that is, they contain not merely words, but also
linguistic information on them (part-of-speech, syntax analysis, etc.).
\medskip

  Although some corpus compilation efforts were started in the 1960s,
corpus linguistics has reached it highest popularity in recent years,
mainly due to the success of statistical methods as well as to the 
increase in computational and storing capacity of computer systems.
\medskip

  When a corpus compilation project is started, some important issues must be 
taken into account.

  First, whether the corpus should be balanced or not. This is an open question
that has not found a definitive answer in years. As stated in \cite{Church93}, 
it comes down to a tradeoff between quantity and quality: While American 
industrial laboratories (e.g. IBM, ATT\&T) tend to favour quantity, the BNC, NERC,
and many dictionary publishers --specially in Europe-- tend to favour quality.
\cite{Biber93} claims for quality, since poor sampling methods or inappropriate
assumptions can produce misleading results.

  Second, which annotations will be included in the corpus, and
how will be the annotation task performed. Automatic annotation introduces
a certain amount of errors in the corpus, while manual annotation is 
very expensive in terms of human resources. Some research aiming to reduce the
human effort when annotating training corpus is presented in \cite{Engelson96}. It 
consists of algorithms which select the most informative samples that should be 
annotated to be later used in training. The same idea is present in the
work by \cite{Lehmann96}, who developed a database containing positive and negative 
examples of different linguistic phenomena, so that a test or training corpus
focused on a certain phenomena can be build at a low cost. See \cite{Atkins92}
for further information on corpus design and development.
\medskip

  The most well-known corpora are probably the Brown Corpus (BC) and the 
London-Oslo-Bergen corpus (LOB). The BC \cite{Francis82} contains over 
a million words of American English and was tagged in 1979 
using the TAGGIT tagger \cite{Greene71} plus hand post--edition. The LOB
corpus contains the same amount of British English and was also tagged
in 1979.

  Nowadays, corpora tend to be much larger, and are compiled mainly through
projects and initiatives such as the Linguistic Data Consortium (LDC), the 
Consortium for Lexical Research (CLR), the Electronic Dictionary Research (EDR),
the European Corpus Initiative (ECI) or the ACL's Data Collection Initiative (ACL/DCI).

  Those associations provide corpora as the Wall Street Journal (WSJ, $300$ million 
words of American English), the Hansard Corpus (bilingual corpus containing $6$ years
of Canadian Parliament sessions), the Lancaster Spoken English Corpus (SEC), the
Longman/Lancaster English Language Corpus, the Nijmegen TOSCA corpus,
the $200$-million-word Bank of English corpus (BoE) --tagged using the ENGCG 
environment \cite{Jarvinen94}--, or the $100$-million-word British National 
Corpus (BNC) tagged with the CLAWS tagger \cite{Leech94}. Surveys on existing
resources can be found in \cite{Edwards93,Wilks96a}.
\medskip

  Although most corpora limit their annotation level to part-of-speech
tags, some offer higher level annotations and constitute an important 
source of knowledge for those researching in NLP.
We find, for instance, syntactically analyzed corpora such as the Susanne 
corpus, the Penn Treebank ($3$ million words) \cite{Marcus93}
or the IBM/Lancaster treebank. Also, 
SemCor \cite{Miller93} contains over $200,000$ words of the Penn Treebank 
semantically disambiguated with WordNet synsets. A review of the state of
art in using parsed corpora can be found in \cite{Souter94}.
\medskip

  Until a few years ago, the existing corpora were all of the English language.
Nevertheless, the success and applicability of corpus in linguistics as well
as in NLP, has raised a wide interest and caused its quick extension to 
other languages. For instance, the Tr\'esor de la Lange Fran\c{c}aise (TLF) which 
contains $150$ million words of written French, the LEXESP corpus \cite{Acebo94,Cervell95}
that will contain over $5$ millions of balanced text in Spanish, or the CTILC, which
compiles over $50$ million words of modern Catalan. A good information source on
Spanish lexical resources is \cite{IC96}.

\subsection{Part-of-speech Tagging}
\label{sec:tagging-POS}

  POS tagging consists of assigning to each word of a sentence a
part-of-speech tag which indicates the function of that word in that
specific context. Although it depends on how fine--grained is the used
tagset --which may vary from 20 to 500 tags--,
it can be considered an easy task, since many words --between 80\% and 90\%-- 
either have only one possible part-of-speech, or the context in which they
appear restricts the choice to only one tag. But in the remaining percentage
of cases, the ambiguity solution may be very difficult to find, many times
requiring semantic or even common sense knowledge.

\subsubsection{Some considerations on tagger evaluation and comparison}
  When evaluating the performance of any system, one must be
very prudent, since a higher accuracy percentage does not necessarily mean
a better tagging system. Thus, comparing taggers is not as straightforward
as it might seem.

  The factors that affect most the accuracy of a tagger are the tagset, 
and the way in which unknown words are handled. If the tagger uses a
statistical model, the noise in the train and test corpus also
plays a role in distorting the computation of the real tagger 
performance. This issue is further discussed in section \ref{sec:results-POS-errors}.
\medskip

\noindent\underline{\bf The Tagset}
\nopagebreak\medskip

\noindent With respect to the tagset, the main feature that concerns us is
its granularity --which is directly related with its size--. 

If the tagset is too coarse, the tagger accuracy will be much higher,
since only important distinctions are considered and thus the task
to perform is much easier, but the results would supply an excessively poor
information. 

If the tagset is too fine--grained, the tagger precision will
be much lower, because the model will have to be much richer
and so, more difficult to obtain and more likely to 
contain flaws\footnote{If the model is build manually, flaws will 
                       be caused by humans, who are error-prone 
                       proportionally to the complexity of the task.
                       If it is build statistically, huge amounts 
                       of data are required to correctly estimate the model.}.
In addition some very fine distinctions may not be solved on 
syntactic or context information only, but need semantic or even pragmatic
knowledge.
\medskip

  Some samples of commonly used tagsets can be found in \cite{Kren97}, who
classify the word level tags --such as POS tags-- in two classes, 
according to the number of linguistic dimensions they specify:

\begin{itemize}
\item Single--dimension tags, which will usually contain the syntactic category of
the word, such as {\bf N}, {\bf V}, {\bf ADJ}, {\bf DET}, etc. ({\em noun}, {\em verb},
{\em adjective}, {\em determiner}, etc.).
\item Multiple--dimension tags, which incorporate additional word features
such as gender, number, person, etc. For instance, the tag {\bf VIP3S} could indicate
that a word form is {\em verb, indicative, present, third person, singular}.
\item Combination of separate multiple dimensions in sets or {\em readings}.
As in Constraint Grammars formalism, a word would have a set of labels, each one containing
information on a single linguistic feature. For instance, {\bf ($<$SVO$>$ V PRES -SG3 VFIN)}
states that a word is {\em transitive, verb, present, non-third singular, finite}.
This representation has the following advantages: it can be graded, that is, one can choose
which features is interested in and ignore the others, and it also enables the 
introduction of new dimensions, as for instance syntactic roles or semantic information. 
\end{itemize}

  Some studies on the tagset size influence on a tagger results have been done. 
For instance,
\cite{Sanchez95} proposed a 479-tag tagset for using the Xerox tagger on Spanish, 
and later reduced it to 174 tags since the first proposal was considered 
too fine--grained for a probabilistic tagger.
\cite{Elworthy94a} states that the tagset sizes (48 tags for Penn Treebank 
and 134 for LOB corpus) do not affect greatly to the behaviour patterns of the 
re-estimation algorithms. 
The work in \cite{Briscoe94a} is also related with this topic,
since POS experiments on different languages (English, Dutch, French 
and Spanish), each with different corpus and tagset were tested and compared.
\medskip

\noindent\underline{\bf Handling Unknown Words}
\nopagebreak\medskip

\noindent Another factor that can affect tagger accuracy is how are unknown words
handled. The most usual methods are:
\begin{itemize}
\item Do not consider the possibility of unknown words. That is, assume a 
morphological analyzer which gives an analysis for any unknown word. 
This is usually simulated by analyzing all the words appearing in the 
used test corpus. Obviously this approach will tend to produce higher
performance results, though it is in fact less robust than the following.
\item Assume that unknown words may potentially take any tag --excluding
those tags corresponding to closed categories (preposition, determiner, \ldots), 
which are considered to be all known--. Although this is more realistic than
the previous method, it introduces more noise, and so the reported 
performance will be lower.
\item Use available information to guess which are the candidate tags
for a given unknown word. This is the most robust and powerful solution,
and has been applied in different ways by several researchers. 
For instance, \cite{Meteer91,Weischedel93} take into account inflectional and derivational
endings as well as capitalization and hyphenation to guess the possible
POS tags for a word, while \cite{Adams93} use
a statistical model of fixed--length suffixes combined with capitalization
features to guess the possibilities for unknown words.
\cite{Ren92} perform a frequency study of the cases when a word is 
{\em actually} unknown or when it is a typewriting error, and
a thorough subclassification of each case is exposed.
 Machine learning techniques are also used to deal with unknown words:
\cite{Mikheev96a,Mikheev96b} learns morphological rules from a 
lexicon and a corpus using unsupervised statistical acquisition. These 
rules can later be applied to guess the possible tags for an unknown word.
\cite{Daelemans96a} uses example based learning to identify the possible 
categories for unknown words, and \cite{Marquez98} apply a decision tree
learning algorithm to acquire trees that can be later used to establish
the categories of words not found in the lexicon. 
\end{itemize}

\subsubsection{Current methods for POS tagging}

  The existing NLP literature describes many methods and algorithms to
reduce as much as possible the small percentage of cases in which the
POS tag for a word has several possibilities, and even in those cases,
to choose the most likely one. 
These methods can be classified in the three broad groups described 
at the beginning of section \ref{sec:disamb}:
linguistic, statistical and machine--learning family. 
See \cite{Abney96} for a clear survey on kinds of POS tagging techniques.
\medskip

\noindent\underline{\bf The Linguistic Approach}
\nopagebreak\medskip

\noindent The linguistic approach consists of coding the necessary knowledge in a 
set of rules written by a linguist after introspection. Early systems
performed rather bad for nowadays standards (below $80\%$ accuracy), like 
the pioneer TAGGIT \cite{Greene71} which was used to create the initial
tagging of the Brown Corpus, which was then hand revised.
Later came the work by the Nijmegen TOSCA group \cite{Oostdijk91}, and
more recently the development of Constraint Grammars \cite{Karlsson95}
and their application to POS tagging \cite{Voutilainen94},
which can be considered the best existing tagger ($99.3\%$ accuracy is reported,
though not all words are fully disambiguated).
Constraint Grammars have also been used to morphologically disambiguate 
agglutinative languages as Basque \cite{Aduriz95} or Turkish \cite{Oflazer97}.

Although the linguistic approach produces high quality language models that 
yield good disambiguation results, it is a high time-consuming one since 
many years of human resources are required to develop a good language
model. 
\medskip

\noindent\underline{\bf The Statistical Approach}
\nopagebreak\medskip

\noindent Another trend that seems to be the most extended at present --since it
requires much less human effort-- is the statistical approach:
A statistical model of language is used to disambiguate the word sequence.
The simplest model would be a most-likely-tag choice for each word.  
A successful model during the last years has been modelling
the sequence of tags in a sentence as a Hidden Markov Model, 
and computing the most probable tag sequence given the word sequence.
An accurate overview on the subject can be found in \cite{Merialdo94}.
\medskip

  To obtain a statistical language model, one needs to estimate the model 
parameters, such as the 
probability that a certain word appears with a certain tag, or the probability
that a tag is followed by another. This estimation is usually done by
computing unigram, bigram or trigram frequencies on tagged corpora. 
The CLAWS system \cite{Garside87}, which was
the probabilistic version of TAGGIT, used bigram information and was improved
in \cite{DeRose88} by using dynamic programming. The tagger by \cite{Church88}
used the Brown corpus as a training set to build a trigram model.

  The corpora from which frequencies are estimated should be 
disambiguated by hand, in order to produce an accurate
estimation. Although this requires also a big deal of human work, it is much less
than in the previous approach and it is currently becoming less important, since
many tagged corpora are available at a low or even zero cost. Although these corpora
still contain tagging errors, they are a good enough starting point, and they
are revised and improved in new releases. 
\medskip

 To reduce the amount of hand tagged corpora needed to obtain such estimations,
the Baum-Welch re-estimation algorithm \cite{Baum72} was used to improve an initial
bigram model --obtained from a small tagged corpus, or even, invented by 
introspection-- iterating over untagged data. A famous example is the Xerox tagger
described in \cite{Cutting92}, which has been improved and adapted by a number
of researchers. For instance, \cite{Sanchez95} transported it to Spanish and
enlarged it with an unknown words handler.
The Baum-Welch algorithm has been also used in \cite{Briscoe94a}, who experimented 
the utility of the algorithm on refining models for
languages different than English and in \cite{Elworthy94a} where a thorough study of 
the conditions in which it is worth using the algorithm is presented.
\medskip

  Recent works \cite{Jung96,Ng96,Saul97} try to enlarge the range of the algorithms, 
that is, not to limit them to a fixed-order n--gram, but to be able to combine 
different order n--grams, statistical information on word morphology, 
long--distance n-grams \cite{Huang93} or triggering pairs \cite{Rosenfeld94}.
\medskip

  Other works that use a statistical-based approach are \cite{Schmid94a} which 
performs energy-function optimization using neural nets and \cite{Ludwig96} 
who disambiguates words on a morphological information basis --for very flexive 
languages where this is possible--. 
\medskip

  Results produced by statistical taggers are really good, giving about $95\%-97\%$
of correctly tagged words.
Some authors try to improve the results by using a set of context constraints 
which are applied to the results of the probabilistic tagger, and correct its 
most common errors. \cite{Brill92,Brill95,Roche95,Aone96} use a simple most-likely 
tag tagger the output of which is corrected by a set of transformations automatically
acquired by an error-driven algorithm. \cite{Moreno94} uses a bigram statistical
tagger whose output is corrected by a set of linguist--written constraints.
\medskip

  There are also hybrid methods that use both knowledge based and statistical 
resources, such as that of \cite{Tzoukermann95} or the research presented 
in this thesis \cite{Padro96a,Voutilainen97,Marquez97a}.
 Comparative discussion on the advantages and disadvantages of linguistic and 
statistical based part-of-speech taggers can be found in \cite{Chanod95,Samuelson97}.
\medskip

  The main flaw of statistical taggers is the difficulty to accurately 
estimate the language model. Since the estimation is usually performed
through {\em Maximum Likelihood Estimate} --that is, the probability 
assigned to each event is proportional to the number of times it occurred
in the training data--, and since MLE does not waste any probability mass for 
events not appearing in the training corpus, the estimation may be more or less 
accurate when the model has a reduced number of parameters --e.g. a bigram model--,
but it turns very inaccurate when the number of parameters grows, since the
necessary amount of training data becomes too large. Then, the main problem encountered
is the low or zero frequency events. The main techniques employed to deal with 
these problems are sketched below. Further details can be found in 
\cite{Jelinek89,Charniak93,Manning96}.
\medskip

\noindent\underline{\it Dealing with insufficient data}
\nopagebreak\medskip

\noindent The low or zero frequency events 
produce inaccurate estimations for the probability of events that happen 
scarcely in the training set, for instance, if event {\bf A} is observed 
to happen once and event {\bf B} to happen twice, the estimated probability 
would be double for {\bf B} than for {\bf A}, when this is not necessarily 
true.

  The zero--frequency events problem is even worse, since zero probability is
assigned to events not observed in the training corpus, when they are not necessarily
{\em impossible} to happen.

  Techniques to deal with scarce events may consist either of 
re-arranging the probability mass in order to keep a part of it for
unobserved events, or of combining information from different sources, 
since only one source is not reliable. 
The most usual techniques are smoothing, backing-off and Maximum Entropy modelling, 
which include methods belonging to both kinds.

\paragraph{Smoothing.}
Smoothing can be done through count re-estimation 
methods such as Add--One --also known as Laplace's law (1775)-- or Good--Turing 
estimation \cite{Good53}, or either by relying on lower--order data, that is,
through linear interpolation (also called deleted interpolation).
\medskip

 Count re-estimation methods try to correct the false estimations of 
rare events by re-arranging the frequency countings before the estimation.

Add-One adds one to all frequencies, thus avoiding zeroes and reducing the
proportion between rare happening events. Lidstone's law is a variation of
Laplace's which adds not one but some smaller positive value $\lambda$.

 Good--Turing redistributes the
amount of observations to favour those events with less observations. Usually
this redistribution is either smoothed or performed only on low--frequency events, 
because it produces unreliable results for high--frequency events. \cite{Church91}
presents a comparison of Add-One and Good-Turing techniques.

  Other methods are those proposed by \cite{Ney93} who present two alternative 
models for discounting frequencies, in order to distribute them among unseen events,
and by \cite{Schmid94b} who estimates n-gram probabilities using decision trees.
\medskip

 Smoothing through linear interpolation \cite{Bahl83,Bahl89} is performed by computing the 
probability of an event as the weighted average of the estimated probabilities for 
its sub--events. For instance, the smoothed probability of a trigram 
could be computed as the weighted average of the estimated probability for the 
trigram itself, and for the corresponding bigram and unigram, that is,
\begin{list}{}{}
\item $P_s(x_n|x_{n-1},x_{n-2}) = \lambda_1 P_1(x_n) + \lambda_2 P_2(x_n|x_{n-1}) + 
\lambda_3 P_3(x_n|x_{n-1},x_{n-2})$, 
\end{list}
where the optimal values for $\lambda_i$ are usually computed with the 
Estimation Maximization (EM) algorithm \cite{Dempster77}. 
\medskip

\paragraph{Backing-off.} 
It is also possible to combine information from different sources in the
style of \cite{Katz87}. This is called {\em back--off}, and consists 
of using the MLE estimation if the event has appeared at least $k$ times. 
Otherwise, the probability of the lower order event is used. For instance, if 
a trigram has occurred less than $k$ times, the corresponding bigram probability 
would be used, provided the bigram has appeared $k$ or more times. If it has not, 
the unigram probability would be used. 
While linear interpolation consists of combining several sources giving
a different weight to each one, backing--off chooses the best one among the 
available information sources. It can be seen as a particular case
of linear interpolation where the $\lambda_i$ are all zero but the one
corresponding to the higher order history that has more than $k$ observations, 
which is set to $1$.

\paragraph{Maximum Entropy Estimate.} 
A recent approach which solves the scarce 
data problem is Maximum Entropy Estimate, which on the contrary than MLE, assume
maximum ignorance (i.e. uniform distribution, maximum entropy) and
observed events tend to lower the model entropy. Under this approach,
unobserved events do not have zero probability, but the maximum they can
given the observations. That is, the model does not assume anything that 
has not been specified.
\medskip

  In classical MLE approaches each knowledge source was
used separately to build a model, and those models were then
combined. Under the Maximum Entropy approach, the model is
build already combined, and attempts to capture the information
provided by each knowledge source. 
\medskip

  Each information source is seen as defining a {\em constraint} 
on the model stating that the {\em average} combined probability
for an event equals its {\em desired expectation}, 
usually computed from the training data.

  That is, we have a set of constraints on the probabilities of
each event much weaker than those that would have been obtained
by MLE, which would have asserted that the probability of an
event must equal its {\em desired expectation}, not in average,
but always.
\medskip

  Once the constraints are established, the Generalized Iterative
Scaling algorithm (GIS) is used to compute the values for the event
probabilities that satisfy all constraints, that is, to obtain a 
combined probabilistic model.
If the constraints are consistent, an unique solution --i.e. an unique
probability distribution-- is guaranteed to
exist and the GIS algorithm is proven to converge to it \cite{Darroch72}.
\medskip

  Summarizing, the Maximum Entropy Principle \cite{Jaines57,Kullback59}
can be stated as follows:
\begin{enumerate}
\item Formulate the different information sources as constraints that 
      must be satisfied by the target combined estimate.
\item Among all probability distributions that satisfy the constraints,
      choose the one with highest entropy.
\end{enumerate}
\medskip

   The advantages of the Maximum entropy approach over MLE are the following:
\begin{itemize}
\item The MLE models provided by different information sources are usually inconsistent,
     the reconciliation needed to combine them is achieved by averaging 
     their answers (linear interpolation) or by choosing one of them (back-off). 
     Maximum Entropy approach eliminates the inconsistency because it imposes 
     weaker conditions on each information source.
\item ME is simple and intuitive. It assumes nothing but the
     constituent constraints.
\item The ME approach is very general. Probability for any event can be
     computed, and many kinds of constraints can be incorporated,
     such as long distance dependencies, or complicated correlations.
\item The information in already existing statistical language models can 
     be absorbed into the ME estimate.
\item The GIS algorithm is incrementally adaptive, that is, new constraints can be added
     at any time. Old constraints can be maintained or allowed to relax.
\item An unique ME solution is guaranteed to exist, and the GIS algorithm
      to converge to it.
\end{itemize}

   The main drawbacks of this approach are:
\begin{itemize}
\item The GIS algorithm is computationally expensive.
\item There is no theoretical bound to GIS algorithm convergence rate.
\item If inconsistent constraints are used the existence, uniqueness 
      and convergence theorems may not hold.
\end{itemize}
  
  As a summary, we can say that ME approach avoids the problems that raise
from the low-frequency events when using MLE, and that it builds a model
which correctly combines information provided by different knowledge sources.
This issue is closely related to the work presented in this thesis, since 
it also describes a method to combine different sources of knowledge.

  For further details on the ME approach, see \cite{Lau93,Rosenfeld94,Ristad97a}. 
\medskip

\noindent\underline{\bf The Machine--Learning Approach}
\nopagebreak\medskip

\noindent The third family is represented by authors who use
learning algorithms which acquire a language model from a training corpus,
in such a way that
the learned model includes more sophisticated information than a n-gram
model: For instance, \cite{Marquez97a,Marquez97b} learn statistical 
decision trees from a
tagged corpora. A similar idea is that of \cite{Daelemans96a} who use an 
example--based learning 
technique and a distance measure to decide which of the previously
learned examples is more similar to the word to be tagged. 
The same idea is used in \cite{Matsukawa93b}, but the learned examples
are used there to correct the most frequent errors made by a
Japanese word segmentator.
\cite{Samuelson96} acquires Constraint Grammars from tagged corpora taking 
into account the tags that appear between pairs of tags which never occur 
consecutive in training corpora. 
The above referenced \cite{Brill92,Brill95} can also be 
considered as belonging to this group, since the algorithm automatically learns 
the series of transformations which best repair the most common errors made by 
a most--likely--tag tagger. A variant of his method is used by \cite{Oflazer96}, who
present a hybrid system which combines hand--written Constraint Grammars with
automatically acquired Brill--like error--driven constraints.

\subsection{Semantic Tagging: Word Sense Disambiguation}
\label{sec:tagging-semantic}

  Word sense disambiguation (or word sense selection) consists of, 
given a sentence, assigning to each content word a sense label 
indicating which is the right meaning for the word in that context.
\medskip

  The above definition for the WSD task leaves open the question of how and what 
should sense labels be. The problem here is similar to that
of tagset granularity on POS tagging, since we can select as sense labels
a very coarse division, such as a {\em topic} or {\em area} identifier,
or a very fine--grained division such as a pointer to a sense entry 
in a Machine Readable Dictionary (MRD) or in a word taxonomy.

  If we choose a coarse division, the disambiguation task would be
easier, but some slight sense distinctions will be lost. For instance,
if we choose that that word {\em host} has three possible senses:
{\em $<$person$>$}, {\em $<$life-form$>$} and {\em $<$horde$>$}, we will
not be able to distinguish between the {\em $<$master-of-ceremonies$>$} and
the {\em $<$innkeeper$>$} senses, which will both be subsumed under the 
{\em $<$person$>$} label. On the other hand, if we choose
very fine--grained sense labels --as usual MRD entries are--, some 
ambiguities will be unsolvable, as for instance, the difference between 
the senses {\em $<$interior-designer$>$} and {\em $<$ornamentalist$>$} 
for word {\em decorator}.

  Semantic labels sets can range from a few dozens to thousands of tags.
For instance, there would be $11$ different possible semantic categories 
for nouns and $573$ for verbs if the sense labels were chosen to be WordNet top 
synsets\footnote{WordNet \cite{Miller91} is a concept hierarchy, where 
                 each sense is represented by a set of synonym words 
                 (a {\em synset}). In addition, synsets are grouped in 
                 thematic files, each one with its own {\em file code}.},
and $26$ for nouns and $15$ for verbs if the chosen labels were WN file codes. 
The Roget's International Thesaurus \cite{Chapman77} distinguishes $1,042$ 
thematic categories. Finally, if we chose as sense labels the WordNet synset codes, 
there would be $60,557$ possible noun semantic classes and $11,363$ for verbs.
\medskip

  The variability in the sense granularity is an issue that makes it very
difficult to compare the accuracy of different sense disambiguation systems,
but there are other factors which make the performance reported by different
systems to vary greatly. Those factors include in the first place the kind of 
knowledge used and the source from which it is obtained --it would seem logical 
that the performance of a system using statistically knowledge acquired in an
unsupervised way was much lower than that of system based on hand--coded
semantic knowledge--. Second, the amount of context considered by the 
disambiguation technique used to apply that knowledge --no context at all, 
local context, full context, \ldots--.
And third, how is the system evaluation performed --over all words, over words
in a certain syntactical category (e.g. nouns), over a chosen subset of words, \ldots--.
Some steps have been done \cite{Gale92b,Miller94} towards establishing a common 
baseline for enabling WSD systems comparison.
\medskip

  A broad classification of the currently existing systems considered from
the point of view of the kind of knowledge they use are the linguistic
or knowledge--based, the statistical, and the hybrid families.
\medskip

\noindent\underline{\bf The Knowledge--Based Approach}
\nopagebreak\medskip

 Methods in the first family are those which rely on linguistic knowledge,
which is usually obtained through lexicographer introspection \cite{Hirst87}.

 This knowledge may take the form of a Machine Readable Dictionary (MRD),
as in the case of 
\cite{Lesk86}, who proposes a method for guessing the right sense in a given context
by counting word overlaps between dictionary definitions, or \cite{Cowie92}, who
use the same idea but avoiding the combinatory explosion by using
simulated annealing. Dictionary definitions are also used by \cite{Guthrie91} to
collect lists of salient words for each subject semantic code of
words in LDOCE\footnote{LDOCE stands for {\em Longman's Dictionary Of Contemporary English}.}.
The co-occurrence data acquired in this way were later used by \cite{Wilks93} 
to construct context word vectors for each word and for each sense.
\cite{Harley94,Harley97} present a multi-tagger, which combines different 
information sources (POS, domain, collocations, \ldots) contained in the 
completely coded {\em Cambridge International Dictionary of English} (CIDE),
to assign to each word an unique entry in the dictionary, and thus disambiguating
it at several levels (POS, sense, lemma, \ldots).

Nevertheless, since dictionaries --even when
they are machine readable-- are intended for human users, they contain loosely 
structured knowledge which often relies on common sense. Thus, if we want a 
computer system to use that knowledge, it is necessary to extract the 
knowledge contained in the MRD and put it in a more tightly structured format.
Works relating to this kind of knowledge extraction are described 
in \cite{Dolan93,Wilks93,Rigau97b}.
\medskip

 Another possibility is the use of knowledge not from a human--oriented source
such as a MRD, but in the form of a thesaurus or a conceptual taxonomy
such as WordNet \cite{Miller91}.
For instance, \cite{Cucchiarelli97} use a thesaurus obtained by selecting from WordNet
a subset of domain--appropriate categories that reduce WordNet overambiguity.
The work presented in \cite{Atserias97} uses different unsupervised lexical methods
--which handle sources including monolingual and bilingual dictionaries-- to link
each sense in a language other than English to an unique WordNet synset, in order to
enable the automatic construction of multilingual WordNets.

  The taxonomy can be used directly, as a lexical source, or else taking 
advantage of the lexical relationships encoded in the hierarchy. 
\cite{Sussna93} measures the {\em conceptual distance} between senses to improve 
precision during document indexing, assuming that co-occurring words will 
tend to have close senses in the taxonomy. The idea is extended to the notion
of {\em conceptual density} by \cite{Agirre95,Agirre96}, who instead of
minimizing pairwise sense distance, try to maximize the density of
the senses for all words in the sentence. \cite{Rigau94} presents a 
methodology to enrich dictionary senses with semantic tags extracted 
from WN, using a conceptual distance measure.
\medskip

\noindent\underline{\bf The Statistical Approach}
\nopagebreak\medskip

The second broad group uses knowledge obtained from statistical
processing of corpora either tagged (supervised training) or
untagged (unsupervised training). Most of the systems rely on
unsupervised training, since semantically annotated corpora are generally
less available than corpora with other kinds of annotation. 

  The collected statistics can be lexical statistics --such as
mutual information, relative entropy, or merely frequencies of 
words and senses--, or lexical distributions, i.e. computing and 
comparing distribution of senses respect to a context --generally
word forms--.
\medskip

  Among the unsupervised techniques, we find the work by \cite{Brown91}, who
extracted a statistical model from the bilingual Hansard Corpus, and by
\cite{Yarowsky92}, who collects
word classes co-occurrences from unsupervised corpus, under the assumption
that {\em the signal overcomes the noise}.
Although \cite{Schutze92} uses unannotated data for training, his model
acquisition procedure is not completely unsupervised: After 
the context vector based automatic generation of clusters from corpus
co-occurrence data, a manual post--process to assign each sense to a 
cluster is performed. In \cite{Schutze95} the idea is extended with the
use of second--order co-occurrences, context vectors are automatically 
clustered in classes representing word senses, and word occurrences are 
disambiguated by assigning them to their closest cluster.

  On the side of the supervised methods, \cite{Gale92a,Gale93} --following the
idea of \cite{Dagan91} which states that {\em two languages are better than one}-- 
use the bilingual Hansard Corpus and consider the
French translation of a word as a semantic tag, assuming that different 
senses will correspond to different French words, thus the Hansard Corpus
can be seen as semantically disambiguated. Obviously this does not 
hold for all words, so experiments are limited to some specific words.
\medskip

   Another important feature of statistical based systems is the amount of 
context they consider when acquiring or applying the statistical model.
From this point of view, we find the whole range of possibilities, from
no context at all to considering all the document as relevant context for
each word.

  Some methods use no context at all, such as \cite{Gale92b,Miller94}
who describe two baseline benchmarks based on no context information (guessing
and most likely) and one based on very local co-occurrence information. 

  Methods which rely on local context information are those which consider
only the words in a small window ($5$ to $10$ words) or in the same sentence
than the focus word. The underlying idea in this approach is stated in 
\cite{Yarowsky93} as the {\em one sense per collocation} principle:
{\em the \underline{same} words are likely to have the \underline{same} 
meanings if they occur in \underline{similar} local contexts}.
Some authors using this idea are \cite{Bruce94a,Bruce94b},
who decompose the probabilistic model that would result of taking several
local context features (morphological, collocation, POS, \ldots) as interdependent,
and \cite{Pedersen97b} who compare three statistical language model acquisition 
algorithms, using either local or global context features.
\cite{Lin97} uses also local features, but converting Yarowsky's {\em one
sense per collocation} principle to a more flexible version: 
{\em \underline{different} words are likely to have \underline{similar} meanings 
if they occur in \underline{identical} local contexts}. This adaptation
enables disambiguating the sense of a word, even though one has not collected
its typical contexts, by using the contexts of similar senses.

  Finally, some methods rely on global context information, which 
corresponds to the {\em one sense per discourse} principle \cite{Gale92a}.
For instance, \cite{Yarowsky92,Gale93} compute the salient words vector for
each class on a global context basis.
\cite{Yarowsky95}, who relies in both {\em one sense per collocation} and
{\em one sense per discourse} principles, uses an unsupervised 
incremental algorithm to classify occurrences of a given word in one
of its possible classes. The algorithm consists of a cycling corpus--based 
procedure which collects local context features (basically salient words 
lists) which can later be used for WSD.
\medskip

  A comparison between different statistical methods can be found in \cite{Leacock95}:
Bayesian, neural networks, and content vectors are compared at performing word
sense disambiguation.
\medskip

\noindent\underline{\bf Hybrid Approaches}
\nopagebreak\medskip

  The last group includes those methods which mix statistical
and linguistic knowledge. The current trend is to combine one or
more lexical knowledge sources --either structured or non-structured--
such as corpora, MRDs, Lexical Knowledge Bases, thesauri, taxonomies, etc.,
with exploitation techniques which usually consist of different similarity or 
distance measures between lexical units.
\medskip

  For instance, \cite{Liddy92} use LDOCE subject semantic codes and the 
WSJ corpus for computing a subject-code correlation matrix which is later
used for word sense disambiguation. \cite{Karov96} describe a system which
learns from a corpus
a set of typical usages for each word sense, using as training contexts 
those of the words appearing in the sense definition in an MRD.
Newly appearing occurrences are compared with the training data 
using a similarity function.

  Although the learning algorithm described in \cite{Yarowsky95} is of 
statistical nature, he points out that it is useful using MRD definitions to
collect the seed words needed to start the iterative acquisition procedure.
\medskip

  There is also the approach of \cite{Ribas95}, who uses WordNet as a 
lexical resource, combined with an
association ratio based algorithm, to automatically extract selectional
restrictions from corpora, which are then used to disambiguate the noun
senses that are heads of verb complement phrases.
\cite{Resnik93,Resnik94,Richardson94,Resnik95} present a method for automatic
WSD based on an information content measure. The similarity between two
classes is computed as the information content of their lowest common hyperonym
in WordNet hierarchy. The information content of a class is proportional 
to its occurrence probability, which is estimated from a corpus.

  The work by \cite{Peh97} presents the combination the mapping of a 
domain-specific hierarchy onto WordNet with semantic distance metrics to get
a wide--coverage method for disambiguating semantic classes.

  A multi-resource combination system is that of \cite{Rigau97a}, who combine 
several heuristics --most of them statistical, but knowledge based lexical resources 
such as WordNet are also used-- in a weighting approach to disambiguate word senses. 
The used techniques and lexical resources range from naive most--likely sense 
assignation to content vector representations built from MRDs, through 
different similarity measures.
\medskip

 Other methods that may be considered hybrid are those that combine
more or less sophisticated lexical resources with machine learning
algorithms, to automatically derive a language model oriented to
WSD. Samples of this approach are the work by \cite{Siegel97}, who
uses machine learning algorithms to acquire a model capable of classifying 
verbs as {\em state} or {\em event}, or by
\cite{Mooney96}, who compares seven classical learning algorithms
(including neural nets, statistical techniques and decision-trees) at the
task of disambiguating among six senses for the word {\em line}, using
local context information. \cite{Ng96} presents an example--based system
which acquires a model that integrates different knowledge sources,
including POS tags, morphology, word co-occurrences, and verb--object
syntactic relationships.

\section{Optimization Techniques in NLP}
\label{sec:optimization}

  In this section we will overview the optimization techniques
most commonly used in Artificial Intelligence, and summarize how
have they been applied to natural language processing.
\medskip

  We understand by {\em optimization} any technique that leads to maximize/minimize 
an --either explicit or implicit-- objective function. We can find gradient step
or mathematical programming in any classical Operational Research course, or approaches
as neural nets or genetic algorithms in more recent works. See
\cite{Gu94} for a review on different optimization algorithms.
\medskip

  Although optimization techniques have not been applied to NLP in a generalized way,
we can find several uses in the literature, which had represented a great success
in the field, such as those of \cite{DeRose88} who optimized the speed of 
\cite{Garside87} tagger by means of dynamic programming
--which is more or less the same that the well--known Viterbi algorithm 
\cite{Viterbi67} does--, the use of simulated annealing to disambiguate word senses 
in \cite{Cowie92}, the neural net POS tagger in \cite{Schmid94a}, or the 
paraphrasing algorithm in \cite{Dras97}.

  We find a more extended use of optimization algorithms for model estimation.
For instance, the well known Baum-Welch algorithm \cite{Baum72} used in 
\cite{Kupiec92,Elworthy94a}, or the Expectation Maximization (EM) algorithm 
\cite{Dempster77} commonly used to perform linear interpolation smoothing, or
as in the case of \cite{Pedersen97b}, to disambiguate word senses.
\medskip

  The recent Maximum Entropy approach can also be considered as 
using optimization methods, since the GIS algorithm used to 
select the most appropriate probability distribution, is actually
a maximization algorithm to pick the maximum entropy model.

\subsection{Neural Nets}
\label{sec:neural-nets}

   Neural nets are models that rely on the interaction 
between a large number of simple computing units (neurons)
connected to other units in the net. When a neuron is 
{\em active}, it causes a neighbour cell to become active
provided that the neuron activity level is high enough, 
and that the link that connects them has enough {\em weight} 
or {\em strength}.
  
   Neural nets were originally developed to model human
brain physiology, and soon were found to have interesting
computing capabilities. Neural nets are energy-function 
optimizers that can be trained to learn a task consisting of
producing a certain output when supplied a certain input.
When properly trained, neural nets have generalization
abilities, that is, they are able to generate the right 
output when faced to a never seen input.

   Knowledge is stored in neural nets in the form of
link weights. When an input is presented, the produced
output depends on how is this input propagated through the net,
which is obviously a function of the link weights. Thus,
training a neural net consists of computing the right
weight for each link. This is usually done through an iterative
error minimization algorithm, such as the well known {\em backpropagation}
algorithm.

  Those interested in neural nets, can find further introductory information
to the field in the books by \cite{McLelland84,Kosko90}.
\medskip

  Due to their learning abilities, and to the success
obtained in other fields, neural nets have been applied
to NLP by several authors.
  The most widely used are feed--forward nets. But since they
can process only fixed-length input,
recurrent neural nets \cite{Elman88} --which do not present
this restriction-- are more commonly used in NLP.

  Some general reviews on this area can be found in 
\cite{Reilly92,Miikkulainen93,Feldman93}. Some sample systems
are those of \cite{Schmid94a}, who performs POS tagging using a
feed--forward net, or \cite{Wermter95} which describes a
symbolic-connectionist hybrid system.
\cite{Lawrence95} perform grammatical inference --in fact,
the net learns to distinguish grammatical sentences, although
no grammar is inferred--, and
\cite{Collier96} uses Hopfield networks to store and recall 
patterns of natural language sentences.

\subsection{Genetic Algorithms}
\label{sec:genetic-alg}

  Genetic algorithms are also energy function optimizers.
They are based on the idea that {\em evolution} and
{\em natural selection} produce solutions which are optimally 
adapted to the environment. 

  One starts
with a population of random solutions --or almost random 
to save convergence time--. Solutions are coded as a sequence of 
features or {\em genes}, all possible values of which should
be present in the starting population.
The solutions are combined in pairs (or any order reproduction 
groups) to create new solutions that will have features (genes)
from both (or all) of their parents.
Only the best solutions (the {\em fittest} ones) are allowed
to survive and procreate. The fitness of a solution is
evaluated through a {\em fitness function}.
This kind of {\em natural selection}
leads to an improvement of the solution population generation
after generation, until it reaches an optimum. Mutation can
also be included as small random changes in descendence genes 
to avoid local optima.

  For further details on Genetic Algorithms techniques and their
applications see the books by \cite{Holland92,Goldberg89}
\medskip

   Genetic algorithms have also been used in NLP, though to a much
minor extent than neural nets. For instance, 
\cite{Smith95} used genetic algorithms to perform grammatical 
inference from a set of sample sentences.

\subsection{Simulated Annealing}
\label{sec:sim-annealing}

  Simulated annealing is an optimization algorithm which
is based on metal annealing processes seen from the point
of view of statistical mechanics. 

  The process starts with 
a high temperature, which causes the current state to be
unstable, and very likely to change. The state is changed 
always in the maximum gain direction, but the temperature
component can make it change in a more random way. As the
temperature decreases and the solution approaches
the optimum, the random component is less and less important.

  Simulated annealing obeys the Boltzmann distribution which
has been proven to lead to a global optimum if the temperature
decrease is slow enough. Further details on its relationship with 
relaxation processes and Boltzmann machines can be found in \cite{Aarts87}.
In \cite{Kirkpatrick83} one can find more about the optimization 
properties of simulated annealing.
\medskip

  The work by \cite{Cowie92,Wilks97} describes the application of 
simulated annealing to perform WSD. Nevertheless, they use
as compatibility constraints only the dictionary definition overlap
for possible senses. Simulated annealing is in fact --as 
described in chapter \ref{cap:application}-- a particular case of 
discrete relaxation labelling, thus, more complex compatibility
constraints --linguistically motivated, statistically acquired,
multi-feature, etc.-- could be used with that algorithm.

\subsection{Relaxation Labelling}
\label{sec:relaxation}

  Relaxation is a generic name for a family of
iterative algorithms which perform function optimization, based on
local information. They are closely related to neural nets \cite{Torras89}
and gradient step \cite{Larrosa95b}.

  Although relaxation operations had been long used in 
engineering fields to solve systems of equations \cite{Southwell40},
they did not get their biggest success until their extension to symbolic domain
--relaxation labelling-- was applied to constraint propagation field, 
specially in low-level vision problems \cite{Waltz75,Rosenfeld76}.
\medskip

  From our point of view, relaxation labelling is a technique that can be
used to solve consistent labelling problems (CLPs) --see \cite{Larrosa95a}--.
A consistent labelling problem consists of, given a set of
variables, assigning to each variable a value compatible with the values of 
the other ones, satisfying --to the maximum possible extent-- a set of 
compatibility constraints. Algorithms to solve consistent labelling problems
and their complexity are studied in \cite{Nudel83}.

  In the Artificial Intelligence field, relaxation has been mainly
used in computer vision  --since it is where it was first used-- to address
problems such as corner and edge recognition or
line and image smoothing \cite{Lloyd83,Richards81}. Nevertheless, many
traditional AI problems can be stated as a labelling problem: the 
travelling salesman problem, n-queens, or any other combinatorial problem
\cite{Aarts87}. 
\medskip

  The utility of the algorithm to perform NLP tasks was pointed out
in \cite{Pelillo94a,Pelillo94b}, where POS tagging was used as a
toy problem to test their methods to improve the computation of
constraint compatibility coefficients for relaxation processes. 
Nevertheless, the first application to a real NLP problems, on
unrestricted text is the work presented in this thesis, and 
published in \cite{Padro96a,Padro96b,Marquez97a,Voutilainen97}, which
in addition enables the use of multi--feature constraints coming
from different sources.


%% file: application.tex
\chapter{Application of Relaxation Labelling to NLP}
\label{cap:application}

 This chapter discusses the use of the relaxation labelling algorithm to perform NLP
tasks. To enable the application of relaxation labelling, the language 
model must be described in terms of algorithm elements --variables, 
labels, constraints, etc.--. In our case, the words
in the sentence to disambiguate will be represented as variables, 
and the possible values for certain linguistic features (POS, sense, etc.) will
correspond to their labels.
\medskip

  Although --as pointed out in chapter \ref{cap:introduction}-- relaxation labelling
has been mainly used in fields other than NLP (engineering, computer vision, \ldots),
some researchers in optimization techniques \cite{Pelillo94a,Pelillo94b} have
used POS tagging as a toy problem to experiment their methods to improve the
performance of relaxation labelling. They used a 1000-word test
corpus, and only binary constraints, which was enough to their
purposes of testing a method for estimating constraint compatibility values.
In our case, the aim is POS tagging itself, so we will have to use more 
sophisticated information and larger corpora. 
\medskip

  We will describe the relaxation labelling algorithm from a general point of view 
in section \ref{sec:relax-alg}. Afterwards, in section \ref{sec:parameterization}
we will explain which ones among the described parameterizations were selected 
as the most suitable for our purposes, and discuss some problems related to the 
convergence of the algorithm. Finally, in section \ref{sec:constraints},
we will consider different ways to obtain the constraints needed to 
feed the algorithm.

\section{Algorithm Description}
\label{sec:relax-alg}

  In this section the relaxation algorithm is described from a general 
point of view. Its application to NLP tasks will be discussed in section
\ref{sec:parameterization}.
\medskip

  Let $V=\{v_1,v_2,\ldots,v_N\}$ be a set of variables.

  Let $T_i=\{t^i_1,t^i_2,\ldots,t^i_{m_i}\}$ be the set of possible
labels for variable $v_i$ (where $m_i$ is the number of different labels
that are possible for $v_i$).

  Let $C$ be a set of constraints between the labels of the variables.
Each constraint is a ``compatibility value'' for a combination of
pairs variable--label. For instance, the constraint
$$
0.53 \;\;\;\; [(v_1,A)(v_3,B)]
$$
states that the combination of variable $v_1$ having label $A$, and
variable $v_3$ having label $B$ has a compatibility value of $0.53$.
Constraints can be of any order, so we can define the compatibility
value for combinations of any number of variables (obviously we can 
have combinations of at most $N$ variables).
\medskip

   The aim of the algorithm is to find a {\em weighted labelling} such that
{\em global consistency} is maximized.

A {\em weighted labelling} is a weight assignation for each possible 
label of each variable:
\begin{list}{}{}
\item $P=(p^1,p^2,\ldots,p^N)$ where each $p^i$ is a vector containing a
weight for each possible label of $v_i$, that is: $p^i=(p^i_1,p^i_2,\ldots,p^i_{m_i})$
\end{list}

 Since relaxation is an 
iterative process, when the time step is relevant, we will note
the weight for label $j$ of variable $i$ at time $n$ as $p^i_j(n)$. 
When the time step is not relevant, we will note it as $p^i_j$.

Maximizing {\em global consistency} is defined as maximizing for each variable $v_i$,
$(1\leq i\leq N)$, the average support for that variable, which is defined as
the weighted sum of the support received by each of its possible labels, that is:
$$
{\displaystyle \sum_{j=1}^{m_i} p^i_j \times S_{ij}}
$$
where $p^i_j$ is the weight 
for label $j$ of variable $v_i$ and
$S_{ij}$ is the support received by that pair from the context. The support
for the pair variable--label expresses {\em how compatible} that
pair is with the labels of neighbouring variables, according to
the constraint set (see section \ref{sec:support}). 

  The performed {\em global consistency} maximization is a vector optimization. It does not 
maximize --as one might think-- the sum of the supports of all variables. It finds
a weighted labelling such that any other choice would not increase the 
support for {\em any} variable given --of course-- that such a labelling exists.
If such a labelling does not exist, the algorithm will end in a local
maximum.
\medskip

  The relaxation algorithm consists of:
\begin{itemize}
\item start in a random labelling $P_0$.
\item for each variable, compute the ``support'' that each label
    receives from the current weights for the labels of the other 
    variables (i.e. see how compatible is the current weighting with 
    the current weightings of the other variables,
    given the set of constraints).
\item Update the weight of each variable label according to the 
    support obtained by each of them (that is, increase
    weight for labels with high support, and decrease weight for those
    with low support).
\item iterate the process until a convergence criterion is met.
\end{itemize}

  The support computing and weight changing must be performed in
parallel, to avoid that changing a weight for a label would affect
the support computation of the others.
\medskip

  We could summarize this algorithm saying that at each time step, a 
variable changes its label weights depending on
how compatible is that label with the labels of the other variables at
that time step. If the constraints are consistent, this process converges 
to a state where each variable has weight $1$ for one of its labels and
weight $0$ for all the others.
\medskip

  Note that the {\em global consistency} idea --defined as the 
maximization of the average support received by each variable
from the context-- makes the algorithm robust, 
since the problem of having mutually
incompatible constraints (so one can not find a combination of label
assignations which satisfies all the constraints) is solved because 
relaxation does not (necessarily) find an exclusive combination of
labels, that is, an unique label for each variable, but a weight
for each possible label such that consistency is maximized (the
constraints are satisfied to the maximum possible degree).  
\medskip

  Advantages of the algorithm are:
\begin{itemize}
\item Its highly local character (each
      variable can compute its new label weights given only the
      state at previous time step). This makes the algorithm highly 
      parallelizable (we could have a
      processor to compute the new label weights for each variable, or even a
      processor to compute the weight for each label of each variable).
\item Its expressiveness, since we state the problem in terms of
      constraints between variable labels.
\item Its flexibility, we do not have to check absolute consistency of constraints.
\item Its robustness, since it can give an answer to problems
      without an exact solution (incompatible constraints,
      insufficient data, \ldots)
\item Its ability to find locally optimal solutions to NP problems in a
      non-exponential time (Only if we have an upper bound for the
      number of iterations, i.e. convergence is fast or the
      algorithm is stopped after a fixed number of iterations).
\end{itemize}
\medskip

   Drawbacks of the algorithm are:
\begin{itemize}
\item Its cost. Being $N$ the number of variables, $v$ the average number
      of possible labels per variable, $c$ the average number of constraints
      per label, and $I$ the average number of iterations until convergence,
      the average cost is  $N\times v\times c\times I$, that is, it depends
      linearly on $N$, but for a problem with many labels and 
      constraints, or if convergence is not quickly achieved, the multiplying 
      terms might be much bigger than $N$.
\item Since it acts as an approximation of gradient step algorithms,
      it has their typical convergence problems: Found optima are local, and
      convergence is not guaranteed, since the chosen step might be
      too large for the function to optimize. 
\item In general, constraints must be written manually, since they are
      the modelling of the problem. This is good for easy-to-model domains
      or reduced constraint-set problems, but in the case of POS tagging or
      WSD constraint are too many and too complicated to be written by hand.
\item The difficulty to state which is the {\em compatibility value} for
      each constraint. If we deal with combinatorial problems with
      an exact solution (e.g. travelling salesman), the constraints will be
      all fully compatible (e.g. stating that it is possible to go to any
      city from any other) or fully incompatible (e.g. stating that it
      is not possible to be twice in the same city). But if we try to model
      more sophisticated or less exact problems (such as POS tagging) things
      will not be black or white. We will have to assign a compatibility
      value to each constraint.
\item The difficulty to choose the support and updating functions more
     suitable for each particular problem.
\end{itemize}

\subsection{Support Function}
\label{sec:support}

  The relaxation labelling algorithm requires a way to compute which is 
the support for a variable label given the constraints and the current 
label weights for the other variables.
 This is called the {\em support function} and it is the heart of the 
algorithm, since it is closely related to what will be maximized.
\medskip

  To define the support received by a variable label from its context, we have
to combine the individual influences of each constraint that can be applied
for that pair in the current context.
  So, we will define $Inf(r,i,j)$ as the influence of
a constraint $r$ on label $j$ for variable $i$. Its formal definition
requires some previous steps:
\medskip

{\em DEF: Constraint.}
  A constraint $r$ consists of a compatibility value $C_r$ and its 
associated set of pairs variable--label. The compatibility values can
be restricted to a certain interval (e.g. $[0,1]$, $[-1,1]$, $[0,+\infty]$ \ldots), or
not restricted at all.

 A constraint expresses a how compatible is a given combination of variable labels.
It can be written as follows:
$$
C_r \;\;\;\; [(v_{i_1},t^{i_1}_{j_1}),\ldots,(v_{i_{n_r}},t^{i_{n_r}}_{j_{n_r}})]
$$
\begin{tabbing}
\` where $1 \leq i_1,\ldots,i_{n_r} \leq N$ and \\
\` $1 \leq j_k \leq m_{i_k}$ for $k = 1\ldots n_r$
\end{tabbing}

where $n_r$ is the constraint {\em degree}, that is, the number of pairs variable--label
it involves, and $(v_{i_1},t^{i_1}_{j_1}),\ldots,(v_{i_{n_r}},t^{i_{n_r}}_{j_{n_r}})$ 
are the pairs involved in the constraint.

  For simplicity we will note label $j$ for variable $i$ as $t_j$ instead of 
$t^i_j$, since the variable $i$ which the label is applied to is already 
present in the pair.
The previous constraint will then be expressed as:
$$
C_r \;\;\;\; [(v_{i_1},t_{j_1}),\ldots,(v_{i_{n_r}},t_{j_{n_r}})]
$$ 
\medskip

{\em DEF: Context weight}.
 Obviously, the influence of a constraint on a given variable label is zero 
if the constraint does not include the pair variable--label. (i.e. that constraint
is not applied).
 Then,  constraints that have an influence on a given pair $(v_i,t_j)$ are only
those that include that pair, i.e., those of the form:
$$
C_r \;\;\;\; [(v_{i_1},t_{j_1}),\ldots,(v_i,t_j),\ldots,(v_{i_{n_r}},t_{j_{n_r}})]
$$ 

  We define the {\em context weight} for a constraint and a pair variable--label 
$W(r,i,j)$ as the product of the current weights for the labels appearing 
in the constraint except $(v_i,t_j)$, or, if preferred,
as though the weight for that label was $1$. 
  
  The {\em context weight} states {\em how applicable} the constraint is
given the current context of $(v_i,t_j)$.
  The {\em constraint compatibility value} $C_r$ 
states {\em how compatible} the pair is with the context. 

 Being $p^s_q(n)$ the weight assigned to label $t_q$ for variable 
$v_s$ at time $n$, the context weight is:
$$ 
W(r,i,j) = p^{i_1}_{j_1}(n) \times \ldots \times p^{i_{n_r}}_{j_{n_r}}(n)
$$
\begin{tabbing}
\` where $p^i_j(n)$  is not included in the product.
\end{tabbing}
\medskip

{\em DEF: Constraint Influence.}
 Once we have defined the constraint compatibility values and the
context weight, we can define the influence of a constraint on the 
pair $(v_i,t_j)$ as:
$$ 
Inf(r,i,j) = C_r \times W(r,i,j)
$$
\medskip

{\em DEF: Support.}
  Once we have computed the influence for each constraint on the
given label of a variable, we can compute the total support received 
by that label combining the influences of all constraints.

 Several support functions are used in the literature, depending on the
problem addressed, to define the support $S_{ij}$ received by label $j$ of 
variable $i$. Different support functions correspond to different ways of
combining constraint influences. See \cite{Kittler86} for further details 
on different possible support functions.

\begin{itemize}
\item The first formula computes the support for a label adding the influences obtained 
from each constraints. Depending on the nature of the compatibility values, 
support values may be {\em negative} indicating {\em incompatibility}. This point is
discussed in section \ref{sec:compatibility}.
\begin{equation}
\label{support-s}
 S_{ij} = \sum_{r}{Inf(r,i,j)}
\end{equation}

\item  Another possible formula is adding the influences of constraints which 
involve exactly the same variables and multiplying the results afterwards.
\begin{equation}
\label{support-p}
S_{ij} = \prod_{G \in {\cal P}(V)} \sum_{r \in G}{Inf(r,i,j)}
\end{equation}
\begin{tabbing}
\` where ${\cal P}(V)$ is the set of all possible subsets of $V$ (the set of variables).
\end{tabbing}

\item And finally, we can also pick the maximum of the influences of constraints which 
involve the same variables and multiply the results afterwards.
\begin{equation}
\label{support-m}
S_{ij} = \prod_{G \in {\cal P}(V)} \max_{r \in G}{Inf(r,i,j)}
\end{equation}
\end{itemize}

\subsection{Updating Function}
\label{sec:updating}

   The algorithm also needs to compute which is the new weight for a variable
label, and this computation must be done in such a way that it can
be proven to meet a certain convergence criterion, at least under 
appropriate conditions\footnote{Convergence has been proven under
certain conditions, but in a complex application such as POS tagging
we will find cases where it is not necessarily achieved.} \cite{Zucker78,Zucker81,Hummel83}.

 This is called the {\em updating function} and it is used to compute and normalize the
new weights for each possible label. 

 Several formulas have been proposed \cite{Rosenfeld76}, and some of them have 
been proven to be approximations of a gradient step algorithm. 

 The updating formulas must increase the weight associated with labels
with a higher support, and decrease those of labels with lower
support. This is achieved by multiplying the current weight of a label by a
factor depending on the support received by that label. Normalization
is performed in order that the weights for all the labels of a
variable add up to one.
\medskip

   Although {\em ad-hoc} updating functions can be used, as in \cite{Deng96},
the most commonly used formulas are the following:

\begin{itemize}
\item This formula increases the weight for a label when
$S_{ij}$ is positive and decreases it when $S_{ij}$ is negative.
Values for  $S_{ij}$ must be in  $[-1,1]$.
\begin{equation}
\label{updating-c}
p^i_j(n+1) = \frac{p^i_j(n) \times(1+S_{ij})}{\displaystyle \sum_{k=1}^{m_i}{p^i_k(n)\times(1+S_{ik})}}
\end{equation}

\item This formula increases the weight when  $S_{ij}>1$ and decreases
it when  $S_{ij}<1$. Values for  $S_{ij}$ must be in $[-1,+\infty]$.
\begin{equation}
\label{updating-p}
p^i_j(n+1) = \frac{p^i_j(n) \times
S_{ij}}{\displaystyle \sum_{k=1}^{m_i}{p^i_k(n)\times S_{ik}}}
\end{equation}

Since the support values $S_{ij}$ are computed using the constraint compatibility 
values $C_r$, which may be unbounded, they do not necessarily belong to the
intervals required by any of the above updating functions. 
Even in the case that the $C_r$ were bounded, if the support 
computation used was additive (\ref{support-s} or \ref{support-p}), the final
support result would not be guaranteed to be in the required interval.

 Thus, it will be necessary to normalize the final support value for each
label, in order to fit in the appropriate interval. This issue is further 
discussed in section \ref{sec:convergence}.

\item The following formula is also used as an updating function:
\begin{equation}
\label{updating-d}
p^i_j(n+1) = \frac{e^{S_{ij}/T}}{\displaystyle \sum_{k=1}^{m_i}{e^{S_{ij}/T}}}
\end{equation}

where $T$ is a temperature parameter which decreases at each time
step. The labelling is non-ambiguous in this case (weights are only
$0$ or $1$) and what we compute is the probability that a variable
changes its label. When $T$ is high, changes occur randomly. As $T$
decreases, support values get more influence. 
\end{itemize}

  If variables take only one label at each time step (that is, one
label has weight $1$, and the others $0$) and updating
function \ref{updating-d} is used, the
procedure is called stochastic
relaxation (which is equivalent to simulated annealing),
while if label weights are not discrete, and the updating
function is \ref{updating-c} or \ref{updating-p}
we talk about continuous deterministic relaxation.
See \cite{Kittler85,Torras89} for clear expositions of what is relaxation labelling
and what kinds of relaxation can we get by combining different support
and updating functions.

\section{Algorithm Parameterization}
\label{sec:parameterization}
  As described in section \ref{sec:relax-alg}, relaxation labelling handles variables,
labels and compatibility constraints. Since we want to use it to perform NLP disambiguation
tasks, we have first to model language in a suitable way for the algorithm.
\medskip

  The most direct way is to model each word as a variable, and each of its possible readings
--either POS-tags, senses, syntactic roles, etc.-- as a possible label for that variable.
In this framework, the constraints will express compatibility between one reading 
for one word 
and another reading for a word in its context. So we will have constraints stating that 
a determiner is very compatible with a noun to its right, but rather not compatible with 
a verb.

These constraints will state how compatible is, say, label $t_1$ for variable
$v_i$ with label $t_2$ for variable $v_j$ (or any other combination of $n$ pairs 
variable--label). These constraints, their complexity, and the kind of information 
they use (morphological, syntactic, semantic, \ldots) will depend on the task we are 
performing. Different ways to obtain them are described in section \ref{sec:constraints}.
\medskip

  Once we have modelled language in terms of variables, labels and constraints, we have
to choose the most suitable parameterizations for the algorithm. 
\cite{Kittler86} study how the choice of the adequate support and
compatibility functions should depend on the contextual information
to be exploited. Experiments described in
\cite{Padro96a} and reported in section \ref{sec:selection}
were used to determine which are the most useful parameters. The obtained 
results are outlined in the following sections.

\subsection{Support Function}
\label{sec:app-support}

  Intuitively, the most suitable support function for NLP tasks seems to be
the additive function described in equation \ref{support-s}, since it does not
multiply constraint influences. 

  The multiplicative combination of influences may produce undesirable effects
when dealing with NLP tasks, since the absence of information (influence zero)
would cause the final result drop.
\medskip

   In natural language models, it will be very unlikely that we have 
{\em all}  imaginable constraints --e.g. all combinations of trigram values--. 
This means that when a constraint is missing the influence will be either zero 
or a tiny value (if we performed some kind of smoothing). 
This is obviously a drawback for multiplicative combination functions, 
since a lack of information such as a missing
constraint, does not necessarily imply support zero for a label. It seems more
natural to add the influences, so when one is missing, it just does not 
contribute at all.
\medskip

   Experiments performed on POS tagging, and reported in \cite{Padro96a,Padro96b}, confirm
that idea, showing that support function \ref{support-s} achieves better results than 
the multiplicative functions \ref{support-p} and \ref{support-m}.

\subsection{Updating Function}
\label{sec:app-updating}

   The choice of the appropriate updating function is tightly bound to the 
compatibility values
nature described in section \ref{sec:compatibility}. If compatibility values are allowed
to be negative --to enable constraints expressing {\em incompatibility}-- then
the support computed for each label may be negative. This will force us to use
updating function \ref{updating-c}, since it can take supports in $[-1,1]$.
If our compatibilities are always positive, we can then use updating function 
\ref{updating-p} or \ref{updating-d}.
\medskip

 In any case, normalization of supports must be performed to ensure they are 
in the appropriate interval to be used by the updating function. 

 Although the support normalization function
could be considered another algorithm parameter --one could choose straight
linear normalization or use some sigmoid-shaped function such the arc-tangent or
the hyperbolic tangent--, performed experiments (see \cite{Padro95}) show
that there is not a significant difference between the different normalization
functions. This made us choose the simplest normalization --linear-- and leave
as a parameter its application domain interval\footnote{Support values falling out of 
                                               the normalization function domain
                                               interval will be mapped to the 
                                               highest possible support value.}.
Finding the right normalization interval still requires further studies.
As discussed in section \ref{sec:convergence}, it seems to depend on the 
average support per label, although the easiest way to obtain it is using
a part of the training corpus as a {\em tuning} set.
\medskip

   Since performed experiments pointed out that computing compatibility values 
as mutual information produces better results than the other tested formulas, 
and this measure can be negative, we will use updating function \ref{updating-c}.

\subsection{Compatibility Values}
\label{sec:compatibility}
  
   The constraints used in relaxation labelling must state a compatibility value
  for each combination of pairs variable--label. These values may be as simple
  as $-1$ for not-compatible and $+1$ for compatible, but the algorithm will 
  perform much better if the constraints are better informed. 

   The compatibility value for a constraint states how compatible is one pair
  word--label $(v,t)$ with a set of pairs in its context. We can either 
  assign them by hand or 
  try to use some probability or information theory measure to estimate them.
\medskip

  That is, we have a constraint of the form:
$$
C_r \;\;\;\; [(v_{i_1},t_{j_1}),\ldots,(v,t),\ldots,(v_{i_{n_r}},t_{j_{n_r}})]
$$ 
and we want to compute its $C_r$. Since $C_r$ must express how compatible is the 
pair $(v,t)$ with the context expressed by the constraint, the possible
ways of computing $C_r$ will have to take into account the event consisting
of an occurrence of the pair $(v,t)$, and the event consisting of an
occurrence of the context $[(v_{i_1},t_{j_1}),\ldots,(v_{i_{n_r}},t_{j_{n_r}})]$,
and see if there is any correlation between them.

That is, being $H$ the event corresponding to an occurrence of the pair $(v,t)$
and $E$ the event of an occurrence of the context described by the constraint,
we can consider that $C_r$ can be computed as a function of those
two events: $C_r = Comp(H,E)$. 
\medskip

The $Comp$ function can take many forms. the most direct one is the 
conditional probability:
$$Comp(H,E) = P(H|E)$$
\smallskip

    Another possibility is to compute Mutual Information 
between the two events $E$ and $H$ \cite{Church90,Cover91}.
$$Comp(H,E) = \log\frac{P(H\cap E)}{P(H)\times P(E)}$$
\smallskip

  or either the Association Score \cite{Resnik93,Ribas94} which combines the 
previous two
$$Comp(H,E) = P(H|E)\times\log\frac{P(H\cap E)}{P(H)\times P(E)}$$
\smallskip

  other possibilities are Relative Entropy \cite{Cover91,Ribas94}
$$Comp(H,E) = \sum_{X\in\{H,\neg H\},Y\in\{E,\neg E\}}P(X\cap Y)\times\log\frac{P(X\cap Y)}{P(X)\times P(Y)}$$
\smallskip

 or statistical correlation
$$Comp(H,E) = \frac{P(H\cap E)-P(E)P(H)}{\sqrt{(P(E)-P(E)^2)(P(H)-P(H)^2)}}$$
\medskip

 Yet another possibility is using Maximum Entropy Estimate --which was introduced 
in section \ref{sec:tagging-POS} (see \cite{Rosenfeld94,Ristad97a} for details)-- 
to compute those compatibility values. Although it has not been used 
in this research, we plan to introduce it in the short run.

\subsection{Convergence and Stopping Criteria}
\label{sec:convergence}

  Relaxation labelling is an iterative algorithm which has been proven to
converge under certain conditions \cite{Zucker78,Zucker81,Hummel83}.
These conditions often require simple models --e.g. consisting only on
binary constraints which must be symmetric-- which are not likely
to hold in complex applications such as those of NLP.

  In addition, relaxation algorithms are
often stopped before convergence, since they either produce better results at
early iterations \cite{Richards81,Lloyd83} or it is not necessary to
wait until convergence to know what the result will be \cite{Zucker81}.
  Different stopping criteria can be found in the literature, although
most of them have a strong {\em ad-hoc} flavour \cite{Eklundh78,Peleg79}.
\cite{Haralick83} presents a conditional probability interpretation of
relaxation labelling which enables a theoretically grounded stopping criterion,
unfortunately, it is only applicable in specific cases (binary constraints only, with 
bounded weight sum for all constraints affecting the same variable).
\medskip

  In our case, many experiments seem to produce slightly better
results --though not statistically significantly better--
in early iterations than at convergence (see section \ref{sec:selection}). 
\medskip

   In order to identify the causes of this phenomenon and find a
better stopping criterion, we took as a reference the aforementioned 
uses of relaxation that were stopped by
different criteria than convergence and tested several stopping criteria 
based on the amount of variation from one iteration to the next.
  
   The tested criteria were: average weight variation in one iteration step,
maximum weight variation in one iteration step, Euclidean distance and 
average Euclidean distance per word (\cite{Eklundh78}). Those measures
tried to capture the distance between the points in weight space obtained in
two successive iterations. 

   Additional tested criteria were the first derivatives of the above 
mentioned distance measures. That is, the variation that each one 
presented from one iteration to the next. Those measures intended to
capture the speed of variation that relaxation presents at different 
evolution stages, in order to find out whether it kept any relationship
with the optimal stopping iteration.

   Unfortunately, none of them led to any significant result, that is, 
no relationship was found between the proposed measures and the optimal
stopping iteration.
\medskip

  Another hypothesis that could explain the best performance at early
iterations was that the noise contained in the training and testing corpora 
caused the algorithm to mistag some words and/or to compute as
errors correct taggings that were mistagged in the test corpora.

  To check to what extent this could be true, we manually analyzed
the errors introduced/corrected  by the algorithm between the optimal stopping
iteration and convergence. Results showed that most of them were due to 
noise in the model and in the test corpus and that convergence is,
if not improving the accuracy, at least not decreasing it.
\medskip
  
  The third approach to finding a suitable stopping criterion was related
to the convergence speed. As in the case of gradient step, relaxation labelling
can converge faster or slower if an appropriate step size is chosen and
it is conveniently decreased. In the case of relaxation algorithms, this
effect is achieved by modifying the normalization factor of the support 
values.

  As described in section \ref{sec:support}, the support values are
usually computed as the sum of several compatibility values ($C_r$), 
thus, the support value $S_{ij}$ for a label is unbounded.
It has also been described in section \ref{sec:updating} that the
global support for a label must be bounded in $[-1,1]$ or in $[0,+\infty]$.
This means that once the global support for a label has been computed, it
must be scaled to fit in the appropriate interval. If this normalization 
yields a large value, the step
taken by the iteration will be large, while if normalization produces a 
relatively small value, the step will be shorter. 

  Our experiments show that changing the normalization factor for the 
support values has the effect of changing the number of necessary iterations
to achieve convergence. In addition, the 
tagging accuracy is also affected: There seems to be an optimal
normalization factor which produces the best accuracy at convergence.
For this optimal value, the difference between the accuracy at convergence
and at the optimal stopping iteration is non-significant. This provides us
with a reasonable stopping criterion: we can wait for convergence, provided 
we use a good normalization factor. To establish the most appropriate
value, we test a range of possible value and choose the value
that produces highest accuracy on tagging a fresh part of the training corpus, 
called {\em tuning set}.

  Although higher accuracy results seem to be obtained at early iterations with
low normalization factors, the difference is either small or non-significant. 
In addition, the difficulties 
to select the right stopping iteration described above, point that the selected
stopping criterion is a reasonable one. Nevertheless, this issue will require
further attention.
\medskip

   More details and results of the experiments on the stopping criterion for the relaxation
algorithm can be found in section \ref{sec:selection-stopping}

\section{Constraint Acquisition}
\label{sec:constraints}

  Although the relaxation labelling algorithm and its application to POS tagging 
described so far maximize the consistency of the tag assignation to each
word, the accuracy of the result is obviously dependent on the quality 
of the used language model.

  To enable the use of the relaxation algorithm, the language model must
be written in the form of constraints. The better the constraint model
describes language, the better the obtained results will be.
\medskip

  In this section, we will describe the different techniques that were 
used to obtain the constraints necessary to feed the relaxation algorithm. 
The described techniques range from the manual writing of constraints by
a linguist to the use of machine learning techniques to acquire them, 
through statistical n-gram model acquisition.
  
  The used constraints cover different NLP phenomena. They were developed
to perform different NLP tasks, as described in chapter \ref{cap:experiments}.
For POS tagging, n-gram models, automatically learned models as well as a few 
hand--written constraints were used. For shallow parsing, we used n-gram models and
a quite good linguist written model. Finally, for word sense disambiguation
POS tag n-grams were combined with other kinds of knowledge ranging
from simple co-occurrence statistics to machine--learned selectional
restrictions (see section \ref{sec:experiments-WSD} for details).

\subsection{Manual Development}
\label{sec:manuals}

  The most obvious way to get a language model is getting a 
linguist who, through introspection, writes a set of constraints
which are supposed to describe the behaviour of language.

  This approach is the most scientific one, since it is based on
the assumption that to understand, predict or simulate any
phenomenon, one has to model it first in an unambiguous way.

  Unfortunately, while this is achievable in physical sciences, 
it appears to be much harder when dealing with cognitive sciences or,
as in our case, with language. This difficulty to model language
is probably caused by the very large number of involved variables,
or by the existence of many exceptional cases. 
In addition, language is a constantly changing phenomenon,
in space --in the form of dialects-- and in time 
--as new words appear, or old words are given new meanings--.

  Nevertheless, efforts have been done to model language. If not
as a whole, at least some phenomena have been very
accurately modelled. 
\medskip

  The main advantage of manual modelling is that the 
resulting constraints have linguistic meaning, and thus
can be revised and tuned to improve the model or to 
detect its weak points.

  The main drawback is that many years of human effort have to
be employed to obtain a model able to cope with more or less 
unrestricted language.
\medskip

  The research presented in this thesis was focused on automatically 
acquired models. But, since relaxation labelling accepts any kind of constraints, 
it can deal also with linguist written models --either on their own or combined
with other constraint models--. To check whether this ability was useful and
linguistic models were correctly applied and combined, we introduced 
small manually--written constraint models.
\medskip

  The first NLP task where we applied relaxation labelling was POS
tagging. Although n-gram models perform reasonably 
well\footnote{As described in section \ref{sec:tagging-POS} current
              n-gram based tagger present an accuracy of about $97\%$.},
we wanted to test the ability of the algorithm to integrate other
sources of knowledge.

  We used the following two kinds of manual constraints for POS tagging.
See section \ref{sec:selection} for details and results. 
A sample of the acquired constraints is presented in appendix \ref{app:constraints}.

The first
one was adapting previously existing context constraints to our
algorithm. The adapted constraints were those used by the
tagger described in \cite{Moreno94}. That tagger is a probabilistic
one where the user could write context constraints that are
applied {\em a posteriori}. Those constraints enabled the 
linguist to correct the most common errors made by the 
probabilistic tagger, and thus improve the final accuracy.
Those constraints had been developed for Spanish, and so
they could be used only in Spanish corpora. 

The second source of manual constraints was developing ourselves
a reduced model. The procedure was the following: the most frequent
errors made by a bigram HMM tagger were selected as difficult
cases and constraints were written to cover them.

 In both cases, a compatibility value has to be assigned to
the constraints in order to enable relaxation labelling to use them. 
Hand assignation of those values seems a very weak procedure
bound to subjective appreciations and prone to errors. For this
reason we applied an automatic procedure to estimate those
compatibilities.

  The hand written constraints were matched to the training
corpus, and the occurrences of the affected word/tag and the
context described by the constraint were computed. The joint
occurrences of both events were also computed. This enables 
us to estimate the probability of any of them, as well as
their conditional probabilities, and thus, compute any of the compatibility
measures described in section \ref{sec:compatibility}.
\medskip

  The constraint model used for the shallow parsing task was
completely developed by a linguist. Anyway, it is not a 
large-coverage model, and its labour cost was only some man hours.
The developing procedure is that of Constraint
Grammars, by successive model refinements over a training corpus.
Details can be found in \cite{Voutilainen97} as well as in 
section \ref{sec:experiments-ShP}. 
\medskip

  For the case of WSD, the necessary model would be much larger
than for the other tasks, since the number of possible combinations
is much higher. That made us rely mainly on automatic models. 
Nevertheless, a few selectional restrictions were hand written
for some high frequency verbs. Examples and results are presented in
section \ref{sec:experiments-WSD}.

\subsection{Statistical Acquisition}
\label{sec:statistical}

  The alternative to manually written models is obtaining
them automatically from existing corpora. The methods most commonly 
used to achieve this rely on a statistical basis. The language model 
is thus coded as a set of co-occurrence frequencies for
different kinds of phenomena. 

  This statistical acquisition is usually found in the form
of n-gram collection --as described is section \ref{sec:basic}--,
but more sophisticated acquisition techniques are also used, as 
for instance the selectional restrictions model described in 
section \ref{sec:semantic}.
\medskip

   A more recently introduced methods are those adapted from
machine learning field. Although some of them were developed
to work in symbolic discrete problems, they can be extended
to statistical environments, where learned knowledge is not 
black or white, but may have any intermediate value. The use
of constraints acquired in this way is described in section
\ref{sec:decision-trees}.

\subsubsection{Basic (Binary/Ternary constraints)}
\label{sec:basic}

  The most straightforward way of acquiring a statistical
language model is computing the co-occurrence frequencies of some 
selected features. These frequencies are then used to estimate
probabilities and derive the model. This is known as 
Maximum Likelihood Estimate (MLE).

   The selected features and kind of co-occurrences counted 
depend on what the model will be representing. For instance,
to get a model for part-of-speech tagging, one may count 
occurrences of tag bigrams or trigrams of consecutive words. 
For word sense disambiguation, it is common to find models
consisting of occurrence counts of word pairs inside a predefined
window, regardless of its exact relative position.
\medskip

   In our experiments we use several kinds of statistical
information collected from tagged corpus: 
For POS tagging, we use tag bigrams and trigrams. For shallow parsing
we use bigrams and trigrams of shallow syntactic roles for consecutive words.
For word sense disambiguation, the collected statistics are
co-occurrences of pairs of WordNet top synsets in the 
same sentence, co-occurrences of WordNet file codes, and finally, salient word
vectors for each WN file code, following the idea described in \cite{Yarowsky92}.

\subsubsection{Advanced (Decision Trees)}
\label{sec:decision-trees}

  The statistical information may be also acquired in
more sophisticated ways, not necessarily through mere
occurrence counting. We can use machine learning techniques 
to acquire that knowledge, either in a pure symbolic 
form, or adding statistical information.

  We acquired a POS model consisting of context constraints
more complex than simple n-grams. The constraints
took into account word forms, as well as context POS tags.
The used method is exposed below. Further details can
be found in \cite{Marquez95,Marquez97a,Marquez97b}.
\medskip

\noindent\underline{\bf Setting}
\nopagebreak\medskip

\noindent Choosing, from a set of possible tags, the proper syntactic tag for a
word in a particular context can be seen as a problem of classification.
Decision trees, recently used in NLP basic tasks such as
tagging and parsing \cite{McCarthy95,Daelemans96a,Magerman96}, are suitable
for performing this task.
\medskip

A decision tree is a {\it n}-ary branching tree that represents a {\it classification rule} 
for classifying the {\it objects} of a certain domain into a set of
mutually exclusive {\it classes}.
The domain objects are described as a set of attribute--value pairs, 
where each {\it attribute} measures a relevant feature of an object
taking a (ideally small) set of discrete, mutually incompatible {\it values}.

Each non--terminal node of a decision tree represents  a question on
(usually) one attribute.  For each possible value of this attribute
there is a branch to follow. Leaf nodes represent concrete classes.

Classify a new object with a decision tree is simply 
following the convenient path through the tree until a leaf is reached.

{\it Statistical} decision trees only differs from common decision trees in that leaf
nodes define a conditional probability distribution on the set of classes.

It is important to note that decision trees can be directly translated 
to rules considering, for each path from the root to a leaf, the conjunction 
of all questions involved in this path as a condition, and the class assigned to the
leaf as the consequence. Statistical decision trees would generate
rules in the same manner but assigning a certain degree of
probability to each answer. 

So the learning process of contextual constraints is performed 
by means of learning one statistical decision tree
for each class of POS ambiguity\footnote{Classes of ambiguity are 
                                         determined by the groups of
                                         possible tags for the words 
                                         in the corpus, i.e, {\it noun-adjective}, 
                                         {\it noun-adjective-verb}, 
                                         {\it preposition-adverb}, etc. }
and converting them to constraints (rules) expressing 
compatibility/incompatibility of concrete tags in certain contexts. 
\medskip

\noindent\underline{\bf Learning Algorithm}
\nopagebreak\medskip

\noindent The algorithm we used for constructing the statistical decision trees
is a non--incremental supervised learning--from--examples algorithm 
of the TDIDT (Top Down Induction of Decision Trees) family.
It constructs the trees in a top--down way, guided by the distributional
information of the examples, but not on the examples order
\cite{Quinlan86}. 
Briefly, the algorithm works as a recursive process that departs from
considering the whole set of examples at the root level and constructs
the tree in a top--down way branching at any non--terminal node
according to a certain {\it selected} attribute. 
The different values of this attribute induce a partition of the
set of examples in the corresponding subsets, in which 
the process is applied recursively in order to generate the 
different subtrees. 
The recursion ends, in a certain node,  either when all (or almost all) the 
remaining examples belong to the same class, or when the number of examples 
is too small. 
These nodes are the leafs of the tree and contain the conditional
probability distribution, of its associated subset of examples, on the 
possible classes.

The heuristic function for selecting the most useful attribute at each
step is of a crucial importance in order to obtain simple trees, since
no backtracking is performed. Attribute--selecting functions commomnly
used belong either to the {\it information}--based \cite{Quinlan86,Lopez91} 
family or to the {\it statistically}--based \cite{Breiman84,Mingers89a} family.
\medskip

\noindent\underline{\it Training Set}
\nopagebreak\medskip

\noindent For each class of POS ambiguity the initial
example set is built by selecting from the training corpus
all the occurrences of the words belonging to this ambiguity class.
More particularly, the set of attributes that describe each example 
consists of the part--of--speech tags of the neighbour words, and the
information about the word itself (orthography and the proper tag in
its context). The window considered in the experiments reported in
section \ref{sec:experiments-POS} is 3 words to the left and 2 to the right.
The following are two real examples from the training set for the 
words that can be preposition and adverb at the same time
(IN--RB conflict)\footnote{See appendix \ref{app:tagsets} for a
tagset description.}.
\medskip

\indent\indent{\tt VB DT NN as\_IN DT JJ}
\newline
\indent\indent{\tt NN IN NN once\_RB VBN TO}
\medskip

Approximately $90\%$ of this set of examples is used for the
construction of the tree. The remaining $10\%$ is used as fresh
test corpus for the pruning process.
\medskip

\noindent\underline{\it Attribute Selection Function}
\nopagebreak\medskip

\noindent For the experiments reported in
section \ref{sec:experiments-POS} we used a
attribute selection function due to \cite{Lopez91}
belonging to the information--based family.
It defines a distance measure between partitions and
selects for branching the attribute that generates the
partition closest to the {\it correct} one according to the 
training set.
\medskip

\noindent\underline{\it Branching Strategy}
\nopagebreak\medskip

\noindent Usual TDIDT algorithms consider a branch for each value of the selected 
attribute. This strategy is not feasible when the number of
values is big (or even infinite). In our case the greatest number of values
for an attribute is 45 ---the tag set size---
which is considerably big (this means that the branching factor could be 
45 at every level of the tree%
\footnote{In real cases the branching factor is much lower since not all 
          tags appear always in all positions of the context.}). 
Some systems perform a previous recasting of the attributes in order
to have only binary-valued attributes and to deal with binary trees
\cite{Magerman96}. 
This can always be done but the resulting features lose their
intuition and direct interpretation, and explode in number. 
We have chosen a mixed approach which consist of splitting for all
values and afterwards joining the resulting subsets into groups
for which we have not enough statistical evidence of being different 
distributions. This statistical evidence is tested with a $\chi^2$ test at a 95\%
confidence rate.
In order to avoid zero probabilities smoothing is performed.

Additionally, all the subsets that do not imply a reduction in the
{\it classification error} are joined together in order to have a bigger set
of examples to be treated in the following step of the tree
construction.
\medskip

\noindent\underline{\it Pruning the Tree}
\nopagebreak\medskip

\noindent Decision trees that correctly classify all examples of the training
set are not always the most predictive ones.
This is due to the phenomenon known as {\it over-fitting}. It occurs
when the training set has a certain amount of misclassified
examples, which is obviously the case of our training corpus (see section 
\ref{sec:experiments-POS-corpus}).
If we force the learning algorithm to completely classify the examples
then the resulting trees would fit also the noisy examples.

The usual solutions to this problem are: 1) Prune the tree, either during the 
construction process \cite{Quinlan93} or afterwards \cite{Mingers89b}; 2) Smooth the 
conditional probability distributions using fresh corpus%
\footnote{Of course, this can be done only in the case of statistical 
          decision trees.} \cite{Magerman96}.

Since another important requirement of our problem is to have small trees
we have implemented a post-pruning technique.
In a first step the tree is completely expanded
and afterwards it is pruned following a minimal cost--complexity 
criterion \cite{Breiman84}. Roughly speaking this is  
a process that iteratively cut those subtrees producing only marginal 
benefits in accuracy, obtaining smaller trees at each step. The
trees of this sequence are tested using a, comparatively small, fresh 
part of the training set in order to decide which is the one with the
highest degree of accuracy on new examples.
Experimental tests \cite{Marquez95} have shown that the pruning process 
reduces tree sizes at about 50\% and improves their accuracy in a 2--5\%.
\medskip

\noindent\underline{\it An Example}
\nopagebreak\medskip

\noindent Finally, we present a real example of the simple acquired contextual 
constraints for the preposition--adverb (IN--RB) conflict.

\begin{figure}[htb]
\hfil\epsfbox{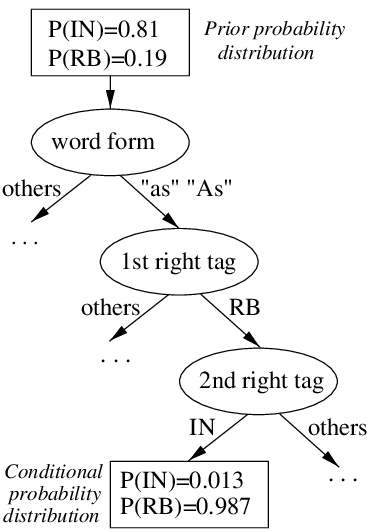}\hfil

\begin{center}
\begin{tabular}{rlccrl} \\
           $-5.81$ &{\tt (IN)}          & & & $2.366$ &{\tt (RB)} \\
                   &{\tt (0 "as" "As")} & & &         &{\tt (0 "as" "As")} \\
                   &{\tt (1 RB)}        & & &         &{\tt (1 RB)} \\
                   &{\tt (2 IN);}       & & &         &{\tt (2 IN);}
\end{tabular}
\end{center}
\caption{Example of a decision tree branch with its equivalent constraints.}\label{fig:arbre}
\end{figure}

Figure \ref{fig:arbre} shows a sample tree branch acquired by the 
algorithm and the constraints into which it is translated.
These constraints express the compatibility (either positive or
negative) of the constraint head --first line-- with the 
context expressed by the conditions following it. The syntax
used here is that of \cite{Karlsson95} Constraint Grammars.
The compatibility value for each constraint is the mutual information 
between the head tag and the context \cite{Cover91}. It is directly computed from
the probabilities in the tree.
Some other sample constraints acquired by the algorithm are presented 
in appendix \ref{app:constraints}.

\subsubsection{Semantic Constraints}
\label{sec:semantic}

  Our interest in methods for obtaining context constraints is 
due to the need of a language model which enables relaxation labelling
to perform disambiguation tasks.

  In previous sections we have seen several techniques to acquire
context constraint, mainly aiming to build a model oriented
to part-of-speech tagging. In this section we will address the 
issue of how to acquire a model to perform word sense 
disambiguation.
\medskip

  Modelling the semantic aspects of language is usually
harder than POS or syntax modelling, and the automatic 
acquisition of semantic constraints is a research field
with still many open questions. As noted in section
\ref{sec:tagging-semantic}, the chosen sense granularity has a
very large influence on a WSD system. In our case,
it also affects greatly the needed constraint model and its acquisition,
since a very fine grained sense distinction will require a much more
precise model, and thus a larger number of constraints and a 
better acquisition procedure than a more coarse sense classification.
\medskip

  The flexibility of the relaxation algorithm used to perform the 
disambiguation will dim the influence of the sense granularity. 
The multi--feature approach we are taking (see sections 
\ref{sec:experiments-ShP} and \ref{sec:experiments-WSD}) will 
enable us to use different granularity levels. 

   That is, since a reading for a word may include different features,
such as POS, lemma, sense, etc., we can include, for instance, a feature 
consisting of a fine--grained sense identifier 
(e.g. WN synset)\footnote{As noted in section \ref{sec:tagging-semantic},
                                              WordNet is a concept hierarchy, where each 
                                              sense is represented by a set of synonym words
                                              (a {\em synset}). In addition, synsets are 
                                              grouped in thematic files, each one with its
                                              own {\em file code}.}, 
another one consisting of a coarse--grained classification (e.g. WN top), and 
even a third one containing some subject information (e.g. WN file code). 

   This will enable the constraints to express relationships between the 
different levels of granularity, according to the needs of every specific case. 
In addition, since the relaxation labelling algorithm will use all available 
constraints, in the case that they do not include all necessary knowledge to 
fully disambiguate the right fine--grained sense, at least a sense with the 
right coarse class or subject will be selected. 
See section \ref{sec:experiments-WSD} for details.
\medskip

  In this research, several automatic techniques for acquiring
a constraint model for WSD have been experimented.
They are briefly described below, from the simplest co-occurrence 
collection to the sophisticated selectional restrictions acquisition
technique developed and applied by \cite{Ribas95}.

\begin{itemize}
\item The simplest methods for acquiring semantic constraints are the use of 
co-occurrence information (see section \ref{sec:basic}). For instance,
computing the co-occurrence ratio of pairs of verb and/or noun senses.

In our case, we used the top synsets
in WordNet hierarchy, regarded as class identifiers\footnote{That is, a verb sense and 
                                                             a noun sense will be considered 
                                                             to have the same co-occurrence 
                                                             ratio than their respective 
                                                             tops.}.
The tops co-occurrences should be computed from a sense--tagged corpus, computing 
each occurrence of a sense as an occurrence of its top. The same technique was 
applied using WN file codes, instead of top synsets, as class identifiers.

\item Another possible method to derive simple semantic constraints 
are collecting salient word lists for each class
(either top synset or WN file code) in the style of \cite{Yarowsky92}. 
This technique can be used either on supervised or unsupervised corpora
and constitutes an easy procedure --although maybe not as precise as one 
might want-- to build semantic models.

   For each word in the corpus, all the content words appearing in its
near context are collected as belonging to the salient words list of the 
focus word sense --if the right sense is known--, or to all the lists of 
all possible senses for the focus word if the corpus is unsupervised. 
Then, a threshold is established 
and only the most relevant context words for each sense are kept. When
disambiguating a new occurrence of a word, the chosen sense is that 
with highest matching ratio between the sense salient words list and
the current context.

\item Another interesting possibility for automatically acquiring semantic 
constraints is using conceptual distance (e.g. over WordNet \cite{Sussna93})
between pairs of noun senses. It seems to be more reliable, since 
the information is not drawn from a corpus, but from a hand--built 
taxonomy. In addition, no sense--tagged corpora is needed to acquire
the model.

  The semantic distance approach is based on the assumption that 
conceptually close synsets will tend to appear in the same context.
This assumption does not always hold, as discussed in 
section \ref{sec:experiments-WSD}. Moreover,
conceptual distance over a taxonomy such as WordNet can only be 
computed between senses belonging to the same sub-hierarchy 
(nouns, verbs, adj, adv), which limits the power of this method.
\end{itemize}

  Probably, none of the above kinds of constraints
would be described as a powerful and natural way 
to express semantic relationships. In addition, experiments reported in
section \ref{sec:experiments-WSD} subscribe the idea that they
do not capture all necessary knowledge to accurately disambiguate
word senses.
\medskip

 One of the most natural way of expressing semantic constraints are the
selectional restrictions that a phrase head imposes to its 
complements and vice-versa, for instance, the noun {\em table} 
accepts adjectives referring to its shape ({\em round, square, \ldots}),
color ({\em brown, dark, \ldots}), size ({\em big, tall, \ldots}), etc., but
it does not accept adjectives such as {\em intelligent, powerful, \ldots}.
In the same way, verbs impose constraints on their objects, for instance,
a subject for verb {\em think} must be {\tt human} --or at least {\tt animate}--, the
direct object for verb {\em eat} must be {\tt food}, etc.
\medskip

 In our case, since this research is mainly on constraint 
{\em application} and not on constraint {\em acquisition}, 
we focused on selectional restrictions imposed by a verb to
its objects. This choice was made in order to be able to use 
the selectional restrictions automatically acquired by 
\cite{Ribas94,Ribas95}. 

  Although the research developed by \cite{Ribas95} was mainly focused
on unsupervised learning --due to the lack of large available sense--tagged 
corpora--, for our purposes of applying a constraint model to perform WSD,
the supervised option seems to provide with more accurate restrictions.
  Thus, although \cite{Ribas95} applied his technique to both cases,
we only will use the constraints he acquired through supervised learning.
\medskip

  The procedure used by \cite{Ribas95} to obtain selectional constraints 
from corpora is outlined below. To find out more about this technique,
either in its supervised or unsupervised version, see \cite{Resnik93,Ribas95}.
\medskip

\noindent\underline{\bf Selectional Restrictions Acquisition.}
\nopagebreak\medskip

\begin{figure}
\begin{itemize}
\item {\bf Previous semantic knowledge}
\end{itemize}

\begin{picture}(0,0)%
\includegraphics{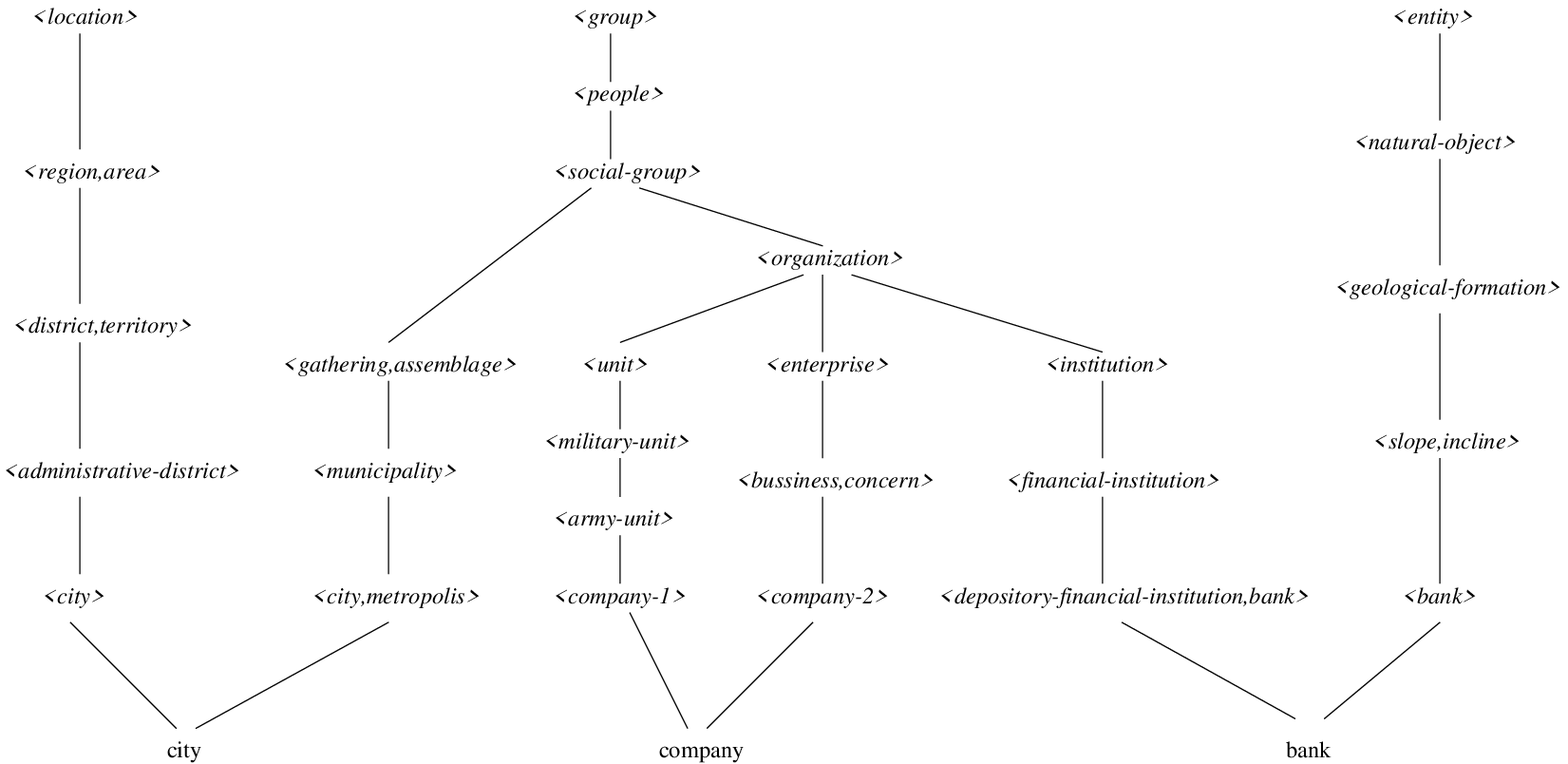}%
\end{picture}%
\setlength{\unitlength}{0.008750in}%
\begin{picture}(680,385)(40,440)
\end{picture}

\begin{itemize}
\item {\bf Three examples of use of {\it pay}}
{\it
\begin{description}
\item For nearly a decade, {\rm banks} have \underline{paid} high interest rates to
small customers.
\item The {\rm company} still has to \underline{pay} its debts to creditors.  
\item The {\rm city} has \underline{paid} \$ 95,142 to Mr. Dinkins in matching  
      funds although his campaign records are incomplete.
\end{description}
}

\item {\bf The acquired Selectional Restriction}
\begin{description}
\item  {\it (pay, SUBJ, $<$social-group$>$)}
\end{description}
\end{itemize}
\caption{Example of the acquisition of Selectional Restrictions.} 
\label{fig:selectional}
\end{figure}

\noindent The scenario on which the acquisition technique 
developed by \cite{Ribas95} should extract
selectional restrictions is displayed in Figure \ref{fig:selectional},
where, departing from the three examples of use of the verb {\sl pay}
and knowing the semantic categorizations of {\sl banks}, {\sl company}
and {\sl city} as {\it social-group}, the system should induce that
the verb {\sl pay} imposes a selectional restriction over its subject
that constrains the content word filling it to be a member of the
semantic type {\it social-group}. Therefore, the aim of the system is
to extract, for each word (being a head and having enough
occurrences) in the corpus and for each of its syntactic complements, a list of
the alternative selectional restrictions that the head word is imposing on
the complement words.

 Although selectional restrictions have been used to express semantic
constraints holding on different syntactic and functional
configurations, the work in \cite{Ribas95} --whose results we are using--
focused only on those holding between
verbs and their complements. The methodology can be easily exported to
other configurations. Moreover, considering the theoretical
and practical controversy on doing the argument/adjunct
distinction \cite{Adams91} and given that the
source of co-occurrences used --the Penn Treebank \cite{Marcus93}--
is not reliably marked with such distinction, it
was not taken into account when acquiring selectional restrictions.
\medskip

\noindent\underline{\it The Basic Technique}
\nopagebreak\medskip

\noindent The basic technique used in \cite{Ribas95} to acquire selectional restrictions
is a slight variation of the methodology first introduced in
\cite{Resnik92} and further developed in \cite{Resnik93}. 

 From the collection of nouns that co-occur as
particular complements of a given verb, the basic method tries to
generalize the appropriate semantic classes (selectional restrictions)
by selecting a level in a taxonomy (WordNet in our case). 
\medskip

The input to the process is a set of co-occurrence triples 
{\it (verb, syntactic-relationship, noun)}\footnote{{\it verb} is the verb lemma,
                                                    {\it noun} is the noun lemma, 
                                                    and {\it syntactic-relationship} may 
                                                    be subject, direct object, indirect 
                                                    object, or prepositional object. 
                                                    In the last case the relationship is
                                                    labelled with the specific preposition.}
extracted from syntactic analysis of the corpus.
Restrictions are only acquired for noun senses, that is, no knowledge about
which is the right verb sense according to its object nous is extracted.
If the algorithm does not know the
appropriate sense for each noun in the co-occurrence triples, it considers
all the noun hyperonyms for all possible noun senses as candidate classes
(unsupervised training).
Otherwise, if the training corpora is sense--tagged, only the hypernonyms 
of the right sense are used as candidate classes (supervised training).
\medskip

  Once the candidate classes have been obtained for each pair 
{\it (verb,syntactic-relationship)}, 
only those classes that generalize triples 
with a higher frequency than a given threshold are further considered. 
Their association is evaluated by means of a statistical measure, 
{\it Association Score} (see section \ref{sec:compatibility}), 
derived from the co-occurrence of verbs and classes of nouns. 

  The statistical association is used by a selection process to 
choose the best classes to convey the selectional restrictions. 
The algorithm, for every pair
({\it verb}, {\it syntactic-relationship}), generalizes a set of
selectional restrictions, i.e. pairs ({\it class}, {\it statistical-preference}).
 This is done by selecting the candidate
class with highest association score, and removing all its hypermonyms and
hyponims from the set of candidate classes. Repeating this procedure until
no candidate classes are left, the resulting selected classes for the verb
and syntactic relationship are mutually disjoint, that is they are not
related by hyperonymy, and are a generalization of the classes in the
co-occurrence triples.
\medskip

\noindent\underline{\it An Example}
\nopagebreak\medskip

  As an example of the different results produced by the supervised and 
unsupervised methods, the candidate classes considered by each kind of training for
the situation presented in figure \ref{fig:selectional} are shown
in tables \ref{taula:candidates-superv} and \ref{taula:candidates-unsuperv},
respectively.
\medskip

\begin{table}[htb] \centering
\begin{tabular}{|l|l|l|} \hline
\multicolumn{1}{|c|}{from {\it city}} &\multicolumn{1}{|c|}{from {\it company}} &\multicolumn{1}{|c|}{from {\it bank}} \\ \hline
 {\it $<$city,metropolis$>$}      &{\it $<$company-2$>$}        &{\it $<$depository-financial-institution,bank$>$} \\ 
 {\it $<$municipality$>$}         &{\it $<$bussiness,concern$>$}&{\it $<$financial-institution$>$} \\
 {\it $<$gathering,assemblage$>$} &{\it $<$enterprise$>$}       &{\it $<$institution$>$} \\
 {\it $<$social-group$>$}         &{\it $<$organization$>$}     &{\it $<$organization$>$} \\
 {\it $<$people$>$}               &{\it $<$social-group$>$}     &{\it $<$social-group$>$} \\
 {\it $<$group$>$}                &{\it $<$people$>$}           &{\it $<$people$>$} \\
                                  &{\it $<$group$>$}            &{\it $<$group$>$} \\ \hline
\end{tabular}      
\caption{Candidate classes for {\it (pay, SUBJ)} using supervised training.}
\label{taula:candidates-superv}
\end{table}

\begin{table}[htb] \centering
\begin{tabular}{|l|l|} \hline
\multicolumn{2}{|c|}{from {\it city}} \\ \hline
   {\it $<$city$>$}                    &{\it $<$city,metropolis$>$} \\
   {\it $<$administrative-district$>$} &{\it $<$municipality$>$}    \\
   {\it $<$district,territory$>$}      &{\it $<$gathering,assemblage$>$} \\
   {\it $<$region,area$>$}             &{\it $<$social-group$>$}    \\
   {\it $<$location$>$}                &{\it $<$people$>$}  \\
                                       &{\it $<$group$>$}   \\ \hline \hline
\multicolumn{2}{|c|}{from {\it company}} \\ \hline
   {\it $<$company-1$>$}      & {\it $<$company-2$>$}  \\
   {\it $<$army-unit$>$}      & {\it $<$bussiness,concern$>$}  \\
   {\it $<$military-unit$>$}  & {\it $<$enterprise$>$}  \\
   {\it $<$unit$>$}           & {\it $<$organization$>$}  \\
   {\it $<$organization$>$}   & {\it $<$social-group$>$}  \\
   {\it $<$social-group$>$}   & {\it $<$people$>$}  \\
   {\it $<$people$>$}         & {\it $<$group$>$}  \\
   {\it $<$group$>$}          &                     \\ \hline  \hline
\multicolumn{2}{|c|}{from {\it bank}} \\ \hline
   {\it $<$depository-financial-institution,bank$>$} & {\it $<$bank$>$}  \\
   {\it $<$financial-institution$>$}                 & {\it $<$slope,incline$>$}  \\
   {\it $<$institution$>$}                           & {\it $<$geological-formation$>$} \\
   {\it $<$organization$>$}                          & {\it $<$natural-object$>$}  \\
   {\it $<$social-group$>$}                          & {\it $<$entity$>$}  \\
   {\it $<$people$>$}                                &                     \\
   {\it $<$group$>$}                                 &                     \\ \hline
\end{tabular}      
\caption{Candidate classes for {\it (pay, SUBJ)} using unsupervised training.}
\label{taula:candidates-unsuperv}
\end{table}

  The usupervised acquisition technique would present the 
following behaviour: Assuming that the {\it $<$social-group$>$} sense
has a higher {\it Association Score} than its relative (hyponym or hyperonym) 
senses, it would be selected as the best candidate. Its relatives would then 
be eliminated from the candidate classes set, and thus, the constraint
{\it (pay, SUBJ, $<$social-group$>$)} would be extracted. 
Nevertheless --unless they had been removed by the threshold filtering--
the {\it $<$location$>$} and {\it $<$entity$>$} class
families are still candidates, thus, 
the sense with highest score in each family would be selected. 
This would cause the final result to consist of three selectional
constraints: the first one is the expected solution, 
with a high association score, and the other two --with a presumably lower 
association score-- are caused by the noise introduced by the unsupervised training
and a too low threshold.
\begin{list}{}{}
\item {\it (pay, SUBJ, $<$social-group$>$)}
\vspace{-1ex}
\item {\it (pay, SUBJ, $<$entity$>$)} 
\vspace{-1ex}
\item {\it (pay, SUBJ, $<$location$>$)} 
\end{list}
\medskip

  In supervised training, the process would be the same, but since 
the {\it $<$location$>$} and {\it $<$entity$>$} class families are not included in the
candidate classes set, the corresponding restrictions would not be extracted.
Thus, the only acquired restriction would be:
\begin{list}{}{}
\item {\it (pay, SUBJ, $<$social-group$>$)}
\end{list}


%% file: experiments.tex
\chapter{Experiments and Results}
\label{cap:experiments}

  In this chapter we will describe the experiments performed to test
the utility of relaxation labelling for NLP purposes. The reported experiments
can be classified on three classes: First, a set of tests --using POS tagging as
a benchmark-- that were performed in order to establish which is the most
appropriate parameterization for the relaxation algorithm in our case. 
Second, tests of the application of relaxation labelling to POS-tagging aiming
to establish its ability to deal with different kinds of information, and to establish
whether it can outperform current POS taggers.
Third, experiments on applying relaxation labelling to NLP tasks other
than POS tagging, namely, word sense disambiguation and shallow parsing. The later
experiments were also performed to test the ability of the algorithm to deal with
multi-feature models as well as its ability to integrate multi-source knowledge.
\medskip

  In section \ref{sec:selection} the experiments performed to
establish the best parameterization are described. 
Section \ref{sec:experiments-POS} describes the set of experiments 
on applying the algorithm to POS-tagging, and section \ref{sec:experiments-NLP} 
exposes how relaxation labelling was applied to 
shallow parsing and word sense disambiguation and the results obtained.

\section{Parameter selection experiments}
\label{sec:selection}
 The first set of experiments were performed on the task of POS tagging because it
is one the simplest and most straightforward NLP tasks. In addition, it is 
almost straightforward to model it to be solved by the relaxation labelling algorithm. 

The performed experiments aimed to find the most appropriate
parameters for relaxation labelling when applied to this kind of tasks, in order
to establish a starting point for further use of the algorithm at
more complex NLP tasks.
\medskip

 The experiments consisted of tagging a corpus using all logical combinations of
parameters for the algorithm. The algorithm parameters are those described 
in section \ref{sec:parameterization},
that is: support function, updating function and compatibility values.

 Different kinds of constraints (bigrams, trigrams, hand-written, 
and all combinations of them) were used, as a first test of the 
algorithm flexibility respect to the used language model. The different
constraints where used separately as well as combined.

 We also tested different normalization functions for support values, 
and made some trials looking for a support function
specifically designed for the case of POS-tagging as well as
applying the different kinds of constraints in a back--off hierarchy.
\medskip

  The experiments were repeated on the following three corpora. Each one of 
them had some feature that made its use interesting.

\begin{list}{}{}
\item[\underline{Corpus {\bf SN} (Spanish Novel)}]
      Train set: $15$ Kw. Test set: $2$ Kw. Tag set\footnote{A listing of tags and
                                                             descriptions can be found
                                                             in appendix \ref{app:tagsets}.} 
      size: 68. \\
      This corpus was chosen to test the algorithm in a language
      distinct than English, and because previous work
      \cite{Moreno94} on it provides us with a good benchmark
      and with linguist written constraints. 
\item[\underline{Corpus {\bf Sus} (Susanne)}]
        Train set: $141$ Kw, Test set: $6$ Kw. Tag set size: 150.\\
        The interest of this corpus is to test the algorithm with a
        large tag set.
\item[\underline{Corpus {\bf WSJ} (Wall Street Journal)}] 
        Train set: $1055$ Kw. Test set: $6$ Kw. Tag set size: 48.\\
        The interest of this corpus is obviously its size, which gives
        a good statistical evidence for automatic constraints acquisition.
\end{list}

The performed experiments with their results 
and conclusions are published in \cite{Padro95,Padro96a,Padro96b}.

\subsection{Baseline results}

   In order to have a comparison reference we will evaluate the
performance of two taggers: A blind most-likely-tag tagger and a bigram HMM
tagger \cite{Elworthy93} performing Viterbi algorithm.
The training and test corpora will be the same for all taggers.

   Results obtained by the baseline taggers are found in 
table \ref{taula-millors-base} (figures show precision percentage
over ambiguous words).

\begin{table}[htb] \centering
\begin{tabular}{|l|r|r|r|r|} \hline
                &SN            &Sus               &WSJ           \\ \hline
Most-likely     &$69.62\%$   &$86.01\%$       &$88.52\%$   \\ \hline
HMM             &$94.62\%$   &$93.20\%$       &$93.63\%$   \\ \hline
\end{tabular}
\caption{Results achieved by conventional taggers.}
\label{taula-millors-base}
\end{table}

   The Most-likely tagger produces poorer results on the {\bf SN} corpus 
than on the others because of the reduced size of this corpus, which does 
not provide enough evidence for a most-likely model.

\subsection{Relaxation labelling results}
\label{sec:selection-results}

   In this section we will expose the results achieved by the 
relaxation labelling algorithm on the three test corpus.

   Although results for each combination of parameters were obtained,
the tables presented here contain only the best results produced by
{\em any} parameter combination. As noted below, the best results
happened to be obtained in most cases by the same parameterizations.
\medskip

   For each parameter combination, the algorithm was tested with all 
possible combinations of constraints. The sets of constraints used 
were bigram constraints ({\bf B}), trigram constraints ({\bf T}) and 
hand-written constraints ({\bf H}).
\medskip

   The sets of hand-written constraints were built according to
the procedure described in section \ref{sec:manuals}, which can
be summarized as follows:

For {\bf WSJ} and {\bf Sus} corpora, the test corpus was tagged using
the baseline HMM tagger. The most frequent errors made by the 
HMM tagger were analyzed, and constraints to cover those cases were 
hand-written. That produced a set of $12$ constraints for {\bf WSJ} 
corpus and a set of $66$ constraints for {\bf Sus}.
 
For {\bf SN} corpus, we adapted some $50$ context constraints proposed 
by \cite{Moreno94}, who used them to correct the most common errors 
of his probabilistic tagger.

  The compatibility value for these constraints were
computed from their occurrences in the corpus, that is, the 
number of occurrences of the affected word or tag and the number of
occurrences of the context described by the constraint is collected
from the training corpus. This provides the necessary information to
compute the compatibility value for the constraint in any of the
forms described in section \ref{sec:compatibility}.
\medskip

Best results --in precision over
ambiguous words-- obtained by relaxation using every combination of 
constraint kinds are shown in table \ref{taula-millors-tots}. 

\begin{table}[htb] \centering
\begin{tabular}{|l|r|r|r|r|} \hline
    &SN                  &Sus                 &WSJ          \\ \hline
B   &$95.77\%$         &$91.65\%$         &$89.34\%$  \\ \hline
BH  &$96.54\%$         &$92.50\%$         &$89.24\%$  \\ \hline
T   &$90.00\%$         &$88.60\%$         &$90.87\%$  \\ \hline
BT  &$93.85\%$         &$89.33\%$         &$90.81\%$  \\ \hline
TH  &$92.31\%$         &$89.02\%$         &$90.78\%$  \\ \hline
BTH &$95.00\%$         &$89.83\%$         &$90.94\%$  \\ \hline
\end{tabular}
\caption{Best relaxation results using every combination of constraint kinds.}
\label{taula-millors-tots}
\end{table}

  The results presented in table \ref{taula-millors-tots} are the best results
obtained for any parameter combination.
Nevertheless, it is interesting to state that all of them
were obtained using support function described in equation (\ref{support-s}) 
and most of them with the updating function in equation (\ref{updating-c}) 
and using mutual information as compatibility values. 

   This suggests that this parameter combination is the most appropriate
for this kind of task. Further discussion on this issue can be found in
section \ref{sec:results-POS}.
\medskip

  Some general issues we can draw from this results are:
\begin{itemize}
\item In the same conditions than HMM taggers -i.e. using only bigram information-- 
relaxation only performs better than HMM for the
small corpus {\bf SN}, and the bigger the corpus is, the worse
results relaxation obtains.
\item Using trigrams
is only helpful in {\bf WSJ}. This is because the training corpus for {\bf WSJ}
is much bigger than in the other cases, and so the trigram model
obtained is good, while for the other corpora, the training set seems
to be too small to provide a good trigram information. 
\item We can observe that there is a general tendency to ``the
more information, the better results'', that is, when using {\bf BTH} we get
better results that with {\bf BT}, which is in turn better than {\bf T} alone.
\item In all corpora results improve when adding hand-written constraints,
except in {\bf WSJ}. This is because the constraints used in this case
are few (about 12) and only cover a few specific error cases (mainly
the distinction past/participle following verbs {\it to have} or {\it
to be}).
\end{itemize}

\subsection{Stopping before convergence}
\label{sec:selection-stopping}

  All results presented in section \ref{sec:selection-results} were obtained 
stopping the relaxation algorithm
when it reaches convergence (no new significant changes are produced from
one iteration to the next), but relaxation labelling algorithms do not
give necessarily their best results at convergence\footnote{
         This is due to two main reasons: (1)The optimum of the support
         function does not correspond {\em exactly} to the best solution for the
         problem, that is, the chosen function is only an approximation of the
         desired one. And (2) performing too much iterations can produce a more
         probable solution, which will not necessarily be the correct one.} 
\cite{Eklundh78,Richards81,Lloyd83}, 
or not always one needs to achieve convergence to know what the result
will be \cite{Zucker81}. 
So they are often stopped
after a few iterations. Actually, what we are doing is changing our
convergence criterion to one more heuristic than ``stop when
there are no more changes''.
\medskip

\begin{table}[htb] \centering
\begin{tabular}{|r|r|r|r|} \hline 
  SN                      &Sus                    &WSJ              \\ \hline
  $96.92\%$ (12)        &$93.78\%$ (6)        &$94.17\%$ (6)  \\ \hline
\end{tabular}
\caption{Best results stopping before convergence.}
\label{taula-millors-stop}
\end{table}

  The results presented in table \ref{taula-millors-stop} are the 
best overall results that we would obtain if we had a criterion which stopped the
iteration process when the result obtained was an optimum. The number
in parenthesis is the iteration at which the algorithm should be
stopped. These results are clearly better than those obtained at relaxation convergence,
 and also outperform the established baseline taggers.
\medskip

  To find out which one was the right moment to stop iteration, three lines 
of research were used (see \ref{sec:convergence}): 
\medskip

First, several convergence criteria were tested,
all of them based on the variation produced from one iteration to the next,
to check whether there was any relationship between those measures and
the optimal iteration.
The tested criteria were: global Euclidean distance (taking each weight of each 
tag as a dimension of a n-dimensional space), average Euclidean distance 
per word (\cite{Eklundh78}), average tag support variation, maximum tag 
support variation, and their respective 
first derivatives (that is, the variation on the variation from one iteration
to the next). 

None of these criteria seemed to keep any relationship with
the optimal stopping iteration, that is, none of them had any
particular behaviour when the algorithm went through the iteration
where the optimal result was obtained.
\medskip

 Second, hand analysis of the errors made or solved by the 
algorithm when approaching convergence was performed. That implied 
tagging a test corpus of $50$ Kw both waiting for convergence and
stopping the algorithm at the iteration which yielded the best result.
Then the $72$ errors introduced by convergence and the $52$ errors that it 
corrected were hand analyzed. 

 Those analysis showed that 
the errors introduced by convergence were mainly due to noise in
the training or test corpora, while the corrected ones were mostly
real corrections. See section \ref{sec:selection-conclusions} for
further discussion.
\medskip

 Third, the algorithm convergence is closely related to the normalization
factor for support values\footnote{
                   As described in section \ref{sec:updating} and discussed
                   in section \ref{sec:convergence}, when
                   using equation \ref{updating-c} support values must be 
                   in $[-1,1]$. Since mutual information is not necessarily 
                   in this range, normalization must be performed.}
since modifying the normalization interval has an effect similar to changing
the step size in a gradient algorithm. So, experiments were performed in 
order to find an objective manner to establish the most suitable normalization
factor, and to establish its relation with the stopping criterion.

  The results of this line showed that changing the normalization factor
changes the iteration at which the optimal result is obtained, as well as
the optimal result itself, and that the highest result is obtained when
the normalization factor selects as the stopping iteration that of convergence.

  As an example, table \ref{taula-convg} shows the accuracy obtained at 
convergence and at the optimal stopping iteration for different normalization
factor values.

\begin{table}[htb] \centering
\begin{tabular}{|c|c|c|} \hline
  Normalization  &convergence  &optimal iteration      \\
  factor         &accuracy     & (it.\#) -- accuracy  \\ \hline
             $5$ & $86.84$     &  $(2)$ -- $93.62$  \\
            $10$ & $89.63$     &  $(6)$ -- $94.26$  \\
            $15$ & $90.79$     &  $(9)$ -- $94.35$  \\
            $20$ & $91.57$     & $(12)$ -- $94.33$  \\
            $25$ & $92.34$     & $(13)$ -- $94.34$  \\
            $30$ & $92.83$     & $(16)$ -- $94.35$  \\
            $35$ & $93.10$     & $(19)$ -- $94.34$  \\
            $40$ & $93.41$     & $(23)$ -- $94.35$  \\
            $45$ & $93.63$     & $(24)$ -- $94.34$  \\
            $50$ & $93.75$     & $(27)$ -- $94.28$  \\
            $55$ & $93.89$     & $(31)$ -- $94.24$  \\
            $60$ & $93.93$     & $(34)$ -- $94.19$  \\
            $65$ & $93.94$     & $(37)$ -- $94.12$  \\
            $70$ & $93.99$     & $(39)$ -- $94.06$  \\
            $75$ & $93.97$     & $(42)$ -- $94.01$  \\
            $80$ & $93.99$     & $(62)$ -- $94.00$  \\
            $85$ & $93.93$     & $(64)$ -- $93.94$  \\
            $90$ & $93.87$     & $(50)$ -- $93.88$  \\
            $95$ & $93.77$     & $(85)$ -- $93.78$  \\
           $100$ & $93.66$     & $(89)$ -- $93.66$  \\ \hline
\end{tabular}
\caption{Results at convergence and at the optimal stopping iteration for different
normalization factor values.}
\label{taula-convg}
\end{table}

\subsection{Searching a more specific support function}
\label{sec:specific-supp}

  The support functions described in section \ref{sec:support} 
are traditionally used in relaxation algorithms. 
It seems better for our purpose to choose an additive one,
since the multiplicative functions might yield zero or tiny
values when -as in our case- for a certain variable or tag no
constraints are available for a given subset of variables.
\medskip

  Since those are general--purpose functions, we attempted to find a support
function more specific for our problem, inspired on the sequence probability
maximization performed by HMMs. 

  Since HMMs find the maximum sequence probability and relaxation is a 
maximization algorithm, we can try to make
relaxation maximize the sequence probability and we should get similar
results, which could be improved afterwards by adding new information
to the model. As relaxation labelling performs a vector optimization 
--as described in section \ref{sec:relax-alg}-- 
mainly dependent on the support function, to make
the algorithm maximize the sequence probability, we defined 
the support function as the sequence probability, computed in
the same way than in a classical probabilistic tagger.
\medskip

Being: \\
\indent\indent $t^k$ the tag for variable $v_k$ with highest weight value at
the current time step.\\
\indent\indent $\pi(v_1,t^1)$ the probability for the sequence to start in tag
$t^1$.\\
\indent\indent $P(v,t)$ the lexical probability for the word represented by
$v$ to have tag $t$.\\
\indent\indent $T(t_1,t_2)$ the probability that tag $t_2$ follows tag $t_1$, (bigram probability).
\medskip

We define: \\
\indent $B_{ij}=\pi(v_1,t^1) \times ({\displaystyle \prod_{k=1}^{i-2}P(v_k,t^k) \times 
        T(t^k,t^{k+1})}) \times P(v_{i-1},t^{i-1}) \ \times$ \hfill \\

\hfill $T(t^{i-1},t_j^i) \times P(v_i,t_j^i) \times T(t_j^i,t^{i+1}) \times
      ({\displaystyle \prod_{k=i+1}^{N-1}P(v_k,t^k) \times T(t^k,t^{k+1})}) 
       \times P(v_N,t^N)$\\

 Since it incorporates only bigram information (the $T(t^k,t^{k+1})$ transitions),
using $B_{ij}$ as as support function would have enabled us to use only binary 
constraints, so we included in our new support function the contribution of 
higher order constraints.
\medskip

The contribution of trigram constraints,
$$
T_{ij}={\displaystyle \sum_{r \in R^3_{ij}} Inf(r,i,j)}
$$
\begin{tabbing}
\`where $R^3_{ij}$ is the set of all trigram\\
\`constraints on tag $j$ for word $i$.
\end{tabbing}
\medskip

And the contribution of higher--order constraints
$$
C_{ij}={\displaystyle \sum_{r \in R^H_{ij}} Inf(r,i,j)}
$$
\begin{tabbing}
\`where $R^H_{ij}$ is the set of all hand-written\\
\`constraints on tag $j$ for word $i$.
\end{tabbing}
\medskip

   We chose to combine the support provided by bigrams ($B_{ij}$) with
the support provided by trigrams ($T_{ij}$) and hand-written
constraints ($C_{ij}$) in a multiplicative form because since $B_{ij}$
is computed as the probability of the whole sequence, it is many
magnitude orders smaller than $T_{ij}$ and $C_{ij}$, which are
computed locally; thus, adding them would have the effect of losing
the information provided by $B_{ij}$, since it would be too small to
affect the other figures. 

  But just multiplying them yields another problem: we do not have trigram
or hand written constraints for each word or tag. Then a tag with no
such an information will have $T_{ij} = C_{ij} = 0$ (or, if we perform
some kind of smoothing, a tiny value), and multiplying this value by
$B_{ij}$ would make the support value drop. That is, a tag with trigram
or hand-written constraints information would have less support than
another one with only bigram information, even when the trigram
information was {\em positive}.
  Since we want trigrams
and other constraints to {\em increase} the support when positive and
to {\em decrease} it when negative, we add one to the
value before multiplying it, so when no trigrams are used, support
remains unchanged, but if extra information is available, it
increases/decreases the support.
\medskip

 Thus, we obtain the new support function:
\begin{equation} 
\label{suport-q}
S_{ij} = B_{ij} \times (1 + T_{ij}) \times (1 + C_{ij})
\end{equation}

  Results obtained with this specific support function
are summarized in table \ref{taula-millors-Sq}.

\begin{table}[htb] \centering
\begin{tabular}{|r|r|r|r|} \hline
  SN                      &Sus                  &WSJ            \\ \hline
  $94.23\%$ (1-3)       &$92.31\%$ (6)      &$93.60\%$(1) \\ \hline
\end{tabular}
\caption{Best results using a specific support function.}
\label{taula-millors-Sq}
\end{table}
   
   Using this new support function we obtain results slightly below
those of the HMM tagger. Although our support function is based on 
the sequence probability,
which is what HMM taggers maximize, we get worse results. 
There are two main reasons for that. The first one is that we are not
optimizing {\em exactly} the sequence probability, but a support function 
based on it. The second reason is that relaxation is not an
algorithm that finds global optima and can be trapped in local
maxima.

\subsection{Combining information in a back-off hierarchy}

   We also experimented combining bigram and trigram information in a back-off
 mechanism: Use trigrams if available and bigrams when not. 
   
   Results obtained with that technique are shown in table \ref{taula-millors-K}
\medskip

\begin{table}[htb] \centering
\begin{tabular}{|r|r|r|r|} \hline
   SN                     &Sus                    &WSJ             \\ \hline
   $92.31\%$ (3-4)      &$93.66\%$ (4)        &$94.29\%$ (4) \\ \hline
\end{tabular}
\caption{Best results using a back-off technique.}
\label{taula-millors-K}
\end{table}
   
   The results here point to the same conclusions than the use of
trigrams: if we have a good trigram model (as in {\bf WSJ}) then the
back-off technique is useful. In this case, the result obtained with
the back--off model was better than the results for any
other constraint combination in this corpus.
 If the trigram model is not so good, results are not
better than the obtained with bigrams alone.

\subsection{Experiment conclusions}
\label{sec:selection-conclusions}

 The main conclusions of those experiments were the following:
\begin{itemize}
\item The most suitable support function is that described in equation \ref{support-s}.
 This is an expectable result, since this is the additive formula for computing 
support. Since zero compatibility constraints will be usual in our application 
--there may be many phenomena not described by our constraints, or that did not
occur in the training set-- a multiplicative formula would have the effect of 
making the support drop to zero when, for instance, a non-observed bigram 
was found. This makes the additive formula much a more logical choice, and this
intuition is confirmed by the experiments.
\item The alternative support function proposed in section \ref{sec:specific-supp} 
does not produce better results than the 
others. Although trying to simulate a bigram HMM with relaxation algorithms
could be an appealing idea --since then we would have a generalization
of the Markovian taggers which could be improved easily adding higher 
order information-- The already existing support functions seem to
combine the different kinds of constraints in a more efficient way. 
Nevertheless, we tried only one proposal, and this is still an open
issue.
\item The better results are obtained when modelling compatibility as mutual 
information. This is probably caused by the fact that mutual information can be
negative or positive, thus, it enables modelling {\em incompatibility} as well as
{\em compatibility}.
\item The updating function which experiments pointed out as the best choice was 
the zero-centered function, described
in equation \ref{updating-c}, but this is a
secondary effect of choosing mutual information as compatibility values, which 
requires an updating function able to deal with negative support values.
\item None of the tested stopping criteria performed significantly
 better than the others, nor than convergence.
\item The difference of $20$ errors (52 {\em vs}. 72, as described in section
\ref{sec:selection-stopping}) between the best iteration and
the convergence is not significant in a $50$ Kw corpus.
\item The hand analysis of the errors showed that most of the introduced errors
were due either to noise in the language model --caused by noise in the training
corpus-- or to noise in the test corpus itself, while most of the corrected
tags were real corrections. That changed the balance to a difference of 
some $20$ errors corrected by convergence that is also non-significant.
\item The experiments on finding the most suitable normalization factor for
support values showed that when the normalization factor is chosen in such
a way that the convergence stopping criterion produces its best results, the
difference between convergence and the best iteration is either zero or
non-significant. That is, the right normalization factor makes the optimal 
stopping iteration be that of convergence.
\item The most suitable normalization factor seems to be directly proportional
to the average support received by a word in the corpus. Although it has still
to be checked whether this proportionality depends on other factors such as
the used corpus, tag set, the ambiguity rate of the lexicon, etc. The procedure
currently used to establish this factor is the use of a part of the training
corpus as a tuning set, and choose as a normalization factor the value which
produces better results on the tuning set.
\end{itemize}

\section{Experiments on Part-of-Speech Tagging}
\label{sec:experiments-POS}

  Being the part-of-speech tagging task a basic one in natural language processing,
it has been addressed for long and from a range of approaches, from the early
linguistic--knowledge based work in \cite{Greene71}, to many different statistical 
approaches \cite{Garside87,Church88,Cutting92}. Great improvements have been done
from the seventies, but almost all systems still have about a $3\%$ error rate. 
The best currently performing system is that of \cite{Karlsson95,Voutilainen95}, which
achieves over a $99\%$ recall, although it does not fully disambiguate all words.

  Comparison between systems is difficult, since most of them use different
test corpora and different tagsets. Choosing an appropriate tagset is a 
crucial issue: if the tagset is too coarse, it would provide an excessively poor
information. If the tagset is too fine--grained, the tagger precision will
be much lower, because the model will be worse estimated (since much more
training data are needed to estimate a finer--grained model), and because some
ambiguities can not be solved on syntactic or context information only.

  In order to minimize the need for tagged data, several researchers
as \cite{Cutting92,Elworthy94a,Briscoe94a,Sanchez95}, use an initial
model --either hand build or estimated from a small tagged corpus--,
which is further refined using non-tagged data with the Baum-Welch
algorithm. \cite{Briscoe94a} applied this technique to tag different languages 
and tagsets, and conclude that a model acquired from relatively small tagged corpus
can be improved up to a reasonably good model through re-estimation. \cite{Elworthy94a} 
studies in which cases is worth using this technique, and how good will be the 
obtained models depending on the re-estimation starting point. He concludes that
although it is possible to obtain a fairly good model through re-estimation, the use
of as much tagged data as possible is the best policy to obtain accurate
n-gram models.
\medskip

  In the case of relaxation labelling the importance of the tagset size 
is relative, since the constraints are not required to be pure n-grams. 
They can be written in a coarser level than those of tags. For instance,
if tags include information about category, number and gender, the 
used constraints may take into account only category, or a finer grained 
distinctions depending on the case. With respect to model re-estimation, 
relaxation labelling can obviously use re-estimated models, but this is a point
that loses relevance as more and more tagged corpora become available. Moreover,
the interest of an algorithm such as relaxation labelling is the ability to use
complex constraint models, so there is no point in using it with simple models
that are more efficiently applied by Markovian taggers.
\medskip

  The experiments performed on POS tagging described in this section were used 
--once the most appropriate algorithm parameterization had been selected-- to 
check that the POS tagging task was accurately performed by our system, and that it
properly combines constraints from multiple sources.

  The experiments consisted of tagging the same corpus with different language models:
a bigram model, a trigram model, an automatically acquired decision-tree model, and a 
small set of hand-written constraints. These different models were combined to 
check whether their collaboration improved the separately obtained results.
\medskip

  The constraint acquisition procedure  has been exposed in 
section \ref{sec:decision-trees}. For further information 
on this topic see \cite{Marquez97a,Marquez97b}.
\medskip

\subsection{Corpus description}
\label{sec:experiments-POS-corpus}

 We used the Wall Street Journal corpus to train and test the system.
We divided it in three parts: $1,100$ Kw were used as a training set,
$20$ Kw as a model--tuning set, and $50$ Kw as a test set.
 
 The tag set size is 48 tags\footnote{See appendix \ref{app:tagsets} for
                                      a detailed listing.}. 
$36.4\%$ of the words in the corpus are ambiguous, and the ambiguity ratio 
is $2.45$ tags/word over the ambiguous words, $1.52$ overall.
\medskip

 We used a lexicon derived from training corpora, that contains
all possible tags for a word, as well as their lexical probabilities.
For the words in test corpora not appearing in the train set, we stored all
possible tags, but no lexical probability (i.e. we assume uniform
distribution)\footnote{That is, we assumed a
morphological analyzer that provides all possible tags for unknown
words.}.

 The noise in the lexicon was filtered by manually checking the lexicon entries
for the 200 most frequent words in the corpus\footnote{The 200 most frequent words
in the corpus cover over half of it.} to eliminate the tags due to
errors in the training set.
For instance the original lexicon entry 
(numbers indicate frequencies in the training corpus)
for the very common word {\it the} was
\medskip

\noindent\hfil{\tt the: CD 1, DT 47715, JJ 7, NN 1, NNP 6, VBP 1.}\hfil
\medskip

\noindent since it appears in the corpus 
with the six different tags: CD (cardinal), DT (determiner), JJ
(adjective), NN (noun), NNP (proper-noun) and VBP (verb:personal-form). 
It is obvious that the only correct reading for {\it the} is
determiner.
\medskip

 The training set was used to estimate bi/trigram statistics and to
 perform the constraint learning.

 The model--tuning set was used to tune the algorithm
parameterizations, and to write the linguistic part of the model. 

 The resulting models were tested in the fresh test set.

\subsection{Language model}

We will use a hybrid language model consisting of
an automatically acquired part and a linguist--written part.
\medskip

The automatically acquired part is divided in two kinds of
information: 
\medskip

 On the one hand, we have bigrams and trigrams collected
from the annotated training corpus: we obtained 1404 bigram 
restrictions and 17387 trigram restrictions from the 
training corpus.
\medskip

 On the other hand, we have context constraints learned
from the same training corpus using statistical decision trees
acquired for each representative ambiguity class.

 The whole WSJ corpus contains 241 different classes of ambiguity.
The 40 most representative classes\footnote{In
terms of number of examples.} were selected for acquiring the
corresponding decision trees.
That produced 40 trees totaling up to 2995 leaf nodes, and covering 
83.95\% of the ambiguous words. Given that
each tree branch produces as many constraints as tags its leaf
involves, these trees were translated into 8473 context constraints.
\medskip

 The linguistic part is very small --since there were no available 
resources to develop it further-- and covers only very few cases, but it
is included to illustrate the flexibility of the algorithm.
It was written as follows: the model--tuning set was tagged using a bigram model.
Then, the most common errors made by the bigram tagger were selected, and
some 20 constraints were manually written to cover those cases.
\medskip

A sample rule of the linguistic part is the following:
\begin{tabbing} xxxxxxxx \= xxxx \= xxxxxxxxxxxxxxxxx \= \kill
\> {\tt 10.0} \> {\tt (VBN)} \\
\>            \> {\tt (*-1 VAUX BARRIER (VBN) OR (IN) OR (<,>) OR} \\
\>            \>                \> {\tt (<:>) OR (JJ) OR (JJS) OR (JJR));}
\end{tabbing}

  This rule states that a tag {\em past participle} ({\bf VBN}) is very
compatible (10.0) with a left context consisting of a {\bf VAUX} 
(previously defined macro which includes all forms of ``have'' and ``be'') 
provided that all the words in between do not have any of the tags in the set
{\bf\{VBN IN $<$,$>$ $<$:$>$ JJ JJS JJR\}}. That is, this rule raises the support for
the tag {\em past participle} when there is an auxiliary verb to the left but
only if there is not another candidate to be a past participle or an 
adjective in-between. The tags {\bf\{IN $<$,$>$ $<$:$>$\}} prevent the rule from being
applied when the auxiliary verb and the participle are in two different
phrases (a comma, a colon or a preposition are considered to mark the 
beginning of another phrase).

  The constraint language used in this example is
the Constraint Grammar formalism \cite{Karlsson95}, with
the additional feature of an unrestricted numerical weight for each 
constraint, instead of only two possible values (SELECT/REMOVE).
\medskip

\subsection{Experiment description and results}
\label{sec:experiments-POS-results}

 Once the different language models had been obtained,
the tagger was tested on the $50$ Kw test set using all the
possible combinations of the models.
\medskip

 As a detailed example of the model behaviour, the effect of the acquired 
rules on the number of errors for some of the most common cases is shown 
in table \ref{taula:experiments-POS-separats}\footnote{XX/YY stands for an 
                                                       error consisting of a word 
                                                       tagged YY when it should 
                                                       have been XX. The meaning 
                                                       of the involved tags
                                                       can be found in appendix 
                                                       \ref{app:tagsets}.}.
\medskip

 In the tables presented in this section, {\bf C} stands for 
the acquired context constraints, {\bf B} for
the 1404--bigram model, {\bf T} for the 17387--trigram model, and 
{\bf H} for a small set of 20 hand-written constraints.
In addition, {\bf ML} indicates a baseline model containing no constraints
(this will result in a most-likely tagger) and {\bf HMM} stands for a
hidden Markov model bigram tagger \cite{Elworthy93}.
\medskip

{\scriptsize
\begin{table}[htb] \centering
\begin{tabular}{|l|r||r||r|r||r|r||r|r|} \hline
                 &ML      &C     &B      &BC     &T      &TC    &BT     &BTC   \\ \hline
JJ/NN+NN/JJ      &73+137  &70+94 &73+112 &69+102 &57+103 &61+95 &67+101 &62+93 \\ \hline
VBD/VBN+VBN/VBD  &176+190 &71+66 &88+69  &63+56  &56+57  &55+57 &65+60  &59+61 \\ \hline
IN/RB+RB/IN      &31+132  &40+69 &66+107 &43+17  &77+68  &47+67 &65+98  &46+83 \\ \hline
VB/VBP+VBP/VB    &128+147 &30+26 &49+43  &32+27  &31+32  &32+18 &28+32  &28+32 \\ \hline 
NN/NNP+NNP/NN    &70+11   &44+12 &72+17  &45+16  &69+27  &50+18 &71+20  &62+15 \\ \hline
NNP/NNPS+NNPS/NNP&45+14   &37+19 &45+13  &46+15  &54+12  &51+12 &53+14  &51+14 \\ \hline
``that''         &187     &53    &66     &45     &60     &40    &57     &45    \\ \hline \hline
Total            &1341    &631   &820    &630    &703    &603   &731    &651   \\ \hline
\end{tabular}
\caption{Number of some common errors made by each model.}
\label{taula:experiments-POS-separats}
\end{table}
}

 Figures in table \ref{taula:experiments-POS-separats} show that in 
all cases the extension of a statistical model with 
the machine--learned constraints led to a reduction in the number
of errors.

  It is remarkable that when using {\bf C} alone, the
number of errors for these cases is lower than with any bigram and/or 
trigram model, that is, the acquired model performs better than the 
others estimated from the same training corpus.
\medskip

  The global results on the test corpus obtained by the baseline taggers 
can be found in table \ref{taula:experiments-POS-resultats-base}
and the results obtained using all the learned constraints together 
with the bi/trigram models in table \ref{taula:experiments-POS-resultats-auto}.

\begin{table}[htb] \centering
\begin{tabular}{|l|r|r|} \hline
     &ambiguous &overall    \\ \hline \hline
ML   &$85.31\%$ &$94.66\%$  \\ \hline
HMM  &$91.75\%$ &$97.00\%$  \\ \hline
\end{tabular}
\caption{Results of the baseline taggers.}
\label{taula:experiments-POS-resultats-base}
\end{table}

\begin{table}[htb] \centering
\begin{tabular}{|l|r|r|} \hline
     &ambiguous &overall    \\ \hline \hline
B    &$91.35\%$ &$96.86\%$  \\ \hline
T    &$91.82\%$ &$97.03\%$  \\ \hline
BT   &$91.92\%$ &$97.06\%$  \\ \hline \hline
C    &$91.96\%$ &$97.08\%$  \\ \hline
BC   &$92.72\%$ &$97.36\%$  \\ \hline
TC   &$92.82\%$ &$97.39\%$  \\ \hline
BTC  &$92.55\%$ &$97.29\%$  \\ \hline
\end{tabular}
\caption{Results of our tagger using every combination of constraint kinds.}
\label{taula:experiments-POS-resultats-auto}
\end{table}

On the one hand, the results in tables \ref{taula:experiments-POS-resultats-base} 
and \ref{taula:experiments-POS-resultats-auto} show that our tagger
performs slightly worse than a HMM tagger in the same 
conditions\footnote{Hand analysis of the errors made by the algorithm
                    suggest that the worse results may be due to noise in 
                    the training and test corpora, i.e., relaxation 
                    algorithm seems to be more noise--sensitive than a 
                    Markov model. Further research is required on this point.}, 
that is, when using only bigram information.

On the other hand, those results also show that since our tagger is 
more flexible than a HMM, it can easily accept more complex information to 
improve its results up to $97.39\%$ without modifying the algorithm.
\medskip

 Table \ref{taula:experiments-POS-resultats-manuals} shows the results 
adding the hand written constraints. 
 The hand written set is very small and only covers
a few common error cases. That produces poor results when using them
alone ({\bf H}), but they are good enough to raise the results given by the
automatically acquired models up to $97.45\%$.

\begin{table}[ht] \centering
\begin{tabular}{|l|r|r|} \hline
      &ambiguous &overall    \\ \hline \hline
H     &$86.41\%$ &$95.06\%$  \\ \hline
BH    &$91.88\%$ &$97.05\%$  \\ \hline
TH    &$92.04\%$ &$97.11\%$  \\ \hline
BTH   &$92.32\%$ &$97.21\%$  \\ \hline \hline
CH    &$91.97\%$ &$97.08\%$  \\ \hline
BCH   &$92.76\%$ &$97.37\%$  \\ \hline
TCH   &$92.98\%$ &$97.45\%$  \\ \hline
BTCH  &$92.71\%$ &$97.35\%$  \\ \hline
\end{tabular}
\caption{Results of our tagger using every combination of constraint kinds and hand written constraints.}
\label{taula:experiments-POS-resultats-manuals}
\end{table}

  Although the improvement obtained might seem small, the difference 
is statistically significant when the decision--tree model is incorporated
to any n-gram model. In addition, it must be taken
into account that we are moving very close to the best achievable
result with the current techniques and resources. This item is further 
discussed in section \ref{sec:results-POS-errors}.

\section{Experiments on other NLP tasks}
\label{sec:experiments-NLP}

  The second group of experiments consisted of applying the algorithm to 
different NLP tasks other than POS tagging. The experiments presented in
section \ref{sec:experiments-POS} had shown 
that the performance on POS tagging is at least as good as that of current 
statistical taggers, and that the relaxation algorithm is able to combine
constraints obtained from different knowledge sources.
\medskip 

 The set of experiments described in this section was used to test 
whether the algorithm could easily cope with constraints on features other
than part-of-speech and perform other disambiguation tasks, as well as
its ability to simultaneously disambiguate more than one feature.

\subsection{Shallow Parsing}
\label{sec:experiments-ShP}

   The use of language models based on context constraints has a successful representative 
in the Constraint Grammar formalism \cite{Karlsson95} and related work 
\cite{Voutilainen95,Samuelson96,Samuelson97}. They employ only 
constraints written by linguists and successively refined through the 
use of real text corpora.
\medskip

   Since our system also deals with context constraint models, we set up a collaboration
to test a hybrid model, where hand written context constraints could cooperate
with statistically acquired constraints, such as bigrams or trigrams. This would enable
us to compare the performances of a purely linguistic model with a purely statistical
one, and also to check whether they can collaborate to produce better results.

   Those experiments were performed on shallow parsing, and consisted of analyzing a test
corpus with different models and algorithms. The algorithms were the CG-2 Constraint Grammar 
environment \cite{Tapanainen96} and the relaxation labelling algorithm. The language 
models are: a linguist-written language model, the bi/trigram models and all possible 
combinations of them. Since the CG-2 environment is not able to deal with statistical 
information, it will only be used with the linguist-written model. The statistical and
hybrid models will be applied with relaxation labelling.
This work has been published in \cite{Voutilainen97}.

\subsubsection{Setting}
\label{experiments-ShP-setting}
Most hybrid approaches combine statistical information with
automatically extracted rule-based information \cite{Brill95,Daelemans96a}.
Relatively little attention has been paid
to models where the statistical approach is combined with a truly
linguistic model (i.e.\ one generated by a linguist).  This experiment
is based on one such approach: syntactic rules written by a linguist are
combined with statistical information using the relaxation labelling
algorithm.
\medskip

\begin{figure}[htb]
\hfil\epsfbox{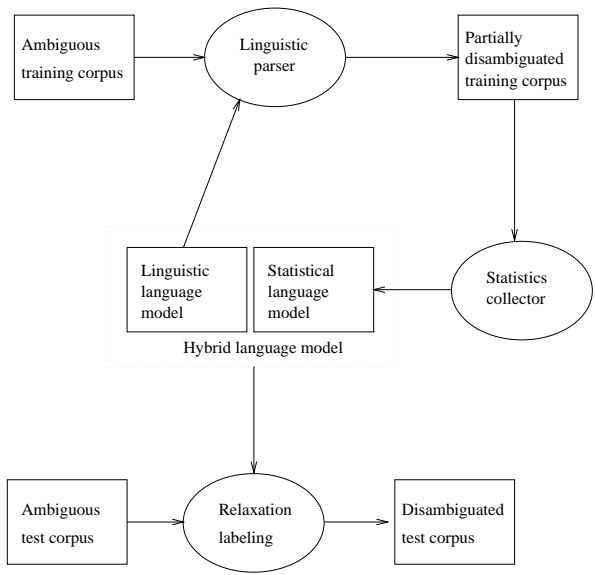}\hfil
\caption{Parser architecture.}\label{fig:parser}
\end{figure}

 In this case, the application is very shallow parsing: identification of verbs,
premodifiers, nominal and adverbial heads, and certain kinds of
postmodifiers. We call this parser a noun phrase parser. 
The system architecture is presented in figure \ref{fig:parser}, 
and combines two approaches:
\begin{itemize}
\item[(i)] a linguistic language model which is used 
       as a model to parse the test corpus as well as a model to disambiguate
       the training corpus and thus obtain a source of almost--supervised
       knowledge to acquire statistical models from. 
\item[(ii)] two n-gram statistical language models --namely, bigram and trigram--
       acquired from the aforementioned training corpus.
\end{itemize}

The input is English text morphologically tagged with a rule-based
tagger called EngCG \cite{Voutilainen92,Karlsson95}. Syntactic
word-tags --described below-- are added as alternatives (e.g. each adjective gets a
premodifier tag, postmodifier tag and a nominal head tag as
alternatives). The system should remove contextually illegitimate tags
and leave intact each word's most appropriate tag. In other words, the
syntactic language model is applied by a disambiguator.

The parser has a {\em recall} of 100\% if all words retain the correct
morphological and syntactic reading; the system's {\em precision} is
100\% if the output contains no illegitimate morphological or
syntactic readings. In practice, some correct analyses are discarded,
and some ambiguities remain unresolved.
\medskip

The system can use linguistic rules and corpus-based statistics.
Notable about the system is that minimal human effort was needed for
creating its language models (the linguistic consisting of syntactic
disambiguation rules based on the Constraint Grammar framework
\cite{Karlsson90,Karlsson95}; the corpus-based consisting of bigrams
and trigrams):

\begin{itemize}
\item Only one day was spent on writing the 107 syntactic
    disambiguation rules used by the linguistic parser.
\item No human annotators were needed for annotating the training corpus
    (218,000 words of journalese) used by the data-driven learning
    modules of this system: the training corpus was annotated by the following 
    procedure:
    \begin{enumerate}
    \item It was tagged using the EngCG morphological tagger.
    \item The tagged text was made syntactically ambiguous by adding
    the alternative syntactic tags to the words.
    \item Finally, the syntactic ambiguities were solved by applying 
    the parser with the 107 disambiguation rules.
    \end{enumerate}
\end{itemize}
\medskip

\noindent The system was tested against a fresh sample of five texts (6,500
words). The system's recall and precision was measured by comparing
its output to a manually disambiguated version of the text. 
 Recall is the percentage of words that get the correct
tag among the tags proposed by the system. Precision is the percentage
of tags proposed by the system that are correct.

Also the relative contributions of the linguistic and statistical
components were evaluated. The linguistic rules seldom discard the
correct tag, i.e.\ they have a very high recall, but their problem is
remaining ambiguity. The problems of the statistical components are
the opposite: their recall is considerably lower, but more (if not
all) ambiguities are resolved. When these components are used in a
balanced way, the system's overall recall is $97.2\%$ -- that is,
$97.2\%$ of all words get the correct analysis -- and its precision
is $96.1\%$ -- that is, of the readings returned by the system,
$96.1\%$ are correct.

\subsubsection{Grammatical representation}

The input of the parser is morphologically analyzed and disambiguated
text enriched with alternative syntactic tags, e.g.

\begin{verbatim}
"<others>"
   "other" PRON NOM PL @>N @NH
"<moved>"
   "move" <SV> <SVO> V PAST VFIN @V
"<away>"
   "away" ADV ADVL @>A @AH
"<from>"
   "from" PREP @DUMMY
"<traditional>"
   "traditional" A ABS @>N @N< @NH
"<jazz>"
   "jazz" <-Indef> N NOM SG @>N @NH
"<practice>"
   "practice" N NOM SG @>N @NH
   "practice" <SVO> V PRES -SG3 VFIN @V
\end{verbatim}

Every indented line represents a morphological analysis.
Syntactic tags start with the "@" sign. A word is syntactically
ambiguous if it has more than one syntactic tags (e.g. {\it practice}
above has three alternative syntactic tags). The above sample
shows that some morphological ambiguities are not resolved by the
rule-based EngCG morphological disambiguator.
\medskip

Next we describe the syntactic tags:

\begin{itemize}
\item @$>$N represents premodifiers and determiners.
\item @N$<$ represents a restricted range of postmodifiers and the
determiner "enough" following its nominal head.
\item @NH represents nominal heads (nouns, adjectives, pronouns,
  numerals, ING-forms and non-finite ED-forms).
\item @$>$A represents those adverbs that premodify (intensify) adjectives
(including adjectival ING-forms and non-finite ED-forms), adverbs and
various kinds of quantifiers (certain determiners, pronouns and
numerals).
\item @AH represents adverbs that function as head of an adverbial phrase.
\item @A$<$ represents the postmodifying adverb "enough".
\item @V represents verbs and auxiliaries (including the infinitive marker
"to").
\item @$>$CC represents words introducing a coordination ("either",
"neither", "both").
\item @CC represents coordinating conjunctions.
\item @CS represents subordinating conjunctions.
\item @DUMMY represents all prepositions, i.e. the parser does not
  address the attachment of prepositional phrases.
\end{itemize}

\subsubsection{Syntactic rules}
\medskip

\noindent\underline{\bf Rule formalism}
\nopagebreak\medskip

\noindent The rules follow the Constraint Grammar formalism, and they were
applied using the recent parser-compiler CG-2 \cite{Tapanainen96}. The
parser reads a sentence at a time and discards those
ambiguity-forming readings that are disallowed by a constraint.
\medskip

Next we describe some basic features of the rule formalism. The rule
\begin{tabbing} xxxxxxxx \= xxxxxxxx \= \kill
\> {\tt REMOVE} \> {\tt (@>N)} \\
\>              \> {\tt (*1C <<< OR (@V) OR (@CS) BARRIER (@NH));}
\end{tabbing}

\noindent removes the premodifier tag @$>$N from an ambiguous reading if
somewhere to the right (*1) there is an unambiguous (C) occurrence of
a member of the set $<<<$ (sentence boundary symbols) or the verb tag
@V or the subordinating conjunction tag @CS, and there are no
intervening tags for nominal heads (@NH).  \medskip

This is a partial rule about coordination:
\begin{tabbing} xxxxxxxx \= xxxxxxxx \= \kill
\> {\tt REMOVE} \> {\tt (@>N)} \\
\>              \> {\tt (NOT 0 (DET) OR (NUM) OR (A))} \\
\>              \> {\tt (1C (CC))} \\
\>              \> {\tt (2C (DET));}
\end{tabbing}

\noindent It removes the premodifier tag if all three context-conditions are satisfied:

\begin{itemize}
\item the word to be disambiguated (0) is not a determiner, numeral or adjective,
\item the first word to the right (1) is an unambiguous coordinating
  conjunction, and
\item the second word to the right is an unambiguous determiner.
\end{itemize}

The rules can refer to words and tags directly or by means of
predefined sets. They can refer not only to any fixed context
positions; also reference to contextual patterns is possible.  The
rules never discard a last reading, so every word retains at least one
analysis. On the other hand, an ambiguity remains unresolved if there
are no rules for that particular type of ambiguity.
\medskip

\noindent\underline{\bf Grammar development}
\nopagebreak\medskip

\noindent A day was spent on writing 107 constraints; about 15,000 words of the
parser's output were proof-read during the process. The routine was the
following:

\begin{enumerate}
\item The current grammar (containing e.g.\ 2 rules) is applied to the
    ambiguous input in a `trace' mode in which the parser also indicates,
    which rule discarded which analysis,
\item The grammarian observes remaining ambiguities and proposes new rules
    for disambiguating them, and
\item He also tries to identify misanalyses (cases where the correct
    tag is discarded) and, using the trace information, corrects the
    faulty rule.
\end{enumerate}

This routine is useful if the development time is very restricted, and
only the most common ambiguity types have to be resolved with
reasonable success.  However, if the grammar should be of a very high
quality (extremely few mispredictions, high degree of ambiguity
resolution), a large test corpus, formally similar to the input except
for the manually added extra information about the correct analysis,
should be used. This kind of test corpus would enable the automatic
identification of mispredictions as well as counting of various
performance statistics for the rules. However, manually disambiguating
a test corpus of a few hundred thousand words would probably require a
human effort of at least a month.
\medskip

\noindent\underline{\bf Sample output}
\nopagebreak\medskip

\noindent The following is genuine output of the linguistic (CG-2) parser using
the 107 syntactic disambiguation rules. The traces starting with "S:"
indicate the line on which the applied rule is in the grammar file.
One syntactic (and morphological) ambiguity remains unresolved: {\it
  until} remains ambiguous due to preposition and subordinating
conjunction readings.

\begin{verbatim}
"<aachen>" S:46
   "aachen" <*> <Proper> N NOM SG @NH
"<remained>"
   "remain" <SVC/N> <SVC/A> V PAST VFIN @V
"<a>"
   "a" <Indef> DET CENTRAL ART SG @>N
"<free>" S:316, 49
   "free" A ABS @>N
"<imperial>" S:49, 57
   "imperial" A ABS @>N
"<city>" S:46
   "city" N NOM SG @NH
"<until>"
   "until" PREP @DUMMY
   "until" <**CLB> CS @CS
"<occupied>" S:116, 345, 46
   "occupy" <SVO> PCP2 @V
"<by>"
   "by" PREP @DUMMY
"<france>" S:46
   "france" <*> <Proper> N NOM SG @NH
"<in>"
   "in" PREP @DUMMY
"<1794>" S:121, 49
   "1794" <1900> NUM CARD @NH
"<$.>"
\end{verbatim}

\subsubsection{Hybrid language model}
\label{experiments-ShP-hybrid}

  To solve shallow parsing with the relaxation labelling algorithm
we model each word in the sentence as a variable, and each of its
possible readings as a label for that variable. We start with a
uniform weight distribution.

   We will use the algorithm to select the right syntactic tag for
every word. Each iteration will increase
the weight for the tag which is currently most compatible with the
context and decrease the weights for the others. 

   Since constraints are used to decide {\em how compatible} a tag is with
its context, they have to assess the compatibility of a combination of
readings. We adapt CG constraints described above.
\medskip

  The {\bf REMOVE} constraints express total
incompatibility\footnote{We model compatibility values
                         using mutual information \cite{Cover91},
                         which enables us to use negative numbers 
                         to state {\em incompatibility}.}  
and {\bf SELECT} constraints express total
compatibility (actually, they express incompatibility of all
other possibilities).

  The compatibility value for these should be at least as strong as
the strongest value for a statistically obtained constraint (see below), 
which happens to be about $\pm 10$. 

  But because we want the linguistic part of the model to be more
important than the statistical part and because a given label will
receive the influence of about two bigrams and three
trigrams\footnote{The algorithm tends to select one label per
                  variable, so there is always a bi/trigram which is 
                  applied more significantly than the others.}, 
a single linguistic constraint
might have to override five statistical constraints. So we will
make the compatibility values for linguistic rules six times stronger
than the strongest statistical constraint, that is, $\pm 60$.
\medskip

  Since in our implementation of the CG parser \cite{Tapanainen96}
constraints tend to be applied in a certain order -- e.g.\ {\bf SELECT} 
constraints are usually applied {\em before} {\bf REMOVE}
constraints -- we adjust the compatibility values to get a similar
effect: if the value for {\bf SELECT} constraints is $+60$, the
value for {\bf REMOVE} constraints will be lower in absolute value,
(i.e.\ $-50$). With this we ensure that two contradictory
constraints (if there are any) do not cancel each other.  
The {\bf SELECT} constraint will win, as if it had been applied before.
\medskip

  This enables using any Constraint Grammar with this algorithm although
we are applying it more flexibly: we do not decide whether a
constraint is applied or not. It is always applied with an influence
(perhaps zero) that depends on the weights of the labels.

  If the algorithm should apply the constraints in a more strict way, we
can introduce an influence threshold under which a constraint does not
have enough influence, i.e. it is not applied.
\medskip

  We can add more information to our model in the form of
statistically derived constraints. Here we use bigrams and
trigrams as constraints.

  The 218,000-word corpus of
journalese from which these constraints were extracted was
build as described in section \ref{experiments-ShP-setting}.

  It is noticeable that no human effort was spent on creating 
this training corpus.
The training corpus is partly ambiguous, so the
bi/trigram information acquired will be slightly noisy, but
accurate enough to provide an {\em almost} supervised statistical model.
 
  For instance, the following constraints have been statistically
extracted from bi/trigram occurrences in the training corpus. 

\begin{tabbing} xxxxxxxx \= xxxxxxxx \= xxxxxxxxxxxxxxxxxxxxxx \= xxxxxxx \= \kill
\> {\tt -0.4153} \> {\tt (@V)}        \> {\tt 4.2808} \> {\tt (@>A)} \\
\>                 \> {\tt (1 (@>N));}  \>               \> {\tt (-1 (@>A))} \\
\>                 \>                   \>               \> {\tt (1 (@AH));}
\end{tabbing}

   The compatibility value assigned to these constraints
is the mutual information between the affected syntactic tag and
the context described by the constraint. It is computed from the probabilities
estimated from the training corpus. We do not need to manually assign the
compatibility values here, since we can estimate them from the corpus.
\medskip

   The compatibility values assigned to the hand--written constraints
express the strength of these constraints compared to the
statistical ones. Modifing those values means changing the relative
weights of the linguistic and statistical parts of the model.

\subsubsection{Preparation of the benchmark corpus}

  For evaluating the systems, five roughly equal-sized benchmark corpora
not used in the development of our parsers and taggers were prepared.
The texts, totaling 6,500 words, were copied from the Gutenberg e-text
archive, and they represent present-day American English. One text is
from an article about AIDS; another concerns brainwashing techniques;
the third describes guerilla warfare tactics; the fourth addresses the
assassination of J.~F.~Kennedy; the last is an extract from a speech
by Noam Chomsky.

The texts were first analysed by a recent version of the morphological
analyser and rule-based disambiguator EngCG, then the syntactic
ambiguities were added with a simple lookup module. The ambiguous text
was then manually disambiguated. The disambiguated texts were also
proof-read afterwards.  Usually, this practice resulted in one analysis
per word. However, there were two types of exception:

\begin{enumerate}
\item The input did not contain the desired alternative (due to a
   morphological disambiguation error). In these cases, no reading was
   marked as correct. Two such words were found in the corpora; they
   detract from the performance figures.
\item The input contained more than one analyses all of which seemed
   equally legitimate, even when semantic and textual criteria were
   consulted. In these cases, all the equal alternatives were marked
   as correct. The benchmark corpus contains 18 words (mainly ING-forms and
   nonfinite ED-forms) with two correct syntactic analyses.
\end{enumerate}

The number of multiple analyses could probably be made even smaller by
specifying the grammatical representation (usage principles of the
syntactic tags) in more detail, in particular incorporating some
analysis conventions for certain apparent borderline cases (for a
discussion of specifying a parser's linguistic task, see
\cite{Voutilainen95}).

\subsubsection{Experiments and results}

We tested linguistic, statistical and hybrid language models, using
the CG-2 parser described in \cite{Tapanainen96} and the relaxation 
labelling algorithm.
\medskip

The statistical models were obtained from a training corpus of 218,000
words of journalese, syntactically annotated using the linguistic
parser (section \ref{experiments-ShP-hybrid}).  

Although the linguistic CG-2 parser does not disambiguate completely,
it seems to have an almost perfect recall (Table \ref{taula:experiments-ShP-L}),
and the noise introduced by the remaining ambiguity is assumed to be
sufficiently lower than the signal, following the idea used in
\cite{Yarowsky92}.

The collected statistics were bigram and trigram occurrences.
\medskip

The algorithms and models were tested against the above described
hand--disambiguated benchmark corpus.
\medskip

 Models are coded as follows: {\bf B} stands for
bigrams, {\bf T} for trigrams and {\bf C} for hand--written
constraints. All combinations of information types are tested. Since
the CG-2 parser handles only Constraint Grammars, we can not test this
algorithm with statistical models.
\medskip

\begin{table}[hbt] \centering
\begin{tabular}{|l|c|c|} \hline
           &{\bf CG-2 parser}          &{\bf Relaxation labelling} \\
           &{\bf precision - recall} &{\bf precision - recall} \\ \hline
   {\bf C} &$90.8\%-99.7\%$       &$93.3\%-98.4\%$        \\ \hline
   {\bf forced-C}&$95.0\%-95.0\%$ &$95.8\%-95.8\%$     \\ \hline
\end{tabular}
\caption{Results obtained with the linguistic model.}
\label{taula:experiments-ShP-L}
\end{table}

    Table \ref{taula:experiments-ShP-L} summarizes the results obtained when using 
only a linguistic model. Results are given in precision and recall since the
model does not disambiguate completely. Results when forcing complete disambiguation
through random selection are also presented.

\begin{table}[htb] \centering
\begin{tabular}{|l|c|} \hline
            &{\bf Relaxation labelling}  \\
            &{\bf precision - recall}  \\ \hline
   {\bf B}  &$87.4\%-88.0\%$     \\ \hline
   {\bf T}  &$87.6\%-88.4\%$     \\ \hline
   {\bf BT} &$88.1\%-88.8\%$     \\ \hline
  {\bf forced-BT}&$88.5\%-88.5\%$      \\ \hline
\end{tabular}
\caption{Results obtained with statistical models.}
\label{taula:experiments-ShP-S}
\end{table}

    Table \ref{taula:experiments-ShP-S} shows the results given by the statistical 
models, which are rather worse, since shallow parsing is a task more difficult to
capture in a n-gram model than POS tagging.

\begin{table}[hbt] \centering
\begin{tabular}{|l|c|} \hline
             &{\bf Relaxation labelling} \\
             &{\bf precision - recall} \\ \hline
  {\bf BC}   &$96.0\%-97.0\%$    \\ \hline
  {\bf TC}   &$95.9\%-97.0\%$    \\ \hline
  {\bf BTC}  &$96.1\%-97.2\%$    \\ \hline
  {\bf forced-BTC}&$96.7\%-96.7\%$      \\ \hline
\end{tabular}
\caption{Results obtained with hybrid models.}
\label{taula:experiments-ShP-H}
\end{table}

    Finally, table \ref{taula:experiments-ShP-H} presents the results produced 
by the hybrid models, which are significantly better than the previous ones, that is,
the collaboration between models improved the performance in this case as well
as in POS-tagging (section \ref{sec:experiments-POS}).
\medskip

These results suggest the following conclusions: 

\begin{itemize}
\item Using the same language model (107 rules), the relaxation
  algorithm disambiguates more than the CG-2 parser.  This is due to
  the weighted rule application, and results in more misanalyses and
  less remaining ambiguity.
\item The statistical models are clearly worse than the linguistic
  one. This could be due to the noise in the training corpus, but it
  is more likely caused by the inherent difficulty of the task: we are dealing
  here with shallow syntactic parsing, which is probably more
  difficult to capture in a statistical model than POS tagging.
\item The hybrid models produce less ambiguous results than the other
  models. The number of errors is much lower than was the case with
  the statistical models, and somewhat higher than was the case with
  the linguistic model. The gain in precision seems to be enough to
  compensate for the loss in recall, although, obviously, this depends on
  the flexibility of one's requirements.
\item There does not seem to be much difference between BC and TC
  hybrid models. The reason is probably that the job is mainly done by
  the linguistic part of the model -- which has a higher relative
  weight -- and that the statistical part only helps to disambiguate
  cases where the linguistic model does not make a prediction.  The BTC
  hybrid model is slightly better than the other two.
\item The small difference between the hybrid models suggest
  that some reasonable statistics provide enough disambiguation, and
  that not very sophisticated information is needed.
\end{itemize}

\subsection{Word Sense Disambiguation}
\label{sec:experiments-WSD}

   The utility of the constraint-based models applied through relaxation labelling
algorithms was also checked in the task of word sense disambiguation. Nevertheless,
the work described in this section is in an early stage and the obtained results 
have still to be improved.
Factors that affect this part of the work are, apart from the intrinsic difficulty 
of the task, the lack of large sense-tagged corpus to perform training --we use 
SemCor \cite{Miller93,Miller94}, which contains only some $230$ Kwords--
and the difficulty to obtain accurate context constraints involving word senses. 
\medskip

Since, as stated in \cite{Wilks96b,Wilks97},
knowing the part-of-speech tag for a word helps to reduce the sense
ambiguity  in a large amount of cases, we addressed the combined problem POS+WSD.
 The way in which this was performed was the following: we considered that
what is assigned to each word is not a single {\em tag} but a {\em reading}, 
being a reading a set of word features, that may include --among others-- 
part-of-speech tag, sense, lemma, etc.

   Then, the task of disambiguation consists of selecting the most appropriate 
reading for the current context, and this can be done through relaxation algorithms
if context constraints on the existing features are available. 

\subsubsection{Searching for appropriate semantic constraints}
\label{sec:experiments-WSD-constraints}

   Due to their higher complexity, context constraints on semantic features 
are more difficult to obtain than other kinds of models, such as statistical 
information for POS tagging. This difficulty is found not only in automatically 
acquired models --since the complexity of the task overwhelms most
acquisition algorithms, and not enough supervised data is available to
feed them--, but also in manually developed models, since the larger
number of items to deal with makes it a high labour cost task to manually
develop a linguistic model for WSD. 
\medskip

  In order to obtain semantic constraints to feed the relaxation algorithm with,
we tried to extract knowledge from different sources, and use them either
combined or separately.
 
  The used constraints included the following knowledge:
\begin{itemize}
\item POS bigrams, which will perform the part-of-speech disambiguation.
\item Most likely sense selection once the POS tag is known. The senses
      are considered to be output by WordNet sorted from most likely 
      to less likely. 
\item Pairwise conceptual distance \cite{Sussna93,Agirre95} among noun 
      senses, measured in the WordNet taxonomy \cite{Miller91}.
      These constraints try to capture the {\em topic} the sentence is about.

      Each pair of noun senses in WordNet generates a binary constraint,
      stating that they have a compatibility inversely proportional to
      their distance, so the nearer they are, 
      the more compatible\footnote{Obviously, for efficiency reasons, not
                            {\em all} the possible constraints
                            are generated {\em a priori}. The 
                            distance is computed only when a
                            pair of senses appears. That is, 
                            constraints are dinamically generated.}.
      This raises the support for noun senses that
      are neighbour in the WordNet taxonomy.

      These constraints do not consider the relative position of the words.
      This means that a noun sense is affected by as many constraints 
      as possible senses may have the nouns in its context --being the 
      context the whole sentence where it appears--, regardless of their
      relative position.

        Obviously, this approach assumes that the nouns appearing
      in the same sentence tend to have conceptually near senses.
      This maybe true in some --too obvious-- cases such as 
      {\tt The nurse helps the doctor at the hospital.}, but it also 
      may be misleading in many other cases --more likely to happen 
      in a real corpus--, such as {\tt The child felt sick and the nurse 
      had to take him to the hospital to see the doctor}.
 
        In addition, since WordNet consists of separate hierarchies for
      nouns and verbs, conceptual distance between nouns and verbs can not
      be computed. This prevents us from using this kind of constraints
      to detect the difference between sentences like 
      {\tt The crane ate the fish} and {\tt The crane lifted the fish container}.

\item  WordNet top synsets pairwise co-occurrences, interpreted as class 
      co-occurrences. These constraints try to be a kind of {\em semantic
      bigrams}. 
 
        Each top synset is considered as a class, and all its
      descendant synsets are considered to belong to that class. Then,
      class co-occurrences are computed on a training corpus, and used as
      binary constraints. As in the previous case, no positional 
      information is considered, two classes are considered to co-occur if 
      they appear in the same sentence, regardless of their position.
      Nevertheless, this approach enables deriving verb--noun constraints,
      since there will be co-occurrences of noun and verb classes. 
      Anyway, the flaw here is that verbs are organized in WordNet in a 
      very flat hierarchy, that is, most verbs are tops and constitute
      a class on their own. This produces a large number of possible
      verb--noun constraints, which require a much larger corpus to be estimated.

        This kind of constraints have also been acquired using WordNet 
      file codes instead of top synsets as class identifiers.

        The assumption that this approach requires is that senses 
      belonging to a given class tend to appear more with senses 
      of certain classes than with senses of the others.

\item Automatically acquired selectional restrictions on verb objects. The 
      acquisition procedure is described in \cite{Ribas94,Ribas95} and
      has been outlined in section \ref{sec:semantic}.

        Selectional restrictions try to capture the constraints that a
      phrase head imposes to its complements. In our case, we focused
      on the constraints that a verb imposes to its objects. 

        The restrictions are acquired in such a way that for each verb,
      a numerical value (probability, association ratio, \ldots) is assigned
      to the preferred classes for each of its syntactic positions (subject,
      direct object, indirect object, prepositional object). Converting
      these restrictions to context constraints is straightforward.

        The strongest assumption taken in this approach is that verbs are
      considered as {\em forms} not as senses, i.e. selectional restrictions
      for polysemic verbs do not distinguish the different verb senses. 

\item Hand written selectional restrictions on verb objects. The hand written
      restrictions were only a small subset covering some sample verbs, and
      are not statistically significant, but will enable us to check whether
      appropriate constraints may perform the task accurately.

        The small set of hand written constraints was used to test the system
      with a model without the problems related to
      overgeneralization presented by the automatically acquired selectional 
      restrictions (see section \ref{sec:experiments-WSD-problems}), and 
      without the problems that may be caused by the assumptions described 
      above for the different kinds of constraints employed.
\end{itemize}

\subsubsection{Performance analysis of the proposed constraints}
\label{sec:experiments-WSD-problems}

   The results of the experiments described in section 
\ref{sec:experiments-WSD-constraints} point that the semantic constraints 
do not significantly improve the performance for WSD respect a most-likely
sense assignation once the POS tag is known. 
 Actually, the effect of the semantic constraints is that while they do correct a 
certain number of noun senses they also turn wrong a similar amount.
 
   Anyway, it must be taken into account that the POS-tagging plus most-likely
sense selection produce almost a 58\% of correct synset selection and a 63\% for correct
WN file code selection, {\em on all words}. If we focus on nouns only, the 
results are still better, 63\% for synsets and 68\% for WN file codes. This is 
due to the fact that the most-likely sense order yielded by WN is based on
sense occurrences in SemCor, so we are using over-fitted knowledge, and we have 
a very high baseline which is difficult to outperform.
\medskip

  Other reasons for the poor contribution of tested constraints are:
\begin{itemize}
\item The conceptual distance and top co-occurrences constraints are poorly
informed heuristics that may not contain any new information that was not in
the most-likely sense heuristic. In addition, the later model was acquired 
from a rather small training corpus, which causes the co-occurrences estimations
to be unreliable, specially those involving verb classes since most verbs
in WordNet constitute a class on their own.
\item The automatically acquired selectional restrictions where acquired
taking into account the syntactic position of the noun, and when the constraints
are applied by relaxation, this information is not available, so the first noun
to the left of the verb is considered to be its subject, the first to the 
right the direct object, and so on. This may cause that many constraints are 
either improperly applied or not applied at all when they should be.
\item The automatically acquired selectional restrictions were acquired
from only positive examples, which may lead to over-generalization, thus
they may be applied in cases when they should not.
\item The selectional restrictions do not consider verb sense ambiguity, so, 
a restriction stating that the object of verb {\em eat} must be of class
{\em $<$food$>$}, would --wrongly-- be applied in the sentence: 
``{\tt the acid ate the soap cake.}'', where the verb {\em eat} has the 
sense of {\em $<$corrode$>$}.
Thus the {\em $<$food$>$} sense for {\em cake} would be selected, instead of the
correct {\em $<$artifact$>$} sense.
\end{itemize}

\subsubsection{Using a small hand--written model}
\label{sec:experiments-WSD-exemple}

   In order to analyze in detail how relaxation uses the semantic constraints
and to check whether we can expect better results from it, we focused on one verb and
on some of its selectional restrictions and studied the algorithm behaviour.
The choosen verb had to have a high frequency in the corpus and 
have also a high number of disambiguation errors\footnote{Disambiguation errors
                                                  made by most-likely sense
                                                  selection given the POS tag.} 
in its object nouns, in order
to check whether the constraints could solve them. 
Such conditions were matched by several common verbs such as 
{\em to give}, {\em to find}, {\em to hold}, etc.,
probably due to their high ambiguity, and to the fact that they are 
basis to many phrasal verbs. Although a few restrictions for each of them 
were written, focus was set on verb {\em to give} since it was 
the most frequent verb --apart of the ubiquitous {\em to be} and {\em to have}--
with errors in its object nouns in the corpus.
\medskip

  We extracted from SemCor all the sentences containing any form of the verb {\em to give},
and got a small corpus of $6$ Kw and $220$ sentences. We disambiguated them using the 
model for POS-bigrams plus most-likely sense selection. 
Then we extended the language model in two ways: On the one hand, with the 
selectional constraints for the verb {\em to give} automatically acquired by
the \cite{Ribas95} algorithm (see section \ref{sec:semantic}),
which are listed in table \ref{taula:selectional-auto}.
On the other hand, we manually wrote the $5$ selectional restrictions 
presented in table \ref{taula:selectional-manual}.
Verb ambiguity was not considered in any case ({\em give} has $22$ senses in WN, 
but no distinctions were made).  

\begin{table}[htb] \centering
\begin{tabular}{ll}
  $2.85$ & [~{\tt give} SUBJECT~=~{\em $<$act,~human-action$>$}~] \\
  $2.60$ & [~{\tt give} SUBJECT~=~{\em $<$group,~grouping$>$}~] \\
  $1.11$ & [~{\tt give} SUBJECT~=~{\em $<$person,~individual$>$}~] \\
  $5.94$ & [~{\tt give} OBJECT-1~=~{\em $<$rate~(magnitude-relation)$>$}~] \\
  $3.59$ & [~{\tt give} OBJECT-1~=~{\em $<$information~(content)$>$}~] \\
  $3.19$ & [~{\tt give} OBJECT-1~=~{\em $<$message~(communication)$>$}~] \\
  $2.90$ & [~{\tt give} OBJECT-1~=~{\em $<$group,~grouping$>$}~] \\
  $2.34$ & [~{\tt give} OBJECT-1~=~{\em $<$person,~individual$>$}~] \\
  $2.24$ & [~{\tt give} OBJECT-1~=~{\em $<$state$>$}~] \\
  $1.55$ & [~{\tt give} OBJECT-1~=~{\em $<$act,~human-action$>$}~] \\
  $3.93$ & [~{\tt give} OBJECT-2~=~{\em $<$opportunity,~chance$>$}~] \\
  $3.06$ & [~{\tt give} OBJECT-2~=~{\em $<$activity,~behaviour$>$}~] \\
  $2.79$ & [~{\tt give} OBJECT-2~=~{\em $<$attribute$>$}~] \\
  $2.34$ & [~{\tt give} OBJECT-2~=~{\em $<$cognition,~knowledge$>$}~] \\
\end{tabular}
\caption{Automatically acquired selectional restrictions for verb {\em to give}.}
\label{taula:selectional-auto}
\end{table}

\begin{table}[htb] \centering
\begin{tabular}{ll}
  $10.0$ &[~{\tt give} SUBJECT~=~{\em $<$person,~individual$>$}~] \\
  $10.0$ &[~{\tt give} OBJECT-1~=~{\em $<$possession$>$}~] \\
  $10.0$ &[~{\tt give} OBJECT-1~=~{\em $<$time$>$}~] \\
  $10.0$ &[~{\tt give} OBJECT-1~=~{\em $<$freedom,~liberty$>$}~] \\
  $10.0$ &[~{\tt give} OBJECT-1~=~{\em $<$status,~social-state$>$}~]
\end{tabular}
\caption{Hand written selectional restrictions for verb {\em to give}.}
\label{taula:selectional-manual}
\end{table}

   The compatibility values for automatically acquired constraints were computed
from the occurrences in training corpus. For the hand written constraints, the
compatibility was assigned following the same criterion than in the shallow
parsing case. As described in section \ref{experiments-ShP-hybrid}, the 
compatibility value assigned to the constraints was at least as large as the
largest value for a statistical constraint --in this case, the POS bigrams--.
Since the larger compatibility for a POS binay constraint is about $\pm 10$,
this is the compatibility assigned to the semantic constraints.
\medskip

   Thus, the {\em to give} test corpus was disambiguated using the following models: 
\begin{itemize}
\item POS bigrams plus most-likely sense selection.
\item POS bigrams plus most-likely sense selection plus the $14$ automatically
      acquired {\em to give} constraints.
\item POS bigrams plus most-likely sense selection plus the $5$ manually
      written {\em to give} constraints.
\end{itemize}

   Results point out that automatically acquired constraints seem to perform worse than 
hand written constraints. This is due not only to their higher overgeneralization
degree, but also to the fact that a larger number of constraints may imply a 
larger number of conflicts, and thus a larger amount of wrongly resolved conflicts. 
See the examples below for more details.
\medskip

   Although the used selectional restrictions are few in both cases 
($5$ hand--written and $14$ automatically learned), the obtained results
offer a sample of a wide range of possibilities: Some words are corrected to 
the synset level while others only to the WN file code level, some are turned
wrong because of an incorrect application of a constraint that should distinguish 
verb senses, and some others are turned wrong by the incorrect application of 
a constraint that should use more precise syntactic information.

   The following examples show some effects of the selectional 
restrictions presented above.
A reading marked with {\bf p} indicates wrong POS tag 
(and thus, wrong file code and wrong synset).
When marked with {\bf f}, it indicates right POS tag but wrong WN file code 
(and thus, wrong synset).
 A reading marked with {\bf s} indicates wrong synset but right POS tag and file code. 
Readings marked with {\bf t} are test corpus incoherences (i.e. a noun 
POS-tag with a verb sense) and are left out of performance analysis.

  Note that the synset assigned to the verb {\em give} is always  
{\em $<$give~(state,~say)$>$}. This is because the used selectional constraints 
only restrict noun senses, so verbs are assigned their most likely sense.
\medskip

The manual constraints were successful in assigning the right synset to nouns
in which the most-likely heuristic was wrong, as in the following example sentence, 
where the word {\em award} was correctly changed from
the {\em $<$honour$>$} to the {\em $<$prize$>$} synset.
\begin{tabbing} xxxxxxx \= xxxxxxxxxxx \= xxx \= \kill  
  \> \underline{POS + Most Likely}  \\
  \>{\tt a}        \> \>DT \\
  \>{\tt special}  \> \>JJ special adj.all {\em $<$special$>$} \\
  \>{\tt award}    \> {\bf f} \>NN award noun.communication {\em $<$honour$>$} \\
  \>{\tt was}      \> \> VBD be verb.stative {\em $<$have-the-quality-of-being$>$} \\
  \>{\tt given}    \> {\bf f} \>VBN give verb.possession {\em $<$give (state, say)$>$} \\
  \>{\tt to}       \> \>TO \\
  \>{\tt Bob}      \> \>NP \\
  \>{\tt Nordmann} \> \>NP \\
\\
  \> \underline{POS + Most Likely + Hand--written}  \\
  \>{\tt a}        \> \>DT \\
  \>{\tt special}  \> \>JJ special adj.all {\em $<$special$>$} \\
  \>{\tt award}    \> \>NN award noun.possession {\em $<$prize$>$} \\
  \>{\tt was}      \> \>VBD be verb.stative {\em $<$have-the-quality-of-being$>$} \\
  \>{\tt given}    \> {\bf f} \>VBN give verb.possession {\em $<$give (state, say)$>$} \\
  \>{\tt to}       \> \>TO \\
  \>{\tt Bob}      \> \>NP \\
  \>{\tt Nordmann} \> \>NP
\end{tabbing}
\medskip

 In the following case the algorithm assigned the right WN file code but 
not the right synset, i.e, the word {\em host} file code was corrected from
{\em noun.animal} to {\em noun.person}, but the hand--written constraints were
not specific enough to distinguish between the assigned {\em $<$master-of-ceremonies$>$}
synset and the correct {\em $<$host~(adult)$>$} sense.
\begin{tabbing} xxxxxxx \= xxxxxxxxxxx \= xxx \= \kill  
  \> \underline{POS + Most Likely}  \\
  \>{\tt their}    \>\>PP\$  \\
  \>{\tt host}     \>{\bf f}\>NN host noun.animal {\em $<$host (organism)$>$} \\
  \>{\tt gives}    \>{\bf s}\>VBZ give verb.possession {\em $<$give (state,say)$>$}  \\
  \>{\tt them}     \>\>PP \\
  \>{\tt fresh}    \>\>JJ fresh adj.all {\em $<$fresh$>$}    \\
  \>{\tt clothes}  \>\>NNS clothes noun.artifact {\em $<$clothes$>$} \\ 
\\
  \> \underline{POS + Most Likely + Hand-written}  \\ 
  \>{\tt their}    \>\>PP\$ \\
  \>{\tt host}     \>{\bf s}\>NN host noun.person {\em $<$master-of-ceremonies$>$} \\
  \>{\tt gives}    \>{\bf s}\>VBZ give verb.possession {\em $<$give (state, say)$>$} \\
  \>{\tt them}     \>\>PP \\
  \>{\tt fresh}    \>\>JJ fresh adj.all {\em $<$fresh$>$} \\
  \>{\tt clothes}  \>\>NNS clothes noun.artifact {\em $<$clothes$>$}
\end{tabbing}

 In the same case, the automatically acquired model got wrong 
the word {\em host}, because of a restriction conflict:
The automatically acquired model not only includes the constraint 
[{\tt give}~SUBJECT~=~{\em $<$person,~individual$>$}], but also another
constraint on the subject of the verb {\em give}:
[{\tt give}~SUBJECT~=~{\em $<$group,~grouping$>$}].
Although both constraints are correct, they conflict in the word {\em host},
since it may take either a {\em $<$person,~individual$>$} sense or a 
{\em $<$group,~grouping$>$} one ({\em $<$horde$>$}). 
In this particular case, the later was --wrongly-- selected due to the 
higher compatibility value assigned to the second constraint. 
This points out that more context information should be used to 
correctly disambiguate such cases.
\medskip

  The constraints written to better diambiguate senses, may also help
to correct words which would be assigned a wrong POS tag. This happens in the
next example, where the word {\em rein} was assigned a {\bf VB} (verb) tag,
but the selection of the {\em $<$free-rein, rein$>$} synset due to
the manual constraint [{\tt give}~OBJECT-1~=~{\em $<$freedom,~liberty$>$}]
caused the tag to be correctly changed to {\bf NN} (noun).
\begin{tabbing} xxxxxxx \= xxxxxxxxxxx \= xxx \= \kill  
  \> \underline{POS + Most Likely} \\ 
  \>{\tt had}      \>\> VBD have verb.possession {\em $<$have-got, hold$>$} \\
  \>{\tt not}      \>\>RB not adv.all {\em $<$not$>$} \\
  \>{\tt dared}    \>{\bf s}\> VBD dare verb.social {\em $<$presume-to$>$} \\
  \>{\tt to}       \>\> TO \\
  \>{\tt give}     \>{\bf s}\> VB give verb.possession {\em $<$give (state, say)$>$} \\
  \>{\tt rein}     \>{\bf p}\> VB rein verb.change {\em $<$rule, constrol$>$} \\
  \>{\tt to}       \>\> TO \\
  \>{\tt impulses} \>{\bf f}\> NNS impulse noun.attribute {\em $<$momentum$>$} \\
\\
  \> \underline{POS + Most Likely + Hand--written}  \\
  \>{\tt had}      \>\> VBD have verb.possession {\em $<$have-got, hold$>$} \\
  \>{\tt not}      \>\>RB not adv.all {\em $<$not$>$} \\
  \>{\tt dared}    \>{\bf s}\> VBD dare verb.social {\em $<$presume-to$>$} \\
  \>{\tt to}       \>\> TO \\
  \>{\tt give}     \>{\bf s}\> VB give verb.possession {\em $<$give (state, say)$>$} \\
  \>{\tt rein}     \>\> NN rein noun.state {\em $<$free-rein, rein$>$} \\
  \>{\tt to}       \>\> TO \\
  \>{\tt impulses} \>{\bf f}\> NNS impulse noun.attribute {\em $<$momentum$>$} 
\end{tabbing}
\medskip

  On the other hand, the hand--written model also turned wrong two synsets 
where the right choice was the most likely sense. One of them is presented here:
the synset for word {\em rate} was wrongly changed from 
{\em $<$rate~(magnitude-relation)$>$} to {\em $<$rate~(charge-per-unit)$>$}. 
This is due to verb polysemy, since here {\em give} has the  {\em $<$yield$>$} sense
and not the {\em $<$give~(transfer)$>$} one, the constraint 
[{\tt give}~OBJECT-1~=~{\em$<$possession$>$}] should not be applied. The only way
to prevent the constraint from being applied is that it was specific for the
{\em $<$give~(transfer)$>$} sense.
\begin{tabbing} xxxxxxx \= xxxxxxxxxxx \= xxx \= \kill  
  \> \underline{POS + Most Likely}  \\
  \>{\tt This}     \>\>DT   \\
  \>{\tt gives}    \>{\bf s}\>VBZ give verb.possession {\em $<$give (state, say)$>$}\\
  \>{\tt a}        \>\>DT  \\
  \>{\tt rate}     \>\>NN rate noun.time {\em $<$rate (magnitude-relation)$>$}  \\
  \>{\tt of}       \>\>IN  \\
  \>{\tt shear}    \>\>NN shear noun.phenomenon {\em $<$shear$>$} \\
  \>{\tt of}       \>\>IN  \\
  \>{\tt **f}      \>\>NN \\
\\
  \> \underline{POS + Most Likely + Hand--written}  \\
  \>{\tt This}     \>\>DT \\
  \>{\tt gives}    \>{\bf s}\>VBZ give verb.possession {\em $<$give (state, say)$>$} \\
  \>{\tt a}        \>\>DT \\
  \>{\tt rate}     \>{\bf f}\>NN rate noun.possession {\em $<$rate (charge-per-unit)$>$} \\
  \>{\tt of}       \>\>IN \\
  \>{\tt shear}    \>\>NN shear noun.phenomenon {\em $<$shear$>$} \\
  \>{\tt of}       \>\>IN \\
  \>{\tt **f}      \>\>NN
\end{tabbing}

  On the contrary, in the same case, the automatically acquired model selected
the right sense for the word {\em rate}. Since the automatic model does not 
containt the constraint which selects the wrong {\em $<$possession$>$} sense, 
[{\tt give}~OBJECT-1~=~{\em$<$possession$>$}], it is not selected.
In addition, the automatic model does contain a restriction which reinforces
the most-likely option anyway: 
[{\tt give}~OBJECT-1~=~{\em$<$rate~(magnitude-relation)$>$}].
\medskip

 There was also one error made by the manual model due to a wrong constraint 
application caused by insufficient syntactic information.
The word {\em figure} was wrongly taken as the subject
of {\em gives} and the constraint [{\tt give}~SUBJECT~=~{\em $<$person$>$}] gave 
more support to the 
{\em $<$figure~(important-person)$>$} sense and caused it to be chosen instead of the right 
{\em $<$figure~(illustration)$>$}.
\begin{tabbing} xxxxxxx \= xxxxxxxxxxx \= xxx \= \kill  
  \> \underline{POS + Most Likely}  \\
  \>{\tt The}          \>\>DT \\
  \>{\tt temperature}  \>\>NN temperature noun.attribute {\em $<$temperature$>$} \\
  \>{\tt distribution} \>\>NN distribution noun.cognition {\em $<$distribution$>$} \\
  \>{\tt of}           \>\>IN  \\
  \>{\tt figure}       \>\>NN figure noun.communication {\em $<$figure (illustration)$>$} \\
  \>{\tt 4}            \>\>CD  \\
  \>{\tt gives}        \>\>VBZ give verb.possession {\em $<$give (state, say)$>$} \\
  \>{\tt **f}          \>\>NN   \\
  \>{\tt all}          \>{\bf t}\>DT all adj.all {\em $<$all$>$}  \\
  \>{\tt blowing}      \>{\bf t}\>NN blow verb.weather {\em $<$blow$>$}  \\
  \>{\tt rates}        \>\>NNS rate noun.time {\em $<$rate (magnitude-relation)$>$}\\ 
\\
  \> \underline{POS + Most Likely + Hand--written}  \\ 
  \>{\tt The}          \>\>DT \\
  \>{\tt temperature}  \>\>NN temperature noun.attribute {\em $<$temperature$>$} \\
  \>{\tt distribution} \>\>NN distribution noun.cognition {\em $<$distribution$>$} \\
  \>{\tt of}           \>\>IN \\
  \>{\tt figure}       \>{\bf f}\>NN figure noun.person {\em $<$figure (important-person)$>$} \\
  \>{\tt 4}            \>\>CD \\
  \>{\tt gives}        \>\>VBZ give verb.possession {\em $<$give (state, say)$>$}\\
  \>{\tt **f}          \>\>NN \\
  \>{\tt for}          \>\>IN \\
  \>{\tt all}          \>{\bf t}\>DT all adj.all {\em $<$all$>$} \\
  \>{\tt blowing}      \>{\bf t}\>NN blow verb.weather {\em $<$blow$>$} \\
  \>{\tt rates}        \>\>NNS rate noun.time {\em $<$rate (magnitude-relation)$>$}
\end{tabbing}
\medskip

  The results obtained so far point out that relaxation labelling applies properly
the constraints and that it can be used to apply multi-feature models and 
simultaneously solve different NLP tasks. The main causes of the poor results
in WSD are mainly due to the unapropiateness of the used constraint model
and semantic taxonomy.

  Thus, the WSD issue has to be further addressed to obtain better results 
with relaxation labelling algorithms. The solution seems to be in the
direction of taking the appropriate measures to avoid undesired constraint
applications. This will require a model refining and the use of more precise
selectional constraints, which may be adressed through the following issues:
\begin{itemize}
\item Use a shallow parsing model toghether with the WSD model to more
precisely apply the constraints that require syntactic information.
\item Improve the language model, refining the constraints to achieve that 
they take into account not only verb forms but also verb senses.
\item Use constraint which affect not only noun but also verb senses.
\item Define a suitable sense granularity level in WordNet coarser than synset
level, but finer than top-level.
\item Use richer context information, including not only head words, but also
qualifyers, prepositional phrases, etc.
\end{itemize}


%% file: results.tex
\chapter{Comparative Analysis of Results}
\label{cap:results}

   In chapter \ref{cap:experiments} we described the performed experiments
on applying relaxation algorithms to NLP, and reported the obtained results.
 
   In this chapter we will analyze those results and compare the accuracy obtained
with different language models in different tasks, as well as compare our results
with those produced by other systems. Some considerations on performance evaluation
and systems comparison --specially on POS tagging-- can be found in 
section \ref{sec:results-POS-errors}.

\section{Part-of-speech Tagging}
\label{sec:results-POS}

  The experiments on POS tagging, as described in section \ref{sec:experiments-POS},
consisted of tagging the {\bf WSJ} corpus with different language
models. Those models included bigram, trigram, hand-written constraints as well as
automatically learned decision trees. The knowledge contained in the different models was
combined to take advantage of the collaboration between them. We also used a 
HMM bigram tagger and a most-likely-tag algorithm to tag the test set and 
establish a baseline performance. Results are summarized in table \ref{taula:results-POS}.
\medskip

\begin{table}[htb] \centering
\begin{tabular}{|l|r|r|} \hline
      &ambiguous &overall    \\ \hline \hline
ML    &$85.31\%$ &$94.66\%$  \\ \hline
HMM   &$91.75\%$ &$97.00\%$  \\ \hline \hline
B     &$91.35\%$ &$96.86\%$  \\ \hline
T     &$91.82\%$ &$97.03\%$  \\ \hline
BT    &$91.92\%$ &$97.06\%$  \\ \hline \hline
C     &$91.96\%$ &$97.08\%$  \\ \hline
BC    &$92.72\%$ &$97.36\%$  \\ \hline
TC    &$92.82\%$ &$97.39\%$  \\ \hline
BTC   &$92.55\%$ &$97.29\%$  \\ \hline \hline
H     &$86.41\%$ &$95.06\%$  \\ \hline
BH    &$91.88\%$ &$97.05\%$  \\ \hline
TH    &$92.04\%$ &$97.11\%$  \\ \hline
BTH   &$92.32\%$ &$97.21\%$  \\ \hline
CH    &$91.97\%$ &$97.08\%$  \\ \hline
BCH   &$92.76\%$ &$97.37\%$  \\ \hline
TCH   &$92.98\%$ &$97.45\%$  \\ \hline
BTCH  &$92.71\%$ &$97.35\%$  \\ \hline
\end{tabular}
\caption{Results for POS with different language models.
({\bf ML}~stands for most-likely, {\bf B}~for bigrams, {\bf T}~for trigrams, 
{\bf C}~for automatically acquired constraints and {\bf H}~for hand-written constraints.)}
\label{taula:results-POS}
\end{table}
  
  From those results, we concluded the following:

\begin{itemize}
\item When using only bigram information, the relaxation algorithm is
     worse than the bigram HMM tagger with a 90\% confidence rate. This may 
     be indicating a higher sensitivity of relaxation to noise in the model.
\item The use of trigrams, either alone or combined with bigrams yield a small 
     improvement on the average performance, though not at a significant level. 
     That is, the trigram 
     model is slightly better than the bigram model, and the bigram+trigram model 
     is in turn slightly better than the trigram model, but these improvements 
     are not significant.
\item The use of an automatically acquired model based on statistical decision 
     trees described in \cite{Marquez97b}
     produces results slightly higher than the bigrams and/or trigrams models, but
     --as in the previous case-- there is not a significant difference either.
\item The combination of the statistical models (bigram and/or trigram) plus the
     automatically acquired (decision trees) leads to a significant improvement
     at a 99\% confidence rate respect the bigram/trigram model or the use of
     the decision trees alone. This enables us to conclude two important issues:
     First, that the automatically acquired constraint model captures relevant 
     information that was not contained in the n-gram models and vice-versa, 
     since the joint result is better than those obtained by any of the two 
     models alone. Second, that the collaboration of both models was correctly 
     performed by the relaxation algorithm, which proofs that it is able 
     to correctly combine knowledge from different sources.
\item The use of a small set of some twenty hand written constraints improves 
     performance slightly, although not significantly, when added to a model 
     containing the automatically acquired decision-tree constraints. The improvement
     is significant at a 95\% confidence rate when the hand written constraints are
     added to the bigram model or to the bigram+trigram model.
     This yields the conclusion that the
     hand written constraints contain information that was not included in the
     n-gram models --this is quite obvious since hand written constraints were  
     linguistically motivated-- but that this happens to a much smaller extent
     in the case of the automatically acquired model, which is also reasonable,
     since the reduced size of the hand--written model makes it quite likely that
     the modelled phenomena were already captured by the decision--tree model.
\item Since our tagger is able to easily incorporate more knowledge, the obtained
    results are better than other systems that report experiments on WSJ corpus:
    \cite{Brill92,Brill95} reports a 3-4\% error rate, and \cite{Daelemans96a} report 96.7\%
    accuracy. We obtained about 97.4\% accuracy using trigrams and automatically
    acquired constraints. Nevertheless, a more accurate comparison procedure should be 
    established through the use of the same train and test corpus, since --as mentioned
    in section \ref{sec:experiments-POS}-- the results may depend strongly
    not only on the tagset (which should be the same, since all reported
    researches use WSJ corpus) but also on the size of the training and test 
    corpus. Another important point that strongly affects POS taggers performance
    is the noise in the train corpus --which produces a noisy model-- as well
    as in the test corpus. These and other factors affecting taggers evaluation 
    and comparison are discussed in section \ref{sec:results-POS-errors} below.
\end{itemize}

\subsection{Some considerations on error cases}
\label{sec:results-POS-errors}

\subsubsection{On POS taggers evaluation}

  As stated in section \ref{sec:experiments-POS-results}, measuring a tagger
performance through its precision percentage, is a technique which is reaching 
a point where the error in the measure may be higher that the measured performance
improvement: Tests are usually performed over noisy corpora, which may contain
about $5\%$ of tagging errors, and current taggers perform all above $95\%$. 
Thus, the amount of noise in the test corpus is the same order than the 
tagger error rate. This introduces an uncertainty in the evaluation which 
may be larger than the reported improvements from one system to another.

  Some work related to this issue is presented in \cite{Elworthy94b}, who
uses a variable rejection threshold to decide whether a tagger output is
reliable. The effect of the threshold is enabling an efficiency vs. accuracy
trade-off, i.e. a high threshold will produce less erroneous taggings, but
will leave more words ambiguous.
In a similar direction, \cite{Jost94} estimate a lower bound for a tagger 
error rate, depending on the training corpus size.
\medskip

  For instance, if we had a test corpus {\bf A} which we knew to contain about
a $5\%$ of tagging errors and we had a tagger that reporting $100\%$ performance
on that test set, our tagger, far from being accurate, would be yielding a $5\%$
error rate. And the other way round, if our tagger was {\em actually} 
ideal and thus performed {\em actual}
$100\%$ accuracy --that is, perfect ratio over an error-free corpus--,
only $95\%$ accuracy would be reported when tested on corpus {\bf A}, 
since accuracy is computed taking the test corpus as a reference point.
\medskip

  If our tagger instead of being perfect-ratio was, say, $95\%$ accurate on
an error-free corpus, and assuming that the $95\%$ accuracy holds for either
the words correctly or incorrectly tagged in {\bf A}, when evaluated on 
corpus {\bf A} the tagger would report
between $90.25\%$ and $90.50\%$, depending on the ambiguity ratio of
the words in the corpus. In any case, the obtained value would be significantly 
lower than the actual tagger precision.
Computations are detailed in table \ref{taula:tagger-95-eq}. All figures
in this section are computed considering only ambiguous words.
\begin{table}[htb] \centering
\begin{tabular}{|c|c|c|c|} \hline
{\bf test corpus A} & {\bf tagger} & {\bf evaluated as} & {\bf amount} \\ \hline
      OK   &     OK       &    OK             &$95\% \times 0.95 = 90.25\%$ \\
      OK   &     NOK      &    NOK            &$95\% \times 0.05  = 4.75\%$ \\
      NOK  &     OK       &    NOK            &$5\% \times 0.95 = 4.75\%$ \\
      NOK  &     NOK      &    ?              &$5\% \times 0.05 = 0.25\%$ \\ \hline
         \multicolumn{3}{|r|}{\bf total OK}   &$90.25\% - 90.50\%$ \\ \hline
\end{tabular}
\caption{\small{Detailed computation of reported accuracy for an {\em actual} 
         $95\%$ precise tagger when the probability of rightly tagging a 
         correct/incorrect word in {\bf A} is the same ($0.95$).}}
\label{taula:tagger-95-eq}
\end{table}

  Table {\ref{taula:tagger-95-dis} illustrates the same case, but assuming 
that the words correctly tagged in corpus {\bf A} correspond to easier
ambiguities, and thus they would be more easily solved by the tagger 
(e.g. $99\%$ of the times), and the words incorrectly
tagged in {\bf A} correspond to more difficult ambiguities in which
the tagger would make more errors ($81\%$ accuracy to keep the
assumed $95\%$ overall precision). The reported accuracy would then
range from $94.05\%$ to $98.10\%$ depending on the ambiguity
ratio of the corpus. 
\begin{table}[htb] \centering
\begin{tabular}{|c|c|c|c|} \hline
   {\bf test corpus A} &{\bf tagger} & {\bf evaluated as}  & {\bf amount}\\ \hline
      OK   &    OK    &     OK          &$95\% \times 0.99 = 94.05\%$ \\
      OK   &    NOK   &     NOK         &$95\% \times 0.01  = 0.95\%$ \\
      NOK  &    OK    &     NOK         &$5\% \times 0.19 = 0.95\%$ \\
      NOK  &    NOK   &     ?           &$5\% \times 0.81 = 4.05\%$ \\ \hline
        \multicolumn{3}{|r|}{\bf total OK} &$94.05\% - 98.10\%$ \\ \hline
\end{tabular}
\caption{\small{Detailed computation of reported accuracy for an {\em actual} $95\%$ 
         precise tagger when the probability of rightly tagging a correct/incorrect 
         word in {\bf A} is $0.99/0.81$.}}
\label{taula:tagger-95-dis}
\end{table}

For a corpus with low ambiguity ratio, words 
wrongly tagged both in the test corpus and in the tagger output would
have higher probability of coincidence, and thus of being computed
as a correct tag. For higher ambiguity ratios, this coincidence would
be less likely, and the tagger output would be more often correctly 
computed as an error. Figure \ref{fig:eval-99} shows how the reported 
tagger accuracy would vary depending on the ambiguity ratio of test corpus.

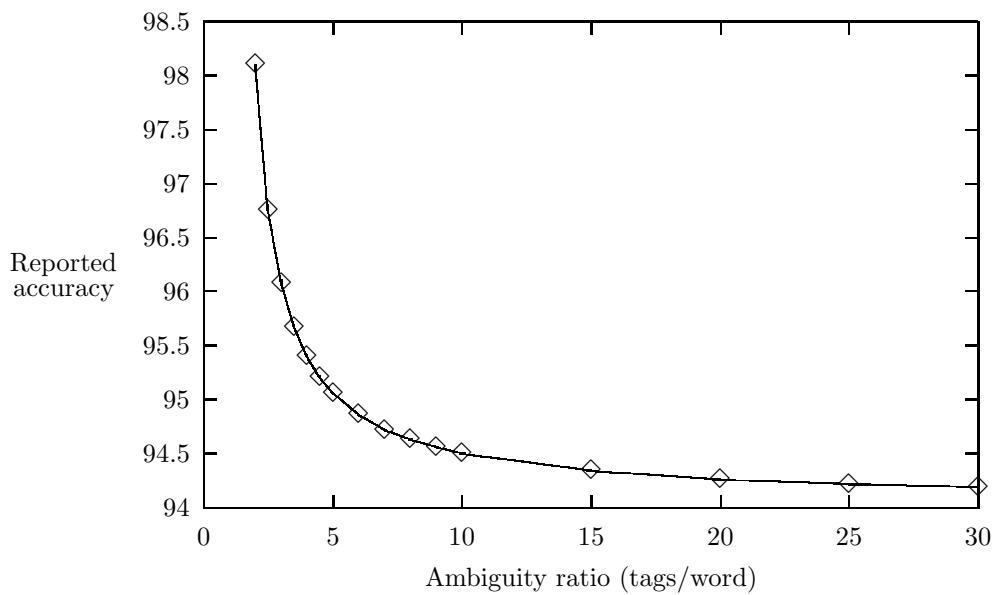
\begin{figure}[htb]
\begin{center}
\input{eval-99.tex}
\caption{\small{Reported accuracy, as a function of the ambiguity ratio, for a 
         an {\em actual} $95\%$ precise tagger when the probability of rightly 
         tagging a correct/incorrect word in {\bf A} is $0.99/0.81$.}}
\label{fig:eval-99}
\end{center}
\end{figure}

  Since the ambiguity ratio for a given corpus is a fixed and easily 
computable value, the main factor affecting the reported accuracy is
the distribution of errors between the correctly and the incorrectly
tagged parts of the test corpus. For instance, table \ref{taula:evaluation}
shows how the evaluated precision of our actual $95\%$ precise tagger 
changes depending on how the probability that the tagger correctly tags a 
word is distributed between words correctly/incorrectly tagged in the 
test corpus {\bf A}. The table is computed taking as the ambiguity ratio
for ambiguous words the value $2.5$ obtained from the {\bf WSJ} corpus used
in the experiments described in section \ref{sec:results-POS}.
\begin{table}[htb] \centering
\begin{tabular}{|c|c|c|} \hline
 probability that tagger is OK &probability that tagger is OK      &  \\
 when {\bf A} is OK ($95\%$ of {\bf A}) &when {\bf A} is NOK ($5\%$ of {\bf A}) & reported accuracy\\ \hline
  $0.950$                 &    $0.950$               &   $90.42\%$ \\
  $0.955$                 &    $0.855$               &   $91.21\%$ \\
  $0.960$                 &    $0.760$               &   $92.00\%$ \\
  $0.965$                 &    $0.665$               &   $92.79\%$ \\
  $0.970$                 &    $0.570$               &   $93.58\%$ \\
  $0.975$                 &    $0.475$               &   $94.38\%$ \\
  $0.980$                 &    $0.380$               &   $95.17\%$ \\
  $0.985$                 &    $0.285$               &   $95.96\%$ \\
  $0.990$                 &    $0.190$               &   $96.75\%$ \\
  $0.995$                 &    $0.095$               &   $97.54\%$ \\
  $1.000$                 &    $0.000$               &   $98.33\%$ \\ \hline
\end{tabular}
\caption{\small{Reported accuracy, as a function of the probability of rightly 
         tagging a correct/incorrect word in {\bf A}, for
         an {\em actual} $95\%$ precise tagger when the 
         corpus ambiguity ratio is $2.5$.}}
\label{taula:evaluation}\medskip
\end{table}

  The first column presents the probability that the tagger correctly
tags a word that has its right tag in the test corpus. The second column
shows the probability that the tagger chooses the right tag for
a word that was wrongly tagged in the test corpus. Both probabilities
are set in such a way that the overall tagger actual performance keeps
being $95\%$. Third column shows how the reported performance for
our tagger would vary between $90.4\%$ and $98.3\%$ only depending
on the tagger behaviour on words that are right or wrongly tagged 
in the test corpus, that is, on to what extent the tagger makes the same 
errors than those found in the corpus {\bf A}.
\medskip

  All this indicates that the reported accuracy of a tagger does not
depend only on the tagset and the train and test corpora sizes, but
also on the corpus itself, specially on its ambiguity ratio and on
how the tagger behaves over errors in the test corpus. That is, if 
the train corpus contains the same kind of errors that the
test corpus --which is quite likely since they are usually 
different parts of the same corpus-- the tagger will probably
learn and make those errors. This will cause the probability of assigning
a right tag to a wrongly tagged word to be lower than for well-tagged words, 
and thus, the tagger performance will be over-evaluated, since more errors will
be computed as right tags. If, on the contrary, 
the tagger makes a  similar proportion of errors in right and wrong
tagged words, it will be drastically under-evaluated.
\medskip

 This makes very difficult to compare systems, since they must be
trained and evaluated in the same corpora to be comparable. In
addition, it makes clear that it is not possible --and nonsense--
achieving further results on POS-tagging using noisy test corpora and
that either error-free test corpora are used, or the distortion
on reported performance must be computed using the ambiguity ratio and
the tagger error distribution over correct/incorrect words in the test corpus.
\medskip

\subsubsection{Some sample error cases}

  Although a systematic study of corpus errors has not been performed,  
we analyzed some cases in WSJ corpus in which our tagger made a 
larger percentage of errors. The main causes of error were identified:
one of them was the above discussed issue of mistagged words in the test corpus,
another, the noise in the train set, and finally, semantic ambiguities 
that could not be solved only with morphosyntactic information. Some samples
of each follow.
\medskip

 Unsolvable ambiguities include the case of semantic ambiguities which can 
only be solved with domain, discourse or context semantic information, such 
as the Noun--Adjective 
ambiguity for word {\em metal} in the phrase {\em the metal container}. 
It could be an adjective, meaning {\em the metallic container}, or a noun,
meaning {\em the container full of metal}. This kind of errors are beyond the 
scope of most current taggers, since they usually deal only with syntactic
and/or morphological information. Nevertheless, our flexible system is able 
to cope with multiple--source knowledge, and constraints relating semantic
and morphological features
could be used to solve this kind of ambiguities in the same way they were
used in section \ref{sec:experiments-WSD} to perform word sense disambiguation
combined with POS-tagging.
\medskip

  Another error case, much more frequent, is the noise in the test and training
corpora. For instance, the WSJ corpus used for the experiments described in section
\ref{sec:results-POS} contains noise (mistagged words) that affects
both the training and the test sets. 

  The noise in the training set
produces noisy --and so less precise-- models. If the same linguistic
structure is not coherently tagged in all its occurrences in the train 
corpus, the model is not correctly estimated and that structure will be
wrongly tagged when the model is applied. If the noise is strong enough
it may cause a certain linguistic structure to be tagged in a most-likely
basis when the model is supposed to do better than that.

  The noise in the test set produces a wrong estimation of accuracy, 
since correct answers are computed as wrong and vice-versa, as was discussed
above.
\medskip

  Samples of some very frequent structures that contain
a very high noise level are the following:

\begin{enumerate}
\item Verb participle forms are sometimes tagged as {\em VBN} (verb-participle)
and in other sentences with no structural differences they are tagged as 
{\em JJ} (adjective). 
\begin{itemize}
\item {\tt\ldots\ failing\_VBG to\_TO voluntarily\_RB submit\_VB \\
       the\_DT {\em requested\_VBN} information\_NN \ldots}
\item {\tt\ldots\ a\_DT large\_JJ sample\_NN of\_IN {\em married\_JJ} women\_NNS \\
       with\_IN at\_IN least\_JJS one\_CD child\_NN \ldots}
\end{itemize}

\item  Another structure not coherently tagged are noun
chains when the nouns are ambiguous and can be also adjectives. Although 
this may obey in many cases to semantic considerations, the same sentence
with the same meaning appears tagged with all possible combinations.
\begin{itemize}
\item {\tt\ldots\ Mr.\_NNP Hahn\_NNP ,\_, the\_DT 62-year-old\_JJ chairman\_NN and\_CC \\
      {\em chief\_NN executive\_JJ officer\_NN} of\_IN Georgia-Pacific\_NNP Corp.\_NNP \ldots}
\item {\tt\ldots\ Burger\_NNP King\_NNP 's\_POS {\em chief\_JJ executive\_NN officer\_NN} ,\_,\\
       Barry\_NNP Gibbons\_NNP ,\_, stars\_VBZ in\_IN ads\_NNS saying\_VBG \ldots}
\item {\tt\ldots\ and\_CC Barrett\_NNP B.\_NNP Weekes\_NNP ,\_, chairman\_NN ,\_, \\
       president\_NN and\_CC {\em chief\_JJ executive\_JJ officer\_NN} .\_. }
\item {\tt\ldots\ the\_DT company\_NN includes\_VBZ Neil\_NNP Davenport\_NNP ,\_, \\
       47\_CD ,\_, president\_NN and\_CC {\em chief\_NN executive\_NN officer\_NN} ;\_: \ldots}
\end{itemize}
\end{enumerate}

  Since the occurrence of the incoherently tagged structure are not
isolated cases --the number of occurrences of the different combinations
are the same order of magnitude--, the noise in these linguistic structures
causes its associated model to be almost most-likely choosing. 
  With respect to the test corpus, a significant amount of occurrences of those noisy 
structures will distort the reported performance.

\section{Shallow Parsing}
\label{sec:results-shallow}

   The experiments on shallow parsing, described in section \ref{sec:experiments-ShP},
consisted of analyzing the same test corpus using different language models and
different analyzers. The used analyzers were the constraint--oriented CG-2 parser 
\cite{Tapanainen96} and the relaxation labelling algorithm. Two kinds of 
language models were employed: the statistically
collected, based on bigrams ({\bf B}) and trigrams ({\bf T}) of shallow syntactic
tags, and the hand--written CG model with linguistic motivation ({\bf C}).
We also used the hybrid models obtained merging them. Results are 
summarized in table \ref{taula:results-ShP}.

\begin{table}[ht] \centering
\begin{tabular}{|l|c|c|} \hline
           &{\bf CG-2 parser}    &{\bf Relaxation labelling} \\
           &{\bf precision - recall} &{\bf precision - recall} \\ \hline \hline
  {\bf C}  &$90.8\%-99.7\%$      &$93.3\%-98.4\%$      \\ \hline
  {\bf forced-C}&$95.0\%-95.0\%$ &$95.8\%-95.8\%$      \\ \hline \hline 
  {\bf B}  &$-$                  &$87.4\%-88.0\%$      \\ \hline
  {\bf T}  &$-$                  &$87.6\%-88.4\%$      \\ \hline
  {\bf BT }&$-$                  &$88.1\%-88.8\%$      \\ \hline
  {\bf forced-BT}&$-$            &$88.5\%-88.5\%$      \\ \hline \hline
  {\bf BC} &$-$                  &$96.0\%-97.0\%$      \\ \hline
  {\bf TC} &$-$                  &$95.9\%-97.0\%$      \\ \hline
  {\bf BTC}&$-$                  &$96.1\%-97.2\%$      \\ \hline
  {\bf forced-BTC}&$-$           &$96.7\%-96.7\%$      \\ \hline
\end{tabular}
\caption{Results for Shallow Parsing with different language models}
\label{taula:results-ShP}
\end{table}

  The conclusions we can derive from the obtained results are:

\begin{itemize}
\item Relaxation disambiguates more words than the CG-2 constraint based parser
     when using the same language model. This is due to the fact that the rules
     are applied in a weighted manner, while the CG-2 parser applies them strictly
     in an established priority order. 
     This has the effect of choosing one among all the possible readings when a 
     small weight difference appears. This obviously causes a higher precision but 
     lower recall for the relaxation algorithm.

     For instance, if a certain reading $R_1$ was
     affected by a {\bf SELECT} constraint and another
     reading $R_2$ for the same word was affected by two different {\bf SELECT} 
     constraints, since CG-2 there are no reasons to discard any of the two 
     readings, the word would be left ambiguous.
     On the other hand, when using relaxation applies, all constraints are applied
     in parallel. Then, the reading $R_2$ would receive a higher support than 
     $R_1$, since it has two positive contributions (constraints) against only 
     one contribution for $R_1$,. This would have the effect of choosing $R_2$ 
     as the best candidate.
\item When forced\footnote{The {\em forced} rows in table \ref{taula:results-ShP} 
                           shows the results when the algorithm was forced to complete 
                           disambiguate all words by randomly choosing a reading.}
     to randomly choose an unique reading for each word among the
     remaining possibilities, relaxation performs significantly better --at a 99\%
     confidence rate-- than the CG-2 parser. This is due to the fact that it had 
     already disambiguated more words, so it is choosing randomly in less cases 
     than CG-2. 
     
     It could be argued that the CG model was written to be used in a CG-2 like
     ordered application, and that there is no point in applying it in parallel, 
     Nevertheless, our experiments show that relaxation labelling results are 
     significantly better --in terms of precision-- than those of CG-2. 
     Maybe this difference could be increased if the constraint model was developed
     under an order--free perspective, more adequate to RL needs. These results
     proof the ability of relaxation labelling to accurately apply linguistic rules, 
     as well as to perform NLP tasks different than POS tagging.
\item When the statistical models are used alone, results are clearly worse
     than the linguist-written
     model. This is very likely caused by the difficulty of the shallow parsing task,  
     which is not as easy to capture in a n-gram model as a simpler task such as
     part-of-speech tagging.
\item The hybrid models produce less ambiguous output than the other models, that is,
     they get a higher precision and a lower recall. The combination of the
     linguistic plus statistical information has also the effect of raising the 
     performance when forced to random disambiguation. The {\bf forced-BTC} results are  
     significantly better (99\% confidence rate) than the {\bf forced-C}. results, 
     indicating that the both kinds of knowledge are correctly cooperating when 
     applied by the relaxation algorithm. That is, the {\bf BT} model contains
     information that was not in the {\bf C} model, and is a useful help to
     further disambiguate the cases where the linguistic model has not enough
     information.
\end{itemize}

\section{Word Sense Disambiguation}
\label{sec:results-WSD}

  The experiments performed on word sense disambiguation reported
in section \ref{sec:experiments-WSD} were neither as
extensive nor as intensive as those performed for the POS tagging or 
shallow parsing tasks. The main reason for that was the lack of 
appropriate linguistic resources (sense tagged corpora, hand--written
semantic constraints, \ldots) and the high labour cost necessary to
develop them. Nevertheless, we tried different statistically acquired
semantic constraints, as well as selectional restrictions which
had been machine--learned by \cite{Ribas95}. We also used a few 
hand written selectional restrictions for some very frequent verbs
such as {\em to give}, {\em to find} or {\em to hold}.

  Although those experiments on applying relaxation labelling to word sense 
disambiguation can be considered prospective, the results obtained up to date 
enable us to draw the following conclusions:

\begin{itemize}
\item The sense disambiguation experiments performed with a reduced set of 
     hand-written selectional restrictions for a particular verb show that
     relaxation labelling is able to deal with multi-feature constraints models
     and to perform several NLP tasks in parallel, in this case, POS tagging 
     and WSD. Examples in section \ref{sec:experiments-POS} show how the semantic
     constraints can be useful not only for performing word sense disambiguation
     but also for assigning the right part-of-speech, and 
     proof the ability of relaxation labelling for taking advantage of 
     cross-feature constraints.
\item The experiments also point out that the constraints ruling sense disambiguation 
     must be very precise and use syntactic information as well as distinguish 
     among verb senses, since an important amount of the errors were due to an incorrect 
     application of constraints that were not specific enough.
     Obtaining more general results for WSD would be possible with a better WSD model.
\item The automatically acquired selectional restrictions for verb objects used
     in the experiments should
     be more strictly applied, that is, applying the selectional restriction for
     a verb subject to the first name to the left of the verb, with few or no other 
     warranty of it being the real subject, can produce wrong sense selections. This
     loose application criterion was selected due to the lack of syntactic
     information in the model.
\item The selectional restrictions acquired in \cite{Ribas95} are, in some cases,
     over-generalized. This, as the author indicates, is due to the lack of negative
     examples, and to the noise introduced by the verbs polysemy. This also
     leads to undesired applications of selectional restrictions..
\item SemCor is not a good test-bench for WSD, since it provides a too small
     training corpus, and the synset level is too fine--grained to perform WSD. 
     Nevertheless, it is one of the few publicly available sense--tagged corpora.
\item Other heuristics tested as possible sense constraints, such as conceptual
     distance and pairwise tops co-occurrence are not significantly useful to help to 
     disambiguate word senses, since although they do correct some sense selections,
     they also spoil a similar amount.
\end{itemize}

    The WSD task should be further addressed from constraint-based language models
and relaxation labelling algorithm. Some directions in which the described problems
can be faced are:
\begin{itemize}
\item Use as senses a less fine--grained set of categories than WordNet synsets, but
     not as coarse as WordNet tops or file codes. That could improve the 
     performance not only when using heuristics such as conceptual distance or 
     top co-occurrence heuristic, but also when using selectional restrictions.
\item Another possibility in the same direction is the use of 
     the {\em Top Ontology} classes or {\em Domains} defined and used in the 
     EuroWordNet project. 
\item Conceptual taxonomies different than WordNet, such as those developed in
     the MicroCosmos \cite{Mahesh95} or Upper Model \cite{Bateman95} projects, 
     could also be taken into account.
\item Use syntactical information to properly apply selectional restrictions. This could
     be achieved combining shallow parsing and WSD language models.
\item Use selectional restrictions that constraint an object sense not only depending on the
     verb, but on the verb class or sense. That would filter out the noise derived 
     from verb polysemy.
\end{itemize}


%% file: eval-99.tex
\setlength{\unitlength}{0.240900pt}
\ifx\plotpoint\undefined\newsavebox{\plotpoint}\fi
\sbox{\plotpoint}{\rule[-0.200pt]{0.400pt}{0.400pt}}%
\begin{picture}(1500,900)(0,0)
\font\gnuplot=cmr10 at 10pt
\gnuplot
\sbox{\plotpoint}{\rule[-0.200pt]{0.400pt}{0.400pt}}%
\put(220.0,113.0){\rule[-0.200pt]{0.400pt}{184.048pt}}
\put(220.0,113.0){\rule[-0.200pt]{4.818pt}{0.400pt}}
\put(198,113){\makebox(0,0)[r]{94}}
\put(1416.0,113.0){\rule[-0.200pt]{4.818pt}{0.400pt}}
\put(220.0,198.0){\rule[-0.200pt]{4.818pt}{0.400pt}}
\put(198,198){\makebox(0,0)[r]{94.5}}
\put(1416.0,198.0){\rule[-0.200pt]{4.818pt}{0.400pt}}
\put(220.0,283.0){\rule[-0.200pt]{4.818pt}{0.400pt}}
\put(198,283){\makebox(0,0)[r]{95}}
\put(1416.0,283.0){\rule[-0.200pt]{4.818pt}{0.400pt}}
\put(220.0,368.0){\rule[-0.200pt]{4.818pt}{0.400pt}}
\put(198,368){\makebox(0,0)[r]{95.5}}
\put(1416.0,368.0){\rule[-0.200pt]{4.818pt}{0.400pt}}
\put(220.0,453.0){\rule[-0.200pt]{4.818pt}{0.400pt}}
\put(198,453){\makebox(0,0)[r]{96}}
\put(1416.0,453.0){\rule[-0.200pt]{4.818pt}{0.400pt}}
\put(220.0,537.0){\rule[-0.200pt]{4.818pt}{0.400pt}}
\put(198,537){\makebox(0,0)[r]{96.5}}
\put(1416.0,537.0){\rule[-0.200pt]{4.818pt}{0.400pt}}
\put(220.0,622.0){\rule[-0.200pt]{4.818pt}{0.400pt}}
\put(198,622){\makebox(0,0)[r]{97}}
\put(1416.0,622.0){\rule[-0.200pt]{4.818pt}{0.400pt}}
\put(220.0,707.0){\rule[-0.200pt]{4.818pt}{0.400pt}}
\put(198,707){\makebox(0,0)[r]{97.5}}
\put(1416.0,707.0){\rule[-0.200pt]{4.818pt}{0.400pt}}
\put(220.0,792.0){\rule[-0.200pt]{4.818pt}{0.400pt}}
\put(198,792){\makebox(0,0)[r]{98}}
\put(1416.0,792.0){\rule[-0.200pt]{4.818pt}{0.400pt}}
\put(220.0,877.0){\rule[-0.200pt]{4.818pt}{0.400pt}}
\put(198,877){\makebox(0,0)[r]{98.5}}
\put(1416.0,877.0){\rule[-0.200pt]{4.818pt}{0.400pt}}
\put(220.0,113.0){\rule[-0.200pt]{0.400pt}{4.818pt}}
\put(220,68){\makebox(0,0){0}}
\put(220.0,857.0){\rule[-0.200pt]{0.400pt}{4.818pt}}
\put(423.0,113.0){\rule[-0.200pt]{0.400pt}{4.818pt}}
\put(423,68){\makebox(0,0){5}}
\put(423.0,857.0){\rule[-0.200pt]{0.400pt}{4.818pt}}
\put(625.0,113.0){\rule[-0.200pt]{0.400pt}{4.818pt}}
\put(625,68){\makebox(0,0){10}}
\put(625.0,857.0){\rule[-0.200pt]{0.400pt}{4.818pt}}
\put(828.0,113.0){\rule[-0.200pt]{0.400pt}{4.818pt}}
\put(828,68){\makebox(0,0){15}}
\put(828.0,857.0){\rule[-0.200pt]{0.400pt}{4.818pt}}
\put(1031.0,113.0){\rule[-0.200pt]{0.400pt}{4.818pt}}
\put(1031,68){\makebox(0,0){20}}
\put(1031.0,857.0){\rule[-0.200pt]{0.400pt}{4.818pt}}
\put(1233.0,113.0){\rule[-0.200pt]{0.400pt}{4.818pt}}
\put(1233,68){\makebox(0,0){25}}
\put(1233.0,857.0){\rule[-0.200pt]{0.400pt}{4.818pt}}
\put(1436.0,113.0){\rule[-0.200pt]{0.400pt}{4.818pt}}
\put(1436,68){\makebox(0,0){30}}
\put(1436.0,857.0){\rule[-0.200pt]{0.400pt}{4.818pt}}
\put(220.0,113.0){\rule[-0.200pt]{292.934pt}{0.400pt}}
\put(1436.0,113.0){\rule[-0.200pt]{0.400pt}{184.048pt}}
\put(220.0,877.0){\rule[-0.200pt]{292.934pt}{0.400pt}}
\put(0,495){\makebox(0,0){Reported}}
\put(0,450){\makebox(0,0){accuracy}}
\put(828,0){\makebox(0,0){Ambiguity ratio (tags/word)}}
\put(220.0,113.0){\rule[-0.200pt]{0.400pt}{184.048pt}}
\put(301,809){\usebox{\plotpoint}}
\multiput(301.58,789.57)(0.496,-5.821){37}{\rule{0.119pt}{4.680pt}}
\multiput(300.17,799.29)(20.000,-219.286){2}{\rule{0.400pt}{2.340pt}}
\multiput(321.58,570.57)(0.496,-2.752){39}{\rule{0.119pt}{2.271pt}}
\multiput(320.17,575.29)(21.000,-109.286){2}{\rule{0.400pt}{1.136pt}}
\multiput(342.58,459.86)(0.496,-1.746){37}{\rule{0.119pt}{1.480pt}}
\multiput(341.17,462.93)(20.000,-65.928){2}{\rule{0.400pt}{0.740pt}}
\multiput(362.58,392.77)(0.496,-1.160){37}{\rule{0.119pt}{1.020pt}}
\multiput(361.17,394.88)(20.000,-43.883){2}{\rule{0.400pt}{0.510pt}}
\multiput(382.58,347.85)(0.496,-0.829){37}{\rule{0.119pt}{0.760pt}}
\multiput(381.17,349.42)(20.000,-31.423){2}{\rule{0.400pt}{0.380pt}}
\multiput(402.58,315.61)(0.496,-0.595){39}{\rule{0.119pt}{0.576pt}}
\multiput(401.17,316.80)(21.000,-23.804){2}{\rule{0.400pt}{0.288pt}}
\multiput(423.00,291.92)(0.588,-0.498){65}{\rule{0.571pt}{0.120pt}}
\multiput(423.00,292.17)(38.816,-34.000){2}{\rule{0.285pt}{0.400pt}}
\multiput(463.00,257.92)(0.858,-0.496){45}{\rule{0.783pt}{0.120pt}}
\multiput(463.00,258.17)(39.374,-24.000){2}{\rule{0.392pt}{0.400pt}}
\multiput(504.00,233.92)(1.352,-0.494){27}{\rule{1.167pt}{0.119pt}}
\multiput(504.00,234.17)(37.579,-15.000){2}{\rule{0.583pt}{0.400pt}}
\multiput(544.00,218.92)(1.746,-0.492){21}{\rule{1.467pt}{0.119pt}}
\multiput(544.00,219.17)(37.956,-12.000){2}{\rule{0.733pt}{0.400pt}}
\multiput(585.00,206.92)(2.059,-0.491){17}{\rule{1.700pt}{0.118pt}}
\multiput(585.00,207.17)(36.472,-10.000){2}{\rule{0.850pt}{0.400pt}}
\multiput(625.00,196.92)(3.802,-0.497){51}{\rule{3.107pt}{0.120pt}}
\multiput(625.00,197.17)(196.550,-27.000){2}{\rule{1.554pt}{0.400pt}}
\multiput(828.00,169.92)(7.438,-0.494){25}{\rule{5.900pt}{0.119pt}}
\multiput(828.00,170.17)(190.754,-14.000){2}{\rule{2.950pt}{0.400pt}}
\multiput(1031.00,155.93)(15.367,-0.485){11}{\rule{11.643pt}{0.117pt}}
\multiput(1031.00,156.17)(177.835,-7.000){2}{\rule{5.821pt}{0.400pt}}
\multiput(1233.00,148.93)(22.529,-0.477){7}{\rule{16.340pt}{0.115pt}}
\multiput(1233.00,149.17)(169.086,-5.000){2}{\rule{8.170pt}{0.400pt}}
\put(301,809){\raisebox{-.8pt}{\makebox(0,0){$\Diamond$}}}
\put(321,580){\raisebox{-.8pt}{\makebox(0,0){$\Diamond$}}}
\put(342,466){\raisebox{-.8pt}{\makebox(0,0){$\Diamond$}}}
\put(362,397){\raisebox{-.8pt}{\makebox(0,0){$\Diamond$}}}
\put(382,351){\raisebox{-.8pt}{\makebox(0,0){$\Diamond$}}}
\put(402,318){\raisebox{-.8pt}{\makebox(0,0){$\Diamond$}}}
\put(423,293){\raisebox{-.8pt}{\makebox(0,0){$\Diamond$}}}
\put(463,259){\raisebox{-.8pt}{\makebox(0,0){$\Diamond$}}}
\put(504,235){\raisebox{-.8pt}{\makebox(0,0){$\Diamond$}}}
\put(544,220){\raisebox{-.8pt}{\makebox(0,0){$\Diamond$}}}
\put(585,208){\raisebox{-.8pt}{\makebox(0,0){$\Diamond$}}}
\put(625,198){\raisebox{-.8pt}{\makebox(0,0){$\Diamond$}}}
\put(828,171){\raisebox{-.8pt}{\makebox(0,0){$\Diamond$}}}
\put(1031,157){\raisebox{-.8pt}{\makebox(0,0){$\Diamond$}}}
\put(1233,150){\raisebox{-.8pt}{\makebox(0,0){$\Diamond$}}}
\put(1436,145){\raisebox{-.8pt}{\makebox(0,0){$\Diamond$}}}
\end{picture}

%% file: conclusions.tex
\chapter{Conclusions}
\label{cap:conclusions}

  This thesis exposes research performed on applying a constraint--based
optimization algorithm --relaxation labelling-- to natural language processing.
The ultimate aim is finding a flexible algorithm able to cope with multi-feature
language models, to integrate knowledge from different sources, and to perform
several NLP tasks, either separately or at the same time.
\medskip

  We tested different parameterizations of the algorithm to find the most
appropriate one to our needs. We then used the algorithm to perform different
NLP tasks: POS tagging, shallow parsing, and word sense disambiguation.

  In addition, we used hybrid language models to perform those tasks. The used 
models included simple statistical information such as bigrams and trigrams,
linguistically motivated hand--written constraints, and automatically acquired
constraints such as decision trees or selectional restrictions. 

  Those language models included also constraints on different word features, and
were used to simultaneously solve more than one NLP task.

\section{Contributions}
\label{sec:concl-contrib}

\subsection{Use of optimization techniques in NLP}
\label{sec:concl-contrib-1}

  One of the main points in this thesis is that optimization techniques in general
and more particularly relaxation labelling are a good option to process natural
language. The main advantage of relaxation labelling over other techniques
is its constraint--based domain description, which makes it very suitable for
many NLP purposes.
\medskip

   When using the relaxation labelling algorithm, the domain is described 
through constraints between variable values. In our case, they are constraints
among word features such as part-of-speech tags, senses, lemmas, etc.

   We proposed and used an extension of the Constraint Grammar formalism, 
in which a {\em compatibility value} is assigned to each constraint, as a powerful
and well-known way of expressing multi-feature context constraints.
\medskip

   With respect to the objective function optimized by the algorithm, we tested
different support functions --which yield different objective functions-- and
choose the most appropriate, the additive function, which was the one that
intuition recommended.

   We also tested a new support function, trying to simulate the sequence 
probability optimized by HMM taggers, but results were not the same, since
relaxation performs a {\em vector optimization}, that is, the objective
function is a vector, and thus, both algorithms are not comparable in these
terms.
\medskip

    As a conclusion, we can state that the optimization algorithm correctly
performs the NLP tasks, when supplied the right constraint--based language model.

    We showed that the model can perform as good as current systems at tasks such
as POS tagging or shallow parsing, and that its flexibility enables it to integrate
and use more sophisticated kinds of knowledge, yielding better results.

\subsection{Application of multi-feature models}
\label{sec:concl-contrib-2}

  Another main point in this thesis is that, for a higher accuracy, 
natural language 
tasks can not be solved independently, since each one needs information from the 
others. This is an idea which is getting support from a growing number of
researchers \cite{Wilks97,Ng97,Oflazer97,Rigau97a,Zavrel97}.
\medskip

  The presented system is able to deal with multi-feature models, that is, 
words are not restricted to have an unique tag, but a set of features. 
  
  The language model can include constraints on any word feature, and thus,
express relationships between one feature for one word and a different one 
for a word in the context, for instance stating that the POS tag for
a given word depends on the semantics of the preceding word.
  The formalism that makes it possible is the Constraint Grammar formalism
described by \cite{Karlsson95}, which was adopted as a standard way
of expressing context rules.
\medskip

  We used multi--feature models to perform shallow parsing and word
sense disambiguation, in the former case the used constraints included information
about word lemmas, syntactic function, POS, case, verb mode, etc. In the later
the used information were POS, senses, WN file codes and lemmas.

  Multi--feature models were also used in POS tagging, although to a minor extent, 
using the word lemma in addition to part-of-speech tag either in constraints 
derived from automatically acquired decision trees or in the case of hand-written 
constraints.
\medskip

  The obtained results proof that the relaxation algorithm properly
combines different kinds of information since it is able to
use constraints relating, for instance, the lemma of a word with
the POS tag of one neighbour word and the syntactic function of 
a third one. 

   Those constraints are properly applied by the 
algorithm and the results are better than when using only one--feature
models. For instance, the application of a small set of 
hand written constraints that used as information not only the POS, 
but also the word lemmas, yielded a significant improvement when
added to a bigram model --which, obviously, used only POS information--.

\subsection{Application of statistical-linguistic hybrid models}
\label{sec:concl-contrib-3}

  The choice to model language through a set of constraints, each of 
them associated to a compatibility value, makes it possible to merge 
knowledge acquired from multiple sources. The way to achieve this is 
converting the different source knowledges to the common formalism of
our language model.

   We successfully applied the relaxation algorithm, and showed that it is
able to integrate knowledge obtained from different sources provided it
is expressed in the form of context constraints.
\medskip

   We used constraints obtained from different sources. For POS tagging, we
combined bigram and trigram constraints with constraints obtained translating
machine--learned decision trees. We used also some sample hand--written constraints.

 For shallow parsing, we used a hybrid model containing bigram and trigram 
information as well as a linguist--written set of constraints.

 For word sense disambiguation, we combined POS bigram constraints with
selectional restrictions on verb objects both automatically acquired \cite{Ribas95}
and manually written. Other kinds of knowledge which were also written in the form of  
context constraints and added to the model were the following:  co-occurrences of
pairs of WordNet top synsets, co-occurrences of pairs of WordNet file codes and
conceptual distance between pairs of noun senses.
\medskip

  The conclusions on this issue are that relaxation 
perfectly combines the different sources knowledge that it is supplied,
and produces results which are better than those that would be obtained
by any of the integrated sources alone, as for instance, in
the POS tagging and shallow parsing experiments reported in 
chapter \ref{cap:experiments}. Nevertheless, experiments in WSD
point out that the knowledge included in the language must be very accurate
to produce good results, specially in complex tasks such as word sense
disambiguation.

\subsection{Simultaneous resolution of NLP tasks}
\label{sec:concl-contrib-4}

  Due to the multi-feature nature of constraints, and to the
parallel way in which relaxation applies them, the algorithm 
can select simultaneously the most appropriate combination for several
word features, that its, it can solve different NLP disambiguations
at the same time.

  This is achieved by assigning to each word not only a unique {\em tag},
but a {\em reading}, that is, a set of features. When a reading is
selected as the correct one, a set of features is being selected and
thus the word is manifold disambiguated. 

  Constraints can express
restrictions on any number of these features, from simple homogeneous
constraints --such as a POS bigram-- to more complex relationships.
The selected reading will be the one that has collected more
positive evidence in the total of its features.

  Modelling word features through {\em readings} has the advantage of 
disabling incoherent combinations,
since readings with, for instance, a {\em verb} POS and a noun sense 
are not considered as candidate readings.
\medskip

  Two of the addressed tasks --shallow parsing and WSD-- were solved 
simultaneously with POS tagging. 
Results showed that constraints on one kind of
knowledge can collaborate to disambiguate the others. 
For instance,
in the WSD experiment described in chapter \ref{cap:experiments}, the hand-written 
constraints for WSD helped in 
correcting some POS tag, since the selection of a noun sense forced
the POS tag to be changed to {\em noun}.

\section{Further Work}
\label{sec:concl-further}

  The research lines opened by this work can be divided in two main groups: those
focused on improving the used constraint language models through both new 
automatic model acquisition algorithms and linguistic manual model development, 
and those aiming to better exploit the relaxation algorithms when applied to 
NLP tasks, including noise analysis, speeding up the algorithm and more
accurate applying the constraint models.
\bigskip

  On the first group, better language models have to be developed, both through the
use of automatic knowledge acquisition techniques and through manual development
of the models. 

\begin{itemize}
\item The future models will have to include constraints on either a 
single disambiguation task or several of them, use single and multi--feature 
constraints, obtained from different automatic or manual sources.

\item For manual constraints, an automatic procedure for computing compatibility
values must be developed. Maximum Entropy seems to be a very promising
approach for this issue.

\item The model for word sense disambiguation should be extended, probably
manually, and will have to include syntactic information to make a
good use of selectional restrictions. This could be achieved combining
the shallow parsing model with the WSD model.
  
\item The WSD model will also have to use a sense codification of an appropriate 
granularity. This could be achieved through the use of the {\em Top Ontology
Classes} defined and used in the EuroWordNet project.

\item To be able to automatically derive accurate language models, the training
corpus must be as noiseless as possible. Thus, debugging techniques should be
applied on available corpora in order to minimize their error rate and to
establish a coherent evaluation method and a upper bound for the achievable accuracy.

 A possible technique to solve this issue could be the comparison of the errors
made by different disambiguators on the same test corpus and the study of the
rate of agreement and disagreement among them. 

\item In the same direction, the distortion in reported performance
introduced by the noise in the test corpus must be further studied, 
to find out whether there is an easy way to estimate it, or on the 
contrary, the only reliable procedure to evaluate a NLP system is 
using noiseless test corpora.
\end{itemize}

  On improving the algorithm performance, several paths are still
to be explored:

\begin{itemize}
\item We plan to further test discrete relaxation, which, as described in section 
\ref{sec:updating}, is equivalent to simulated annealing, and compare
it with continuous relaxation.

\item  Studies on which is the most appropriate normalization factor for support 
values must also be performed, since a correct choice may shorten the number of 
necessary iterations, improve performance, and confirm the assumption that
convergence is the right stopping criterion to choose.

\item  We also plan to investigate whether the fact that relaxation performs slightly
worse than the HMM tagger when both of them use the same bigram model is caused
by a higher sensitivity to noise --and thus, it can be solved using
better training sets-- or else its an intrinsic feature of the algorithms.

\item On improving the algorithm efficiency, a possible future research trend
is the compilation of the context constraints into a finite state transducer
to speed up their application \cite{Roche95,Morawietz97,Tzoukermann97}.
\end{itemize}

  From a more general point of view, we plan to develop language models
as complete as possible for Spanish and Catalan, and use the system
as a basic process in a wider NLP system. The system has already been
integrated it in a NLP environment aimed to perform information extraction, 
as a part of the ITEM project funded by Spanish Research Department 
(CYCIT) TIC96-1243-C03-02.


%% file: appendix1.tex
\chapter{Tagset Descriptions}
\label{app:tagsets}

  This appendix contains the tagsets for the Spanish Novel corpus,
and the WSJ corpus, which were used in the experiments on POS tagging 
described in chapter \ref{cap:experiments}.

\section{WSJ corpus tagset}

\begin{center}
\begin{tabular}{ll|ll}
{\bf Tag}& {\bf Description} & {\bf Tag} & {\bf Description} \\ \hline
  CC     & coordinating conjunction &   TO     & infinitive marker {\em to} \\
  CD     & cardinal number          &   UH     & interjection \\
  DT     & determiner               &   VB     & verb, base form \\
  EX     & exsitential {\em there}  &   VBD    & verb, past tense \\
  FW     & foreign word             &   VBG    & verb, gerund / present participle \\
  IN     & preposition / subordinating conj. &   VBN    & verb, past participle \\
  JJ     & adjective                &   VBP    & verb, non 3rd person singular present\\
  JJR    & adjective, comparative   &   VBZ    & verb, 3rd person singular present \\
  JJS    & adjective, superlative   &   WDT    & {\em wh}-determiner \\
  LS     & list item marker         &   WP     & {\em wh}-pronoun \\
  MD     & modal                    &   WP\$   & possessive {\em wh}-pronoun \\
  NN     & noun, singular or mass   &   WRB    & {\em wh}-adverb \\
  NNS    & noun, plural             &   \#     & pound sign \\
  NP     & proper noun, singular    &   \$     & dollar sign \\
  NPS    & proper noun, plural      &  "       & straight double quote \\
  PDT    & predeterminer            &   ``     & left open double quote \\
  POS    & possessive ending        &   ''     & right close double quote \\
  PP     & personal pronoun         &   `      & left open single quote \\
  PP\$   & possessive pronoun       &   '      & right close single quote \\
  RB     & adverb                   &   (      & left bracket \\
  RBR    & adverb, comparative      &   )      & right bracket \\
  RBS    & adverb, superlative      &   ,      & comma \\
  RP     & particle                 &   .      & sentence final punctuation \\
  SYM    & symbol                   &   :      & colon, semi-colon \\ \hline
\end{tabular}
\end{center}
\newpage

\section{Spanish Novel corpus tagset}

\begin{center}
\begin{tabular}{ll|ll}
{\bf Tag}& {\bf Description} & {\bf Tag} & {\bf Description} \\ \hline
  A   &  adjective                        &  VV  &  verb personal form    \\
  CC  &  coordinating conjunction         &  VP  &  verb personal form + pronoun   \\
  CS  &  subordinating conjunction        &  VS  &  verb personal form + {\em se}   \\
  CA  &  other conjunctions               &  VEV  & {\em ser} personal form    \\
  D   &  adverb                           &  VEP  & {\em ser} personal form + pronoun    \\
  RA  &  preposition+article contracted   &  VHV  & {\em haber} personal form   \\
  RP  &  preposition                      &  VHP  & {\em haber} personal form + pronoun    \\
  TD  &  demonstrative determiner         &  VHS  & {\em haber} personal form + {\em se}   \\
  TP  &  possessive determiner            &  IV   & verb infinitive   \\
  TQ  &  definite quantifyer determiner   &  IP   & verb infinitive + pronoun   \\
  TI  &  indefinite quantifyer determiner &  IS   & verb infinitive + {\em se}   \\
  J   &  article                          &  IEV  & {\em ser} infinitive   \\
  M   &  number                           &  IEP  & {\em ser} infinitive + pronoun   \\
  N   &  noun                             &  IHV  & {\em haber} infinitive   \\
  PD  &  demonstrative pronoun            &  IHP  & {\em haber} infinitive + pronoun   \\
  PN  &  interrogative pronoun            &  IHS  & {\em haber} infinitive + {\em se}  \\
  PL  &  locative pronoun                 &  GV  &  verb gerund   \\
  PO  &  possessive pronoun               &  GP  &  verb gerund + pronoun   \\
  PQ  &  definite quantifyer pronoun      &  GS  &  verb gerund + {\em se}   \\
  PI  &  indefinite quantifyer pronoun    &  GEV  & {\em ser} gerund   \\
  PR  &  relative pronoun                 &  GEP  & {\em ser} gerund + pronoun   \\
  PS  &  personal-subject pronoun         &  GHV  & {\em haber} gerund   \\
  PP  &  personal pronoun                 &  GHP  & {\em haber} gerund + pronoun   \\
  PA  &  other pronouns                   &  GHS  & {\em haber} gerund + {\em se}   \\
  W   &  proper noun                      &  UV   & verb participle   \\
  X   &  {\em se}                         &  UP   & verb participle + pronoun   \\
  Y   &  interjection                     &  US   & verb participle + {\em se}   \\
  Z!` &  punctuation !`                   &  UEV  & {\em ser} participle   \\
  Z!  &  punctuation !                    &  UEP  & {\em ser} participle + pronoun   \\
  Z?` &  punctuation ?`                   &  UHV  & {\em haber} participle   \\
  Z?  &  punctuation ?                    &  UHP  & {\em haber} participle + pronoun  \\
  Z,  &  punctuation ,                    &  UHS  & {\em haber} participle + {\em se}   \\
  Z.  &  punctuation .                    &     \\
  Z;  &  punctuation ;                    &     \\
  Z-  &  punctuation -                    &     \\
  ZX  &  other punctuations               &     \\ \hline
\end{tabular}
\end{center}
\newpage

\section{Susanne Corpus tagset}

  The complete Susanne corpus tagset consists of over $350$ tags which
distinguish gender, number, person, tense and many other morphosyntactic
features. A detailes description can be found in \cite{Sampson95}.
\medskip

   The tagset used in the experiments reported in section \ref{sec:selection}
used the reduced version of the tagset which is listed below. The interested
reader can find the detailed description for each tag in the above 
referenced book by \cite{Sampson95}.
\medskip

\begin{center}
\begin{tabular}{l|l|l|l|l|l|l|l} \hline
!     & CSN   & FB  & MD    & NNT2  & PPHS2 & RRQV & VH0  \\
\$    & CST   & FO  & MF    & NNU   & PPIO1 & RRR  & VHD  \\
(     & CSW   & FW  & ND1   & NNU1  & PPIO2 & RRT  & VHG  \\
)     & DA    & ICS & NN    & NNU2  & PPIS1 & RT   & VHN  \\
,     & DA1   & IF  & NN1   & NP1   & PPIS2 & TO   & VHZ  \\
.     & DA2   & II  & NN2   & NP2   & PPX1  & UH   & VM   \\
...   & DA2R  & IO  & NNJ   & NPD1  & PPX2  & VB0  & VMK  \\
:     & DAR   & IW  & NNJ1  & NPD2  & PPY   & VBDR & VV0  \\
;     & DAT   & JA  & NNJ2  & NPM1  & RA    & VBDZ & VVD  \\
?     & DB    & JB  & NNL   & PN    & REX   & VBG  & VVG  \\
-     & DB2   & JBR & NNL1  & PN1   & RG    & VBM  & VVGK \\
APP\$ & DD    & JBT & NNL2  & PNQO  & RGA   & VBN  & VVN  \\
AT    & DD1   & JJ  & NNO   & PNQS  & RGQ   & VBR  & VVNK \\
AT1   & DD2   & JJR & NNS   & PNQVS & RGQV  & VBZ  & VVZ  \\
BTO   & DDQ   & JJT & NNS1  & PP\$  & RL    & VD0  & XX   \\
CC    & DDQ\$ & LE  & NNS2  & PPH1  & RP    & VDD  & ZZ1  \\
CCB   & DDQV  & MC  & NNSA1 & PPHO1 & RPK   & VDG  &      \\
CS    & EX    & MC1 & NNSB2 & PPHO2 & RR    & VDN  &      \\
CSA   & FA    & MC2 & NNT1  & PPHS1 & RRQ   & VDZ  &      \\ \hline 
\end{tabular}
\end{center}

%% file: appendix2.tex
\chapter{Sample Constraints}
\label{app:constraints}

  This appendix contains some sample constraints which were used
were used in the experiments on POS tagging and on the Shallow Parsing
described in chapter\ref{cap:experiments}.
  Some of the constraints were statistically acquired in
the form of bigrams and trigrams, some others were automatically
extracted using the decision--trees learning algorithm described
in section \ref{sec:decision-trees}, and finally, some of them 
where hand written.

\section{Sample statistically acquired constraints}

   The statistically acquired constraints are binary constraints,
corresponding to bigrams, and ternary constraints, which correspond
to trigram information. Some sample constraints obtained once the
n-gram information has been translated into the
extended Constraint Grammar formalism are the following:
\medskip

   For instance, some binary constraints derived from bigram occurrences
are the following:

   First, a constraint that states a high compatibility for
a {\em verb} tag ({\bf VB}) when preceded by a {\em modal} ({\bf MD}).
\begin{verbatim}
       4.846532  (VB)
                 (-1 (MD));
\end{verbatim}

   The next constraint states a positive compatibility for a {\em determiner}
tag ({\bf DT}) when followed by a {\em noun} ({\bf NN}). 
\begin{verbatim}
       1.760843  (DT)
                 (1 (NN));
\end{verbatim}

   The next constraints state a large incompatibility for a {\em determiner}
tag ({\bf DT}) when followed by a {\em verb} ({\bf VB}), and vice-versa, that is, 
for a {\em verb} tag ({\bf VB}) when preceded by a {\em determiner} ({\bf DT}).
\begin{verbatim}
       -6.776550  (DT)              -6.776550  (VB)
                  (1 (VB));                    (-1 (DT));
\end{verbatim}
\medskip

    Trigram occurrences produce ternary constraints such as the samples below.

    The first constraint expresses that a {\em determiner} tag ({\bf DT}) is 
quite compatible with a right context consisting of an {\em adjective} ({\bf JJ}) 
in the first right position and a {\em noun} ({\bf NN}) in the second.
\begin{verbatim}
       2.352891  (DT)
                 (1 (JJ))
                 (2 (NN));
\end{verbatim}

     The second sample ternary constraint states that a {\em participle} tag 
({\bf VBN}) is rather incompatible with an {\em adjective} ({\bf JJ}) to its
left and a {\em determiner} ({\bf DT}) to its right.
\begin{verbatim}
       -5.682948  (VBN)
                  (-1 (JJ))
                  (1 (DT));
\end{verbatim}

\section{Sample decision--tree learned constraints}

   The sample constraints presented in this section were automatically
acquired by the decision tree learning algorithm \cite{Marquez97a} described 
in section \ref{sec:decision-trees}. They have no linguistic meaning, and
involve a context larger than the immediate one or two words. The context
considered in these constraints consists of two words to the right, three 
to the left and the word form of the focus word.
\medskip

    For instance, the following constraint that the {\em determiner} ({\bf DT})
tag for the word {\em all} is rather incompatible with a context consisting 
of an adverb ({\em RB}) in the first right position and a word with any
of the detailed tags in the second left position.
\begin{verbatim}
    -2.82059 (DT "all")
             (-2 (WDT) OR (VBD) OR (RB) OR (JJ) OR (POS) OR (MD) OR (CC))
             (1 (RB)); 
\end{verbatim}

    The next constraint states the compatibility of an {\em adjective} 
({\bf JJ}) tag for a word that can be also {\em participle} ({\bf VBN})
with a context formed by the specified tags in the two left positions
and in the first right word.
\begin{verbatim}
    1.48853 (JJ)
            (0 (VBN))
            (-1 (VB) OR (IN) OR (DT) OR (<,>))
            (-2 (VBZ))
            (1 (VBP) OR (NNP) OR (NNS) OR (NN) OR (JJ) OR (MD));
\end{verbatim}

    The next two constraints are in fact the same, and state that a
{\bf JJS} tag for the words {\em earliest} or {\em least} is slightly 
compatible with a first left word with any of the detailed tags.
\begin{verbatim}
    0.11497 ("earliest" JJS)
            (1 (VBN) OR (RB) OR (JJ) OR (TO) OR (<(>));

    0.11497 ("least" JJS)
            (1 (VBN) OR (RB) OR (JJ) OR (TO) OR (<(>));
\end{verbatim}

\section{Sample hand--written constraints}

  The third kind of constraints are those which were manually written.
They have some --simple-- linguistic meaning. Their compatibility values
are manually assigned and thus are an arbitrary value. Nevertheless, this value
is chosen to be approximately the same than the highest value obtained for
any automatically acquired constraint (either statistical or learned).

\subsection{POS tagging constraints}

   The following sample constraint were manually written as a part
of a small set aiming both to cover the most frequent errors committed by
the statistical models and to test the ability of the algorithm to deal
with different source information. Thus, although they have some linguistic 
meaning, they are limited and do not cover all possible cases.
\medskip

   For instance, the first constraint states a high compatibility
for a {\em participle} ({\bf VBN}) tag with an auxiliary verb form 
({\bf VAUX}) tagged as a verb\footnote{{\bf VAUX} is previously defined in the grammar
		                       as any possible word form for verbs {\em to be}
				       or {\em to have}. The verb tags are required to 
				       avoid applying the constraint in cases such as
				       nominal uses of {\em being}.},
provided that there is not any other participle nor any phrase break item
(preposition, punctuation or adjective) in between.
\begin{verbatim}
 10 (VBN)
    (*-1 VAUX + (VBD) OR (VB) OR (VBP) OR (VBZ) OR (VBN)
          BARRIER (VBN) OR (IN) OR (<,>) OR (<:>) OR (JJ) OR (JJS) OR (JJR));
\end{verbatim}

   The second sample constraint states a high compatibility for a {\em noun}
({\bf NN}) tag with a left context consisting of a determiner --with no
other nouns in between-- and a right context consisting of no noun tags before
the first noun phrase change (punctuation or determiner).
\begin{verbatim}
 10  (NN)
     (*-1 (DT) BARRIER (NN) OR (NNS))
     (*1 (DT) OR (<.>) OR (<,>) OR (<:>) BARRIER (NN) OR (NNS)));
\end{verbatim}

   The four following constraint deal with comparative constructs of the
form {\em as <adjective> as} and {\em as <adverb> as}. In the WSJ corpus, the 
first {\em as} is tagged as {\bf RB} and the second as {\em IN}. These constraints
state high compatibility for the right choice and high incompatibility
for the wrong one in each case.
\begin{verbatim}
       10  ("as" RB)                  -10  ("as" RB)
           (1 (JJ) OR (RB))                (-1 (JJ) OR (RB))
           (2 ("as"));                     (-2 ("as"));

      -10  ("as" IN)                   10  ("as" IN)
           (1 (JJ) OR (RB))                (-1 (JJ) OR (RB)) 
           (2 ("as"));                     (-2 ("as"));
\end{verbatim}

\subsection{Shallow parsing constraints}

   The constraints used in the shallow parsing experiments
were hand written by a linguist. Although they are not an
exhaustive model, they have a reasonable coverage, and perform
the task accurately. Details about grammar development can be
found in section \ref{sec:experiments-ShP} and in \cite{Voutilainen97}.
\medskip

   Some sample hand written constraints for the shallow parsing task are the
following. The first rule removes the premodifier tag @$>$N from an ambiguous 
reading if somewhere to the right (*1) there is an unambiguous (C) occurrence of
a member of the set $<<<$ (sentence boundary symbols) or the verb tag
@V or the subordinating conjunction tag @CS, and there are no
intervening tags for nominal heads (@NH). 

\begin{verbatim}
       REMOVE (@>N)
              (*1C <<< OR (@V) OR (@CS) BARRIER (@NH));
\end{verbatim}

    Next is a partial rule about coordination, which removes the premodifier 
tag if all three context-conditions are satisfied: (i) the word to be 
disambiguated (0) is not a determiner, numeral or adjective, (ii) the first 
word to the right (1) is an unambiguous coordinating conjunction, 
and (iii) the second word to the right is an unambiguous determiner.

\begin{verbatim}
       REMOVE (@>N)
              (NOT 0 (DET) OR (NUM) OR (A))
              (1C (CC))
              (2C (DET));
\end{verbatim}